\documentclass[sigconf]{acmart}
\usepackage[utf8]{inputenc}
\setcopyright{none}
\renewcommand\footnotetextcopyrightpermission[1]{}
\usepackage{tabularx,booktabs}
\usepackage{amsmath,amsfonts, amsthm}
\usepackage{subcaption}
\usepackage{mathtools}
\usepackage{thmtools, thm-restate}

\usepackage{makecell}
\usepackage[shortlabels]{enumitem}
\usepackage[nobiblatex]{xurl}
\usepackage{multirow}
\usepackage{xcolor}
\usepackage{tikz}
\usepackage{pgfplotstable}
\pgfplotsset{
    compat=1.7,
}

\pgfplotstableread[col sep=comma]{data/motivation_cora.csv}{\loadedtableMotivation}
    \pgfplotstablegetcolsof{\loadedtableMotivation}
    \pgfmathtruncatemacro{\NoOfColsMotivation}{\pgfplotsretval-1}

\pgfplotstableread[col sep=comma]{data/motivation_updated.csv}{\loadedtableMotivationUpdated}
    \pgfplotstablegetcolsof{\loadedtableMotivationUpdated}
    \pgfmathtruncatemacro{\NoOfColsMotivationUpdated}{\pgfplotsretval-1}
    
\pgfplotstableread[col sep=comma]{data/dv_avg_norm_1_test_cora_arch_MLP_eps_0.0.csv}{\loadedtableHomoMLP}
    \pgfplotstablegetcolsof{\loadedtableHomoMLP}
    \pgfmathtruncatemacro{\NoOfColsHomoMLP}{\pgfplotsretval-1}

\pgfplotstableread[col sep=comma]{data/dv_avg_norm_1_test_cora_arch_MMLP_nl_1_eps_0.0.csv}{\loadedtableHomoMMLP}
    \pgfplotstablegetcolsof{\loadedtableHomoMMLP}
    \pgfmathtruncatemacro{\NoOfColsHomoMMLP}{\pgfplotsretval-1}
    
\pgfplotstableread[col sep=comma]{data/dv_avg_norm_1_test_cora_arch_GCN_eps_0.0.csv}{\loadedtableHomoGCN}
    \pgfplotstablegetcolsof{\loadedtableHomoGCN}
    \pgfmathtruncatemacro{\NoOfColsHomoGCN}{\pgfplotsretval-1}
    
\pgfplotstableread[col sep=comma]{data/dv_avg_norm_1_test_cora_ori.csv}{\loadedtableHomoOri}
    \pgfplotstablegetcolsof{\loadedtableHomoOri}
    \pgfmathtruncatemacro{\NoOfColsHomoOri}{\pgfplotsretval-1}

\pgfplotstableread[columns/Dataset/.style={string type}, col sep=comma]{data/Dataset_NonDP_Accuracy.csv}{\loadedtableUtilNonDp}
    \pgfplotstablegetcolsof{\loadedtableUtilNonDp}
    \pgfmathtruncatemacro{\NoOfColsUtilNonDp}{\pgfplotsretval-1}

\pgfplotstableread[col sep=comma]{data/Nl-1_2_Cora_F1.csv}{\loadedtableCoraUtilDp}
    \pgfplotstablegetcolsof{\loadedtableCoraUtilDp}
    \pgfmathtruncatemacro{\NoOfColsCoraUtilDp}{\pgfplotsretval-1}

\pgfplotstableread[col sep=comma]{data/Nl-1_2_Citeseer_F1.csv}{\loadedtableCiteseerUtilDp}
    \pgfplotstablegetcolsof{\loadedtableCiteseerUtilDp}
    \pgfmathtruncatemacro{\NoOfColsCiteseerUtilDp}{\pgfplotsretval-1}

\pgfplotstableread[col sep=comma]{data/Nl-1_2_Pubmed_F1.csv}{\loadedtablePubmedUtilDp}
    \pgfplotstablegetcolsof{\loadedtablePubmedUtilDp}
    \pgfmathtruncatemacro{\NoOfColsPubmedUtilDp}{\pgfplotsretval-1}

\pgfplotstableread[col sep=comma]{data/Nl-1_2_Facebook_page_F1.csv}{\loadedtableFBUtilDp}
    \pgfplotstablegetcolsof{\loadedtableFBUtilDp}
    \pgfmathtruncatemacro{\NoOfColsFBUtilDp}{\pgfplotsretval-1}
    
\pgfplotstableread[col sep=comma]{data/Nl-1_2_Twitch_ES_DE_RareF1.csv}{\loadedtableDEUtilDp}
    \pgfplotstablegetcolsof{\loadedtableDEUtilDp}
    \pgfmathtruncatemacro{\NoOfColsDEUtilDp}{\pgfplotsretval-1}

\pgfplotstableread[col sep=comma]{data/Nl-1_2_Twitch_ES_ENGB_RareF1.csv}{\loadedtableENGBUtilDp}
    \pgfplotstablegetcolsof{\loadedtableENGBUtilDp}
    \pgfmathtruncatemacro{\NoOfColsENGBUtilDp}{\pgfplotsretval-1}

\pgfplotstableread[col sep=comma]{data/Nl-1_2_Twitch_ES_RU_RareF1.csv}{\loadedtableRUUtilDp}
    \pgfplotstablegetcolsof{\loadedtableRUUtilDp}
    \pgfmathtruncatemacro{\NoOfColsRUUtilDp}{\pgfplotsretval-1}

\pgfplotstableread[col sep=comma]{data/Nl-1_2_Twitch_ES_FR_RareF1.csv}{\loadedtableFRUtilDp}
    \pgfplotstablegetcolsof{\loadedtableFRUtilDp}
    \pgfmathtruncatemacro{\NoOfColsFRUtilDp}{\pgfplotsretval-1}

\pgfplotstableread[col sep=comma]{data/Nl-1_2_Twitch_ES_PTBR_RareF1.csv}{\loadedtablePTBRUtilDp}
    \pgfplotstablegetcolsof{\loadedtablePTBRUtilDp}
    \pgfmathtruncatemacro{\NoOfColsPTBRUtilDp}{\pgfplotsretval-1}

\pgfplotstableread[col sep=comma]{data/Nl-1_2_Flickr_F1.csv}{\loadedtableFlickrUtilDp}
    \pgfplotstablegetcolsof{\loadedtableFlickrUtilDp}
    \pgfmathtruncatemacro{\NoOfColsFlickrUtilDp}{\pgfplotsretval-1}

\pgfplotstableread[col sep = comma]{data/utility_attack_data/forfig_cora_baseline_balanced.csv}\loadedtableBaselineCora

\pgfplotstableread[col sep = comma]{data/utility_attack_data/forfig_pubmed_baseline_balanced.csv}\loadedtableBaselinePubMed

\pgfplotstableread[col sep = comma]{data/utility_attack_data/forfig_facebook_page_baseline_balanced.csv}\loadedtableBaselineFbpage

\pgfplotstableread[col sep = comma]{data/utility_attack_data/forfig_citeseer_baseline_balanced.csv}\loadedtableBaselineCiteSeer

\pgfplotstableread[col sep = comma]{data/utility_attack_data/forfig_cora_efficient_balanced.csv}\loadedtableEfficientCora

\pgfplotstableread[col sep = comma]{data/utility_attack_data/forfig_pubmed_efficient_balanced.csv}\loadedtableEfficientPubMed

\pgfplotstableread[col sep = comma]{data/utility_attack_data/forfig_facebook_page_efficient_balanced.csv}\loadedtableEfficientFbpage

\pgfplotstableread[col sep = comma]{data/utility_attack_data/forfig_citeseer_efficient_balanced.csv}\loadedtableEfficientCiteSeer

\pgfplotstableread[col sep = comma]{data/utility_attack_data/forfig_bipartite_baseline_balanced.csv}\loadedtableBaselineBipartite

\pgfplotstableread[col sep = comma]{data/utility_attack_data/forfig_chameleon_baseline_balanced.csv}\loadedtableBaselineChameleon

\pgfplotstableread[col sep = comma]{data/utility_attack_data/forfig_bipartite_efficient_balanced.csv}\loadedtableEfficientBipartite

\pgfplotstableread[col sep = comma]{data/utility_attack_data/forfig_chameleon_efficient_balanced.csv}\loadedtableEfficientChameleon

\usepackage{pgfplots}
\usepackage[ruled,vlined, linesnumbered]{algorithm2e}
\pgfplotsset{compat=1.10}
\usetikzlibrary{pgfplots.groupplots}
\usepgfplotslibrary{fillbetween}

\newcommand{\ash}[1]{{\textcolor{black}{{#1}}}}
\usepackage{xspace}

\newcommand{\fbpage}{\textmd{Facebook}\xspace}
\newcommand{\cora}{\textmd{Cora}\xspace}
\newcommand{\seer}{\textmd{Citeseer}\xspace}
\newcommand{\pubmed}{\textmd{PubMed}\xspace}
\newcommand{\bipartite}{\textmd{Bipartite}\xspace}
\newcommand{\chameleon}{\textmd{Chameleon}\xspace}
\newcommand{\twitch}{\textmd{Twitch}\xspace}

\newcommand{\twitchptbr}{\textmd{Twitch-PTBR}\xspace}
\newcommand{\twitchru}{\textmd{Twitch-RU}\xspace}

\newcommand{\flickr}{\textmd{Flickr}\xspace}
\newcommand{\md}[1]{\textmd{#1}\xspace}

\newcommand{\Gc}{\mathcal{G}}
\newcommand{\Vc}{\mathcal{V}}
\newcommand{\Ec}{\mathcal{E}}

\newcommand{\Ac}{\mathbf{A}}
\newcommand{\emb}{\mathbf{L}}
\newcommand{\model}{\mathbf{M}}
\newcommand{\feat}{\mathbf{F}}
\newcommand{\lab}{\mathbf{Y}}
\newcommand{\dv}{\mathbf{X}}
\newcommand{\concat}{^\frown}

\newcommand{\fc}[1]{\mathbf{#1}} 
\newcommand{\cf}[1]{\mathsf{#1}} 
\newcommand{\arr}[1]{\mathbf{#1}} 
\newcommand{\set}[1]{\mathcal{#1}} 

\newcommand{\sotaAttack}{\textsc{LinkTeller}\xspace}
\newcommand{\baselineAttack}{\textsc{LPA}\xspace}
\newcommand{\baseline}{\textsc{DpGCN}\xspace}
\newcommand{\dpgcn}{\textsc{DpGCN}\xspace}

\newcommand{\tool}{\textsc{LPGNet}\xspace}
\newcommand{\toolbf}{\textsc{\textbf{LPGNet}}\xspace}
\newcommand{\toola}{\textsc{LPGNet-1}\xspace}
\newcommand{\toolb}{\textsc{LPGNet-2}\xspace}
\newcommand{\toolc}{\textsc{LPGNet-3}\xspace}
\newcommand{\gcn}{GCN\xspace}
\newcommand{\mlp}{MLP\xspace}
\newcommand{\mlps}{MLPs\xspace}
\newcommand{\degreevec}{cluster degree vector\xspace}\newcommand{\degreevecs}{cluster degree vectors\xspace}
\newcommand{\homophilyplot}{homophily plot\xspace}

\definecolor{amethyst}{rgb}{0.6, 0.4, 0.8}
\definecolor{azure(colorwheel)}{rgb}{0.0, 0.5, 1.0}
\definecolor{chocolate(traditional)}{rgb}{0.48, 0.25, 0.0}
\definecolor{onyx}{rgb}{0.06, 0.06, 0.06}

\pgfplotscreateplotcyclelist{fycle}{
    {red},
    {blue}, 
    {amethyst},
    {azure(colorwheel)}
    {orange},
    {chocolate(traditional)},
}

\newcommand{\custombox}[2]{\begin{center}\fbox{\parbox{3.1in}{\textit{Result
{#1}: {#2}}}\xspace}\end{center}}

\pgfplotscreateplotcyclelist{exotic}{%
magenta,every mark/.append style={fill=magenta!80!black},mark=*\\%
brown,every mark/.append style={fill=brown!80!black},mark=square*\\%
blue!60!black,every mark/.append style={fill=blue!80!black},mark=otimes*\\%
red!70!white,mark=star\\%
amethyst!80!black,every mark/.append style={fill=amethyst},mark=diamond*\\%
red,densely dashed,every mark/.append style={solid,fill=red!80!black},mark=*\\%
yellow!60!black,densely dashed,
every mark/.append style={solid,fill=yellow!80!black},mark=square*\\%
black,every mark/.append style={solid,fill=gray},mark=otimes*\\%
blue,densely dashed,mark=star,every mark/.append style=solid\\%
red,densely dashed,every mark/.append style={solid,fill=red!80!black},mark=diamond*\\%
}

\pgfplotscreateplotcyclelist{exotic1}{%
magenta,every mark/.append style={fill=magenta!80!black},mark=none,mark size=3pt\\%
black,every mark/.append style={fill=black!80!black},mark=none,mark size=4pt\\%
blue!60!black,every mark/.append style={fill=blue!80!black},mark=none,mark size=3pt\\%
red!70!white,mark=none\\%
amethyst!80!black,every mark/.append style={fill=amethyst},mark=none\\%
red,densely dashed,every mark/.append style={solid,fill=red!80!black},mark=none\\%
yellow!60!black,densely dashed,
every mark/.append style={solid,fill=yellow!80!black},mark=none\\%
black,every mark/.append style={solid,fill=gray},mark=otimes*\\%
blue,densely dashed,mark=star,every mark/.append style=solid\\%
red,densely dashed,every mark/.append style={solid,fill=red!80!black},mark=diamond*\\%
}
\theoremstyle{plain}

\newtheorem{thm}{Theorem}[section]
\theoremstyle{definition}
\newtheorem{definition}[thm]{Definition} 
\theoremstyle{remark}

\makeatletter
\def\endthebibliography{%
  \def\@noitemerr{\@latex@warning{Empty `thebibliography' environment}}%
  \endlist
}
\makeatother



\begin{document}
\date{}

\title{\toolbf: Link Private Graph Networks for Node Classification}

\author{
Aashish Kolluri
}
\email{aashish7@comp.nus.edu.sg}
\affiliation{\institution{National University of Singapore}
\department{School of Computing}
\country{Singapore}
}
\author{
Teodora Baluta
}
\email{teobaluta@comp.nus.edu.sg}
\affiliation{\institution{National University of Singapore}
\department{School of Computing}
\country{Singapore}
}
\author{
Bryan Hooi
}
\email{bhooi@comp.nus.edu.sg}
\affiliation{\institution{National University of Singapore}
\department{School of Computing}
\country{Singapore}
}
\author{
Prateek Saxena
}
\email{prateeks@comp.nus.edu.sg}
\affiliation{\institution{National University of Singapore}
\department{School of Computing}
\country{Singapore}
}
\begin{abstract}
Classification tasks on labeled graph-structured data have many important applications ranging from social recommendation to financial modeling. Deep neural networks are increasingly being used for node classification on graphs, wherein nodes with similar features have to be given the same label. Graph convolutional networks (GCNs) are one such widely studied neural network architecture that perform well on this task. However, powerful link-stealing attacks on GCNs have recently shown that even with black-box access to the trained model, inferring which links (or edges) are present in the training graph is practical. In this paper, we present a new neural network architecture called \tool for training on graphs with privacy-sensitive edges. \tool provides differential privacy (DP) guarantees for edges using a novel design for how graph edge structure is used during training. We empirically show that \tool models often lie in the sweet spot between providing privacy and utility: They can offer better utility than "trivially" private architectures which use no edge information (e.g., vanilla MLPs) and better resilience against existing link-stealing attacks than vanilla GCNs which use the full edge structure. \tool also offers consistently better privacy-utility tradeoffs than \baseline, which is the state-of-the-art mechanism for retrofitting differential privacy into conventional GCNs, in most of our evaluated datasets.
\end{abstract}

\maketitle

\section{Introduction}
\label{sec:intro}

Graph neural networks (GNN) learn node representations from complex graphs similar to how convolutional neural networks do from grid-like images. One of the prominent uses of GNNs is to classify the graph nodes based on their node features~\cite{kipf2016semi,velivckovic2017graph,xu2018powerful}. They are applied to graphs arising in social networks~\cite{fan2019graph,wang2019neural}, computer vision~\cite{sarlin2020superglue,shen2018person}, natural language processing~\cite{bastings2017graph}, and traffic prediction~\cite{zhao2019t}. GNNs are deployed into large-scale recommender systems at Pinterest~\cite{ying2018graph} and Google Maps~\cite{gmaps}. Graph convolutional networks (GCNs) are one of the most successful type of GNNs for node classification and offer close to state-of-the-art performance~\cite{kipf2016semi}. Hence, in this work, we will focus on GCNs for concreteness.




Although node classification can be done purely with the knowledge of node features, the graph edge structure is known to help achieve better accuracy in the classification task. This is why GCNs are popularly used in such tasks. GCNs directly encode, as one of its hidden layers, the adjacency matrix representation of the edges of the given graph. Therefore, they can internally compute features of a node from that of its neighbors. Compared to using vanilla neural networks, such as deep multilayer perceptrons (MLP), they can provide better accuracy in node classification tasks. 
 
In many applications, however, graph edges correspond to sensitive social or financial relationships between people represented as nodes in the graph. Since these are directly used in training GCNs, the privacy risk of leaking which edges are present in the graph is a serious concern. In fact, attacks which can decipher edges present in the original graph used for training given black-box access to the trained GCNs have recently been shown~\cite{he2021stealing,wu2021linkteller}. For instance, the \sotaAttack attack is reported to infer edges from GCNs with high precision without needing any background knowledge.

To defend against such attacks, a principled approach is to use differential privacy (DP) in the GCN training process. The first algorithm to train GCNs with DP guarantees for graph edges, called \baseline, was proposed recently~\cite{wu2021linkteller}. \baseline, much like other DP algorithms, adds noise to hide whether each individual private data element, i.e., an edge, is present. The noise added is based on a privacy budget parameter called $\epsilon$ which determines its privacy-utility tradeoffs.
%
The main issue with \baseline is that it achieves poor privacy-utility tradeoff. 
\baseline models trained with even moderate privacy budgets
($\epsilon \in [1,5]$) perform worse than a standard \mlp by up to $44\%$ in classification accuracy on standard datasets. This implies that MLPs, which only use node features and not the graph edges at all, fare better than \baseline. At privacy budgets ($\epsilon > 8$) \baseline may perform better than an \mlp, but existing attacks achieve similar attack performance on the \baseline models as they do on the non-private GCNs~\cite{wu2021linkteller}. This indicates that \baseline offers little privacy at such $\epsilon$. In short, \baseline offers very few {\em sweet spots} where one can reap the utility benefits of using edges for training while offering some privacy advantage over GCNs.

\paragraph{Our Approach.} 
%
In this work, we present a new neural network architecture called {\bf L}ink {\bf P}rivate {\bf G}raph {\bf Net}works or \tool that offers better privacy-utility tradeoffs. {\em \tool often hits the sweet spots and offers significantly better attack resilience than \baseline while offering similar or better utility}. Our key insight is to move away from the conventional GNNs/GCNs where the raw graph edges (adjacency matrix) are an inseparable part of the neural network architecture. Conventional methods of providing DP such as those used in \baseline drastically change the adjacency matrix and after adding noise can significantly distort the propagation structure inside a GCN. Instead, we propose to separate the graph (edge) structure from the neural network architecture and only query the graph structure when needed. To achieve this, we design a novel architecture using only MLPs to model both the node feature information and some carefully chosen graph structural information. The structure information, which is provided as features to the MLPs, helps neighboring nodes have more similar  representation in the feature space used by MLPs.
 
 What kind of structural information should one use? Observe that GCN-based classification works by exploiting the phenomenon of ``homophily'' which says that neighboring nodes often have similar features, therefore a node's cluster is likely to be the same as that of the majority of its neighbors~\cite{mcpherson2001birds}. This phenomenon has been extensively observed in real-world graphs. In fact, GCNs work well because they use the raw edge structure to internally aggregate cluster labels from a node's neighbors. On the other hand, MLPs which do not use the edge structure at all can easily classify nodes into a cluster different from that of the majority of its neighbors, which does not adhere to the principle behind homophily. 
 
Our insight is thus to compute a special representation on the graph structure called a {\em \degreevec}. For each node, the \degreevec query asks: ``How many neighbors of the node are present in each cluster''. We observe that if the level of homophily in a graph is high, then nodes in the same cluster will have  \degreevecs of high cosine similarity. These vectors capture information about the level of homophily in the graph, and at the same time, use only coarse-grained node degree counts rather than using individual edges. Based on our above observations, our novel \tool architecture stacks layers of MLPs trained on node feature embeddings and splices noisy (differentially private) \degreevecs between successive stacked layers. This improves the classification accuracy iteratively  after each stacked layer because nodes in the same cluster have similar \degreevec representation. Privacy sensitive edges are only queried while computing these \degreevecs.


\paragraph{Evaluation.} We evaluate both the transductive setting, where the inference and training are performed using different nodes of the same graph, and the inductive setting, where the inference graph is different (as in transfer learning). We use $6$ standard benchmark datasets. In the transductive setting,
we find that \tool has {\em consistently better} privacy-utility tradeoffs compared to the state-of-the-art defense, namely \baseline. \tool outperforms  \baseline in classification accuracy by up to $2.6\times$ in $\epsilon \in [1,2]$ and remains superior for all $\epsilon \in (2, 7]$ and across {\em all} datasets evaluated. \baseline offers significantly worse utility compared to a vanilla \mlp (which is trivially private) for all $\epsilon \in [1,7]$, whereas \tool performs better than \mlp at all $\epsilon\geq2$ and close to it for $1\leq \epsilon < 2$ on all datasets.
\ash{
We also evaluate two state-of-the-art link stealing attacks, \baselineAttack~\cite{he2021stealing} and \sotaAttack~\cite{wu2021linkteller}, on our proposed defense.~\footnote{The \baselineAttack paper uses LSA to denote their attack, we call it \baselineAttack.} \tool has better attack resilience than \baseline whenever both of them have a better utility than an \mlp. The attack performance,  measured by the Area Under the Receiver Operating Curve (AUC), is at most $11\%$ higher on \tool compared to \mlp in absolute difference across all datasets and both attacks. In contrast, on \dpgcn the best attack's AUC could go up to $22\%$ higher than \mlp. Further, \sotaAttack regularly achieves an AUC $>0.95$ on \dpgcn which implies that, in such cases, \dpgcn offers no protection to a GNN that leaks edges to \sotaAttack.
}
\ash{In the inductive setting, \tool also exhibits better utility than \baseline for a majority of evaluated configurations. The attack AUC on \tool is always lower than \dpgcn at similar utility and can be up to $34\%$ lower than \dpgcn.}

\ash{Finally, we discuss how to interpret our results and conclude what attacks \tool (or DP in general) protects against. Our discussion contains new insights and pointers to future work which may not be obvious from the prior attacks on graph learning.
}

\paragraph{Contributions.} We propose \tool, a new stacked neural network architecture for learning over graphs. It offers differential privacy guarantees for graph edges used during training. \tool is designed to better retain the signal of homophily present in training graphs without directly using fine-grained edge information. We compare \tool to state-of-the-art DP solution (\baseline). We show that \tool has better resilience against existing link stealing attacks than \baseline models with the same level of utility. 

\subsection*{Availability} \tool is publicly available on GitHub.~\footnote{https://github.com/ashgeek/lpgnet-prototype} We refer to the Appendices several times throughout the text. They can be found in the full version of this paper~\cite{kolluri2022lpgnet}.

\begin{figure*}[t]
    \centering
    \includegraphics[scale=0.13]{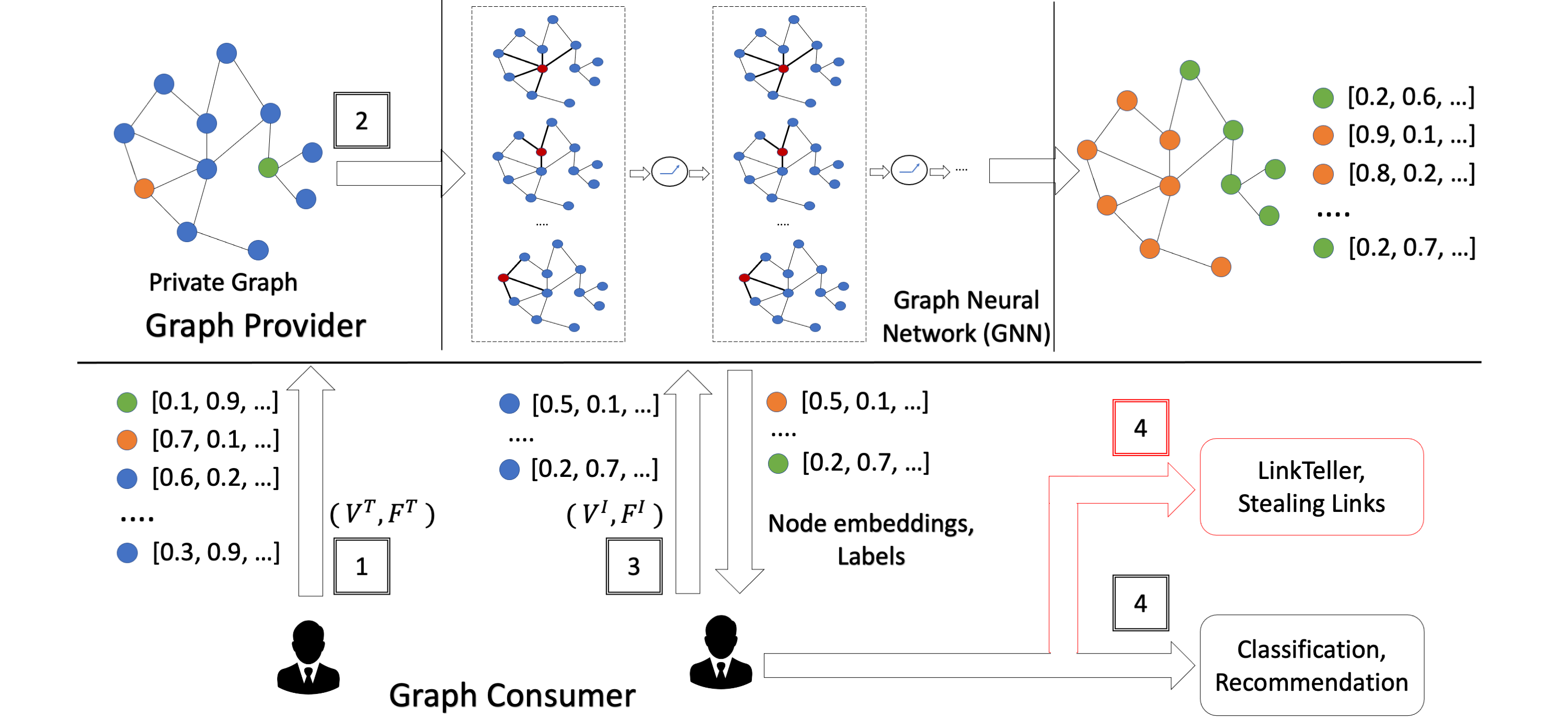}
    \caption{The private graph of users is stored by the graph provider service which is queried by a graph consumer service that has user attributes. In step 1, the graph consumer service sends a set of user features and their labels to the graph provider service. In step 2, the graph provider uses the features and its private graph to train a graph neural network and releases a black-box query access API. The graph consumer queries the API with a new set of node features to infer their node embeddings and their labels. In step 4, the node embeddings may be used for downstream applications such as node classification and recommendations. However, the graph consumer may also attack the GNN using the black-box API to infer the private edges.  }
    \label{fig:setup}
\end{figure*}

\section{Problem}
\label{sec:problem}

GNNs learn low-dimensional representations for nodes (node embeddings) that are used to classify them into meaningful categories. However, to achieve that, GNNs may have to be trained on private data that is stored across multiple entities. We provide the overview of our application setup and the problem in Figure~\ref{fig:setup}.

\subsection{Application Setup \& Goals}
\label{sec:appl-setup}

We focus on the problem of semi-supervised node classification over a graph in which the edges are privacy-sensitive.
To illustrate this, consider a trusted {\em graph provider} (Facebook) which hosts a real-world social networking graph. Its nodes are users and the edges denote their private social links. Consider an untrusted {\em graph consumer}, for instance a music streaming service (Spotify), that holds features of the users such as their music preferences, but not their social links. It may want to use the graph structure along with the user features to improve its recommendations. The graph structure is important since it encodes the principle of ``homophily'', i.e., the neighboring nodes on the graph tend to have similar interests.

The aforementioned example is studied as an instance of a more general problem called the {\em node classification} problem. The goal is to classify nodes that are similar to each other into clusters with distinct {\em labels}, based on their features and the edge structure. In the semi-supervised setup, the cluster labels of a small set of nodes are known and the task is to predict the labels of the remaining nodes. Specifically, a graph $\Gc^{T}:(\Vc^{T},\Ec^{T})$ is available to the graph provider and the features $\feat^{T}$ along with the labels $\lab$ for a small set of nodes in $\Vc^{T}$ are available to the graph consumer.

In this well-studied application setup, the graph consumer gives the features $\feat^T$ and labels $\lab$ to the graph provider. The provider then uses them along with the graph $\Gc^T$ to train an ML model and allows the consumer to query the model hosted on its servers. The consumer can then query the ML model with a set of {\em inference nodes} $\Vc^{I}$ along with their features $\feat^{I}$ to get their low-dimensional embeddings that are output by the ML model~\cite{wu2021linkteller,he2021stealing}. 
%

The learnt model can be queried for labels of  inference nodes, which were not part of the training set. There are two settings in the literature for the classification task. In the {\em transductive} setting, the training and inference are performed on the same graph. The goal is to learn a classifier using the entire graph structure and a few labeled nodes to predict labels of other nodes in it. In the {\em inductive} setting, the graphs used for training and inference are different. The classifier is evaluated on how transferable it is to unseen graphs (or different versions of a dynamically changing graph).

\paragraph{Background: Graph neural networks.} Graph neural networks (GNNs) achieve state-of-the-art performance for the node classification problem. Several GNN architectures have been proposed recently, however, we describe one widely-used architecture called the graph convolutional neural network (\gcn) for concreteness~\cite{kipf2016semi}. It takes a matrix as input with each row corresponding to the features (vector) of a user or node in $\Vc$. The logits obtained from the last layer of the \gcn are the node embeddings which can be converted to probability scores per label using a softmax layer. A \gcn directly splices the graph's edge structure in its internal architecture itself. In every hidden layer, for each node in the graph the \gcn aggregates the features obtained from all its neighbors. Therefore, when $k$ hidden layers are stacked, the features are obtained from all $k$-hop neighbors for aggregation. Formally, every hidden layer $\mathbf{H}^k$ in a \gcn is represented as follows:
$$
\arr{H}^{k} = \sigma(\tilde{\Ac}\cdot \arr{H}^{k-1}\cdot \arr{W})
$$
$\tilde{\Ac}$ is a normalized adjacency matrix, $\arr{H}^0$ is the input node features, $\arr{W}$ is the weight parameters and $\sigma$ is a non-linear activation function.

\paragraph{Privacy Goals.}
Revealing the graph edges to the untrusted data consumer is a serious privacy issue. Many prominent social networking companies protect against querying the direct social links of the user.
\ash{Our goal is to ensure that the untrusted data consumer does not learn about the presence or absence of edges in the data provider's graph by querying the ML model that uses the graph for training or inference. We assume that the untrusted consumer only has black-box access to the trained model.} Even in this restricted setup, recent works~\cite{wu2021linkteller,he2021stealing} have shown that it is possible to identify if an edge was used in training with high (nearing $100\%$) precision in some cases. \ash{We aim to train GNNs that are less likely to reveal that they are using an edge for training or inference using the differential privacy (DP) framework~\cite{privacybook}. By definition, such GNNs are {\em resilient} to link-stealing attacks that infer whether specific edges were used by the GNN during training or inference.}


A differentially private query will ensure that the adversary can get only a bounded amount of additional information about the input dataset after seeing the query outputs.

\begin{definition}[Differential Privacy]
\label{def:dp}
Consider any two datasets, $\set{D}_1, \set{D}_2 \in \set{D}$, such that ($|\set{D}_1-\set{D}_2|\leq1$). A randomized query $\fc{M}:\set{D}\rightarrow{} \set{S}$ satisfies $\epsilon$-DP if, for all such $\set{D}_1, \set{D}_2 \in \set{D}$ and for all $\set{S}_0\subset\fc{M}(\set{D})$, 
\begin{align*}
Pr(\fc{M}(\set{D}_1)\in \set{S}_0)\leq e^\epsilon \cdot Pr(\fc{M}(\set{D}_2) \in \set{S}_0)
\end{align*}
\end{definition}

For graphs, there are $2$ notions of DP: edge-DP~\cite{hay2009accurate} and node-DP~\cite{kasiviswanathan2013analyzing}. Edge-DP is used to protect the existence of any individual edge in the graph. Node-DP is used to protect the existence of any node (and its edges). \baseline is designed with edge-DP as its objective. In our work, we also use the edge-DP framework since our goal is to protect individual edges from being leaked by a GNN. Achieving node-DP is promising future work and usually has worse utility trade-offs than edge-DP in other setups.

\ash{We consider undirected graphs with unweighted edges in this work. If $\Vc$ is given as an ordered set of nodes then every undirected graph $\Gc$ that can be constructed with $\Vc$ is captured by an array of $1$s and $0$s, i.e., $\set{D}_\Gc = \{I(v_i, v_j)| i,j\in\{1,2,\ldots,|\Vc|\} \text{ and } i<j\}$. Here, $I(v_i, v_j)=1$ if the nodes $v_i$ and $v_j$ have an edge in $\Gc$ otherwise $0$. Therefore, to define an edge-DP mechanism, we just view $\set{D}$ in definition~\ref{def:dp} as a set of all possible $\set{D}_\Gc$ and $\set{D}_1, \set{D}_2$  as two datasets (graphs) with a difference of one edge (addition or removal).}

\begin{algorithm}[t]
\SetAlgoLined
\SetKwInOut{Input}{Input}
\SetKwInOut{Output}{Output}
\Input{Graph $\Gc: (\Vc,\Ec)$ as adjacency matrix $\Ac$, Total privacy budget: $\epsilon$, Privacy budget for estimating the number of edges: $\epsilon_r=0.01$ (default).}
\Output{$\hat{\Ac}$}
 $E = |\Ec|$\;
 $\tilde{E} = E + \cf{Lap}(0, \frac{1}{\epsilon_r})$\;
 $\tilde{E} = \lfloor\tilde{E}\rfloor$\;
 $\Ac_{tr}$ = $\cf{UppTriMatrix}(\Ac, 1)$\;
 \tcp{Upper triangular matrix on $1^{st}$ superdiagonal}
 \For{entry in $\Ac_{tr}$}{
    $\Ac_{tr}[entry] += \cf{Lap}(0, \frac{1}{\epsilon-\epsilon_r})$\;
 }
 \tcp{neglecting entries of the lower triangle}
 $top\_indices = arg\,max(\Ac_{tr}, \tilde{E})$\;
 \tcp{get the indices of top $\tilde{E}$ values}
 $\Ac_{tr}[top\_indices] = 1$, $\Ac_{tr}[\neg top\_indices] = 0$\;
 $\hat{\Ac} = \Ac_{tr} + \Ac^{T}_{tr}$\;
 \Return $\hat{\Ac}$
 \caption{An outline of \baseline~\cite{wu2021linkteller}. It adds Laplace noise to all entries of the adjacency matrix and selects the top-$\tilde{E}$ values to maintain the original density of the graph.}
 \label{alg:lapgraph}
\end{algorithm}
\subsection{Existing Solution}
Recently, Wu et al.~\cite{wu2021linkteller} have proposed an edge-DP algorithm to learn a \gcn, called \baseline. In order to achieve edge-DP, \baseline proposes to use the Laplace mechanism on the adjacency matrix before training the \gcn. Here, we define the Laplace mechanism:

\begin{definition}[Laplace Mechanism]
Let the sensitivity of the query $f:\set{D}\xrightarrow{}\set{S}$ be
  $\Delta(\fc{f}) = \max{|\fc{f}(\set{D}_1)-\fc{f}(\set{D}_2)|}\text{ }$ for all $\set{D}_1, \set{D}_2 \in \set{D}$ such that $|\set{D}_1-\set{D}_2|\leq 1$. The mechanism $\fc{M}:\set{D}\xrightarrow{}\set{S}$ defined as follows satisfies $\epsilon$-DP:
\begin{align*}
    &\fc{M}(\set{D}_1)=\fc{f}(\set{D}_1)+\cf{Lap}(0, \frac{\Delta(\fc{f})}{\epsilon})
\end{align*}
\end{definition}

Once the adjacency matrix is differentially private the rest of the training process is also differentially private as given by the {\em post-processing} property of DP mechanisms~\cite{privacybook}. Formally,

\begin{definition}[Post-processing property]
If a mechanism $\fc{M}$ satisfies $\epsilon$-DP then any function $\fc{g} \circ \fc{M}$ also satisfies $\epsilon$-DP.
\end{definition}

Algorithm~\ref{alg:lapgraph} summarizes the \baseline solution. For undirected graphs, the adjacency matrix can be represented by an upper triangular matrix. Therefore, if we add or remove an edge from the graph then only one entry in the adjacency matrix will change. So, adding noise sampled from $\cf{Lap}(0, \frac{1}{\epsilon})$ to each entry in adjacency matrix results in a differentially private one. However, now the adjacency matrix becomes a weighted matrix with a non-zero weight associated with almost every entry (edge). Hence, the noised adjacency matrix represents a very dense and weighted graph. In order to maintain the original graph statistics such as the exact number of edges ($E = |\Ec|$) and density, they propose to take the top-$E$ entries in the DP-adjacency matrix to be the actual edges (replace the top-$E$ with 1 and rest 0). A small privacy budget $\epsilon_r < \epsilon$ is used to compute the number of edges ($\tilde{E} = E + \cf{Lap}(0, \frac{1}{\epsilon_r})$). The remaining privacy budget $\epsilon - \epsilon_r$ is used to noise the adjacency matrix.

\subsection{Attacks \& Utility Tradeoffs}
DP mechanisms trade off utility for privacy by adding noise. In DP, $\epsilon$ measures the level of privacy---lower $\epsilon$ means higher privacy. The utility is expected to decrease with decreasing $\epsilon$.  There is no consensus on how to choose $\epsilon$ in practice, however, $\epsilon$ values ranging from $0.1$ to $10$ have been used consistently to evaluate newly proposed mechanisms~\cite{abadi2016deep,chaudhuri2011differentially,lu2014exponential,xiao2014differentially,wu2021linkteller}. For our problem, existing approaches offer a glaringly poor utility-privacy tradeoff.

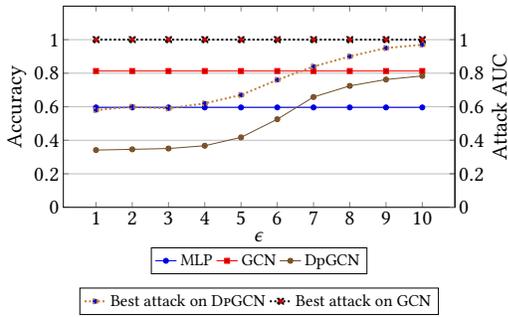
\begin{figure}
\centering
\resizebox{0.80\linewidth}{!}{%
    \begin{tikzpicture}
        \begin{axis}[
            yminorticks=false,
            width=1.5*\axisdefaultwidth,
            height=\axisdefaultheight,
            ylabel={Accuracy},
            xlabel={$\epsilon$},
            y label style={font=\Huge},
            x label style={font=\Huge},
            ytick={0.0, 0.2, 0.4, 0.6, 0.8, 1.0},
            ymin=0.0,
            ymax=1.2,
            y tick label style={font=\Huge},
            xtick=data,
            xticklabels={1,2,3,4,5,6,7,8,9,10},
            x tick label style={font=\Huge},
            legend style={nodes={scale=1.5, transform shape},at={(0.5,-0.2)}, anchor=north, legend columns=4},
            ymajorgrids,
            log origin=infty,
        ]
        \foreach \i in {1,3,9}{
            \addplot+[sharp plot] table[
                    x expr=\coordindex,
                    y index=\i,
                    col sep=comma,
                ] {\loadedtableMotivationUpdated};
                \pgfplotstablegetcolumnnamebyindex{\i}\of{\loadedtableMotivationUpdated}\to{\colname}
        }
        \addlegendentry{MLP}
        \addlegendentry{GCN}
        \addlegendentry{DpGCN}
        \end{axis}
        \begin{axis}[
            yminorticks=false,
            width=1.5*\axisdefaultwidth,
            height=\axisdefaultheight,
            ylabel={Attack AUC},
            y label style={font=\Huge},
            ytick={0.0, 0.2, 0.4, 0.6, 0.8, 1.0},
            ymin=0.0,
            ymax=1.2,
            y tick label style={font=\Huge},
            xtick=data,
            legend style={nodes={scale=1.5, transform shape},at={(0.5,-0.4)}, anchor=north, legend columns=4},
            log origin=infty,
            axis x line = none,
            axis y line*=right,
        ]
        \foreach \i in {15}{
            \addplot+[brown, ultra thick,dotted, axis y line*=right,ylabel=Attack AUC] table[
                    x expr=\coordindex,
                    y index=\i,
                    col sep=comma,
                ] {\loadedtableMotivationUpdated};
                \pgfplotstablegetcolumnnamebyindex{\i}\of{\loadedtableMotivationUpdated}\to{\colname}
        }
        \foreach \i in {16}{
            \addplot+[black, ultra thick,dotted, axis y line*=right,ylabel=Attack AUC] table[
                    x expr=\coordindex,
                    y index=\i,
                    col sep=comma,
                ] {\loadedtableMotivationUpdated};
                \pgfplotstablegetcolumnnamebyindex{\i}\of{\loadedtableMotivationUpdated}\to{\colname}
        }
        \addlegendentry{Best attack on \baseline}
        \addlegendentry{Best attack on \gcn}
        \end{axis}
    \end{tikzpicture}
    }
\caption{The privacy-utility tradeoffs offered by \baseline on Cora dataset. It does not perform better than an MLP until $\epsilon=7$ making it unusable at such $\epsilon$. At higher $\epsilon\geq 9$ the attack performance on \dpgcn is as high as its performance on non-DP \gcn making it unusable there as well.}
\label{fig:privacy-utility-baseline}
\end{figure}

To illustrate the poor tradeoff, we show the node classification performance of \baseline on a commonly used dataset called Cora at $\epsilon \in [1, 10]$, following the evaluation in prior work~\cite{wu2021linkteller}, in  Figure~\ref{fig:privacy-utility-baseline}. Strikingly, the performance is worse than a two-layer fully connected feed-forward neural network (MLP) for $\epsilon < 7$. An \mlp trivially offers DP since it does not use the graph edges at all. This means that if one desires $\epsilon < 7$, then using an MLP would be better than models produced by \baseline. On the other hand, when the $\epsilon \geq 8$, \baseline performs better than the \mlp and closer to the \gcn. However, at that $\epsilon$, \sotaAttack can predict the edges in the original graph with close to $100\%$ precision and an AUC score $\geq0.9$. The attack performance on \baseline is very close to its performance on the non-DP \gcn. This shows that at $\epsilon \geq 8$ the model offers as much privacy as a non-private \gcn. We observe similar trends for other datasets as well, confirming the findings in~\cite{wu2021linkteller}.

There are few $\epsilon$ values for which the attack accuracy is weaker on \dpgcn compared to a non-DP \gcn and utility is better than that of \mlps. For example, see $\epsilon \in [7,8)$ for the \cora dataset in Figure~\ref{fig:privacy-utility-baseline}. We call such values sweet spots. Intuitively, when the number of sweet spots are high, the private algorithm is more useful as a defense against existing attacks since its utility is better than \mlp's which does not use edge structure. Specifically, a defense is objectively better than another defense in a particular setting if the attack performance on that defense is worse, while the utility is better than the other defense. Figure~\ref{fig:privacy-utility-baseline} shows how \baseline fares worse than \mlp for $\epsilon<7$, offering no sweet spots in that area. 
Stated succinctly, our main challenge and central contribution is to show a new way of designing neural network models that often hit sweet spots between privacy and utility. The performance desired should be better than \mlp, which does not use edges, and \baseline which does. Further, it should offer lower (better) $\epsilon$ values, and be as resistant as an \mlp to state-of-the-art attacks (\sotaAttack).
\begin{table}[b]
    \centering
    \caption{Percentage of Noisy edges sampled by \baseline for the Cora dataset with varying $\epsilon \in [1,10]$.}
    \resizebox{\columnwidth}{!}{
    \begin{tabular}{c c c c c c c c c c c}
    \toprule
        $\epsilon$ & 1 & 2 & 3 & 4 & 5 & 6 & 7 & 8 & 9 & 10\\ 
        Noise & 100 & 99 & 98 & 93 & 84 & 66 & 42 & 25 & 15 & 9 \\\hline
    \end{tabular}
    }
    \label{tab:noise}
\end{table}
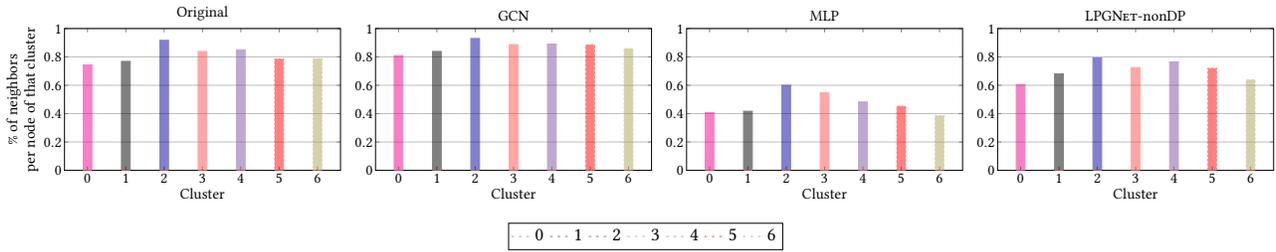
\begin{figure*}[t]
\centering
\resizebox{0.95\linewidth}{!}{%
    \begin{tikzpicture}
        \begin{groupplot}[group style = {group size = 4 by 1, horizontal sep = 40pt},
            width=1/2.5*\textwidth,
            height = 5cm]
        \nextgroupplot[title={Original},
                    title style={font=\Huge},
            yminorticks=false,
            width=1.5*\axisdefaultwidth,
            height=\axisdefaultheight,
            ylabel style={align=center},
            ylabel={\% of neighbors \\ per node of that cluster},
            xlabel={Cluster},
            y label style={font=\Huge},
            x label style={font=\Huge},
            ymax=1.0,
            ymin=0.0,
            ytick={0.0, 0.2, 0.4, 0.6, 0.8, 1.0},
            y tick label style={font=\Huge},
            xtick=data,
            xticklabels={0,1,2,3,4,5,6},
            x tick label style={font=\Huge},
            legend style={column sep = 2pt, legend columns = -1, legend to name = grouplegend, nodes={scale=2.4, transform shape},},
            ymajorgrids,
            log origin=infty,
            cycle list name = exotic1,
        ]
        \foreach \i in {1,...,7}{
            \addplot+[ybar, fill, opacity=0.5] table[
                    x expr=\coordindex,
                    y index=\i,
                    col sep=comma,
                ] {\loadedtableHomoOri};
                \pgfplotstablegetcolumnnamebyindex{\i}\of{\loadedtableHomoOri}\to{\colname}
                \addlegendentryexpanded{\colname};
        }
        ]
        
        \nextgroupplot[title={GCN},
                    title style={font=\Huge},
            yminorticks=false,
            width=1.5*\axisdefaultwidth,
            height=\axisdefaultheight,
            ylabel style={align=center},
            xlabel={Cluster},
            y label style={font=\Huge},
            x label style={font=\Huge},
            ytick={0.0, 0.2, 0.4, 0.6, 0.8, 1.0},
            y tick label style={font=\Huge},
            ymax=1.0,
            ymin=0.0,
            xtick=data,
            xticklabels={0,1,2,3,4,5,6},
            x tick label style={font=\Huge},
            legend style={nodes={scale=2.0, transform shape},at={(0.5,-0.3)}, anchor=north, legend columns=3},
            ymajorgrids,
            log origin=infty,
            cycle list name = exotic1,
        ]
        \foreach \i in {1,...,7}{
            \addplot+[ybar, fill, opacity=0.5] table[
                    x expr=\coordindex,
                    y index=\i,
                    col sep=comma,
                ] {\loadedtableHomoGCN};
                \pgfplotstablegetcolumnnamebyindex{\i}\of{\loadedtableHomoGCN}\to{\colname}
        }
        ]
        
        \nextgroupplot[title={MLP},
                    title style={font=\Huge},
            yminorticks=false,
            width=1.5*\axisdefaultwidth,
            height=\axisdefaultheight,
            ylabel style={align=center},
            xlabel={Cluster},
            y label style={font=\Huge},
            x label style={font=\Huge},
            ytick={0.0, 0.2, 0.4, 0.6, 0.8, 1.0},
            y tick label style={font=\Huge},
            ymax=1.0,
            ymin=0.0,
            xtick=data,
            xticklabels={0,1,2,3,4,5,6},
            x tick label style={font=\Huge},
            legend style={nodes={scale=1.5, transform shape},at={(0.5,-0.3)}, anchor=north, legend columns=3},
            ymajorgrids,
            log origin=infty,
            cycle list name = exotic1,
        ]
        \foreach \i in {1,...,7}{
            \addplot+[ybar, fill, opacity=0.5] table[
                    x expr=\coordindex,
                    y index=\i,
                    col sep=comma,
                ] {\loadedtableHomoMLP};
                \pgfplotstablegetcolumnnamebyindex{\i}\of{\loadedtableHomoMLP}\to{\colname}
        }
        ]
        
        \nextgroupplot[title={\tool-nonDP},
                    title style={font=\Huge},
            yminorticks=false,
            width=1.5*\axisdefaultwidth,
            height=\axisdefaultheight,
            ylabel style={align=center},
            xlabel={Cluster},
            y label style={font=\Huge},
            x label style={font=\Huge},
            ytick={0.0, 0.2, 0.4, 0.6, 0.8, 1.0},
            y tick label style={font=\Huge},
            ymax=1.0,
            ymin=0.0,
            xtick=data,
            xticklabels={0,1,2,3,4,5,6},
            x tick label style={font=\Huge},
            legend style={nodes={scale=1.5, transform shape},at={(0.5,-0.3)}, anchor=north, legend columns=3},
            ymajorgrids,
            log origin=infty,
            cycle list name = exotic1,
        ]
        \foreach \i in {1,...,7}{
            \addplot+[ybar, fill, opacity=0.5] table[
                    x expr=\coordindex,
                    y index=\i,
                    col sep=comma,
                ] {\loadedtableHomoMMLP};
                \pgfplotstablegetcolumnnamebyindex{\i}\of{\loadedtableHomoMMLP}\to{\colname}
        }
        ]
        \end{groupplot}
        \node at ($(group c2r1) + (5.3cm,-5.5cm)$) {\ref{grouplegend}};
    \end{tikzpicture}
    }
\caption{Homophily plots for the Cora dataset. The leftmost plot is for the original graph, the next two for a non-private GCN and MLP respectively, and the rightmost for our \tool without any noise added.} 
\label{fig:homophily_using_dv}
\end{figure*}

\section{Approach}
\label{sec:approach}


Let us see why the \baseline approach offers a poor privacy-utility tradeoff, which motivates our new approach.

\paragraph{Top-down approach does not work.} \baseline takes a top-down approach wherein it retrofits DP to an existing GNN architecture trained for node classification. All GNN architectures use the adjacency matrix in some form or another. Adding noise to the adjacency matrix and using it in the GNN provides DP, but notice that the adjacency matrices have very high sensitivity. Even when a small noise sampled from $Lap(0, \frac{1}{\epsilon})$ is added to each entry of the matrix, it can create $O(|V|^2)$ weighted edges which may not exist in the original graph. We observe that when the \baseline selects top-$E$ edges among them, it cannot differentiate between the original and "noisy" edges at $\epsilon$ such as say $\epsilon=2$. Thus, it selects a lot of noisy edges and the number of such edges gradually increases as $\epsilon$ reduces. To illustrate, $99\%$ of the selected top-$E$ edges are noisy at $\epsilon=2$ for a standard Cora dataset, as shown in the Table~\ref{tab:noise}.
This explains why the \baseline has poor utility-privacy tradeoffs. At moderately low $\epsilon$ values, the graph structure is significantly different from the original which leads to loss of utility. At high $\epsilon$ values, the noise added is small and most of the selected edges are original graph edges, but then at this level, the link-stealing attacks work as effectively on \baseline as on non-private GCNs.

\paragraph{Our approach.} 
We propose \tool, an ML model architecture that can give good utility while avoiding the direct use of the edge structure. \tool starts with an \mlp trained only over node attributes since it is the basic component of most of the GNN architectures. Next, and unlike typical GNNs, \tool computes a {\em \degreevec} for each node that captures coarse-grained graph structure information. To compute \degreevec, nodes are first partitioned into clusters obtained from the labels predicted by the previous \mlp stack layer. For every node \tool computes the number of neighbors it has in each of those clusters. The \degreevec for each node is thus an array of those counts, indexed by the cluster label. The \degreevec is then used with node attributes to train the next \mlp stack layer. The full architectural details of \tool are given in Section~\ref{sec:arch}.

Let us see why our approach works. The idea behind GNNs is that a node tends to have the same label as the majority of its neighbors, a property called homophily which is observed in many real-world networks. But, training an \mlp without the edge structure can often result in the following scenarios which do not adhere to homophily: 1) A node receives a wrong label which is different from the correct label received by a majority of the node's neighbors, 2) A node receives a wrong label and its neighbors receive correct labels but there is no majority label, and 3) A node receives wrong label and its neighbors also receive wrong labels. GNNs suppress these scenarios as the model combines the features of a node's neighbors using the adjacency matrix. This local aggregation pushes two nodes that have many common neighbors towards having similar aggregated features, and eventually, the same cluster labels at each layer. 

In \tool, we indirectly achieve the same goal without using the raw or noisy edge structure directly. We can empirically observe how homophily structure of the graph is preserved in our architecture compared to the alternatives. Specifically, let the fraction of a node's neighbors that reside in the same cluster as itself be its homophily score and $AvgHomophily(i)$ be the average of the homophily scores of all nodes in cluster $i$. If we plot $AvgHomophily(\cdot)$ for all clusters, we obtain what we call a {\em \homophilyplot}. 
Figure~\ref{fig:homophily_using_dv} shows  homophily plots for the \cora dataset. The leftmost plot is for the ground truth clusters in the original graph and the phenomenon of homophily is clearly evident. For instance, nodes of cluster $0$ have about $80\%$ of their neighbors in the same cluster, nodes in cluster $2$ have about $90\%$ of nodes in the same cluster, and so on. 
If we take the clusters produced by non-private GCN (the second plot in Figure~\ref{fig:homophily_using_dv}), the $AvgHomophily(\cdot)$ seems to be very similar to the ones produced with the ground truth clusters. Notice that with clusters obtained by the \mlp, however, the $AvgHomophily(\cdot)$ is significantly lesser than ground truth. This shows how \mlp preserves much weaker
homophily in its output than in the ground truth. Our key idea in \tool is to sharpen the homophily signal, if it exists in the graph used for training, in each stack layer. Observe that homophily implies that two nodes that belong to the same cluster will have similar \degreevecs\footnote{Specifically, we mean cosine similarity here, which ignores the scaling factor due to differing degrees of the two nodes being compared.} as a majority of their neighbors will also belong to the same cluster. By using \degreevecs along with the node features to train the next \mlp stack layer, our architecture is giving similar \degreevecs to nodes in the same cluster, thus increasing the signal-to-noise ratio~\cite{ma2021aunified}. The last homophily plot shows that the \tool architecture preserves the homophily characteristics closer to how GCNs do, at least when no noise is added during training.

Achieving edge-DP is simple in \tool. We add noise to the cluster degree vectors as they are the only computation being made on the graph. If we change one edge in the graph then the degree vectors of the two involved nodes will have changed by $1$ each. Therefore, a noise sampled from $Lap(0,\frac{2}{\epsilon})$ can be added to each count of degree vectors of all nodes. The noise added to degree vectors will have much less impact on them unlike the noise added to the adjacency matrix as they are more coarse-grained than the exact edge information. Specifically, each array count will change by more than $\frac{2}{\epsilon}$ only about $25\%$ of the time and $\frac{4}{\epsilon}$ only $5\%$ of the time.~\footnote{This can be computed using the Hoeffding Inequality on the Laplace random variables corresponding to the noise.} At $\epsilon=1$, the most significant array count that may represent the node's actual cluster can be much higher than the noise added depending on the node's degree. In contrast, the same amount of noise added to an adjacency matrix completely changes the graph since the actual values ($1$s and $0$s) fall within the standard deviation of the noise itself.

\begin{figure*}[t]    
    \centering
    \includegraphics[scale=0.16]{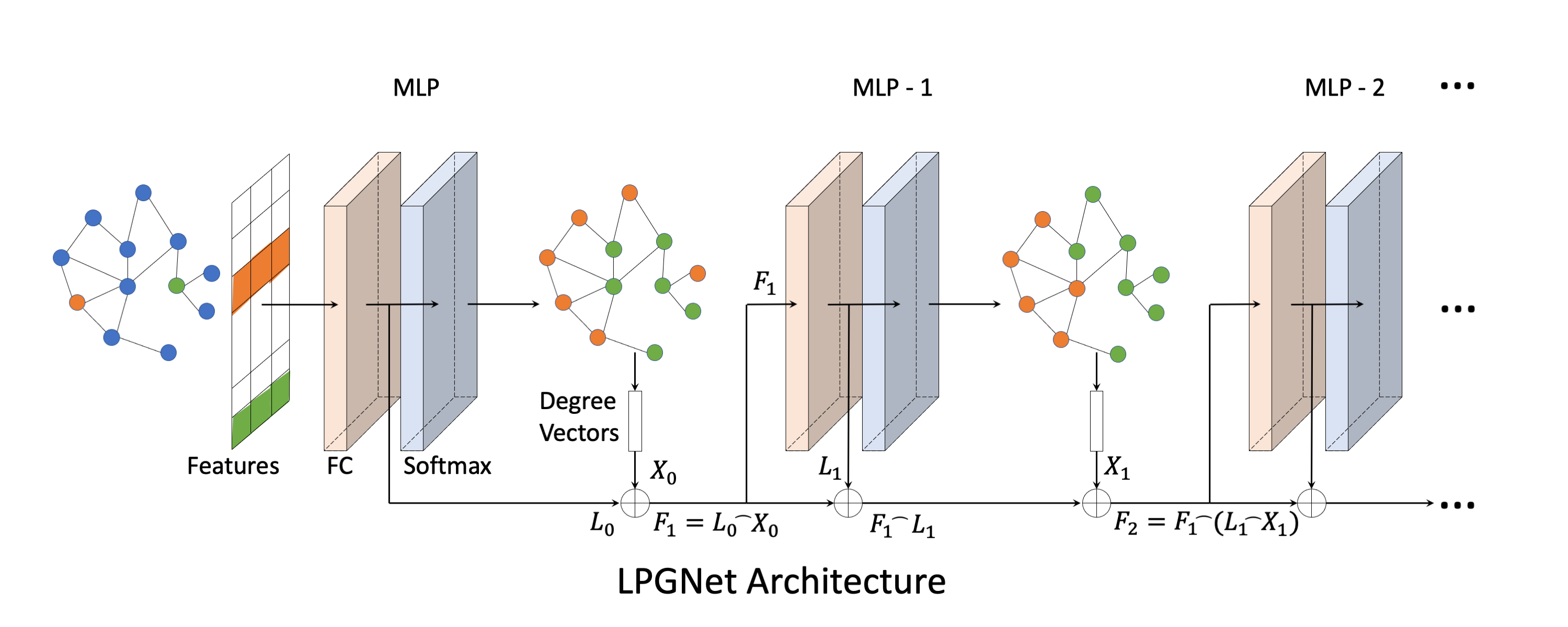}
    \caption{\tool trains MLPs iteratively with the node features and additional graph features encoded in the form of degree vectors. The features provided to every MLP is a concatenation of the features obtained from from all the previous MLPs.}
    \label{fig:lpgnet}
\end{figure*}


\section{Architectural Details of \toolbf}
\label{sec:arch}

\tool represents and implements a chain of MLPs  that  are sequentially trained (one after another independently) in an iterative fashion while taking the help of degree vectors in each iteration. Figure~\ref{fig:lpgnet} captures \tool's architecture and the training process. 

\paragraph{Multi-\mlp architecture} \tool first trains an \mlp on only the node features to obtain the initial set of clusters. Specifically, the input is the feature matrix ($\feat^T$ of size $N\times F$) and the output is a node embedding ($\emb$ of size $N\times C$), where $N=|\Vc^T|$, $F$ is the input feature dimension and $C$ is the number of labels (clusters). Node embeddings are the logit scores output by the \mlp, therefore, we can apply the softmax function over the node embeddings to get the labels for each node. Using these labels \tool obtains the first clusters without any edge information. \tool then begins the training of additional MLPs iteratively.

In each iteration (stack layer) $i$, \tool trains an \mlp based on the degree vectors computed from the clusters obtained in the previous iteration. The degree vectors can be represented by an $N\times C$ matrix ($\dv$) where each row is an array of counts for each node. \tool concatenates the degree vector matrix obtained in the previous iteration $\dv_{i-1}$ to the node embedding matrix from the previous iteration $\emb_{i-1}$, thus, creating $\feat_i = \emb_{i-1}\concat\dv_{i-1}$ an $N\times2C$ matrix that comprises of both node features and the edge information.~\footnote{$\concat$ denotes concatenation.}

\paragraph{Entire history for training} We observe that \tool's performance improves by using the features obtained from all the previous iterations for the current iteration. Specifically, the features in the iteration $i+1$, are the concatenation of $\dv_{i}$, $\emb_{i}$, and $\feat_i$ where $\feat_i$ itself is computed recursively until $i=1$, i.e., $\feat_{i+1} = \feat_{i}\concat(\emb_i\concat\dv_{i})$. 

\paragraph{Transductive vs inductive setting.} In the transductive setting, \tool is trained and tested on the same graph $\Gc^T:(\Vc^T, \Ec^T)$. The features of all nodes are given to the graph provider to train \tool. During inference time, a set of nodes $\Vc^I$ are given to the provider. Note that the provider already has their features $\feat^I$. \tool outputs the logits that are produced by the last \mlp in the training. All the intermediate logits are hidden from the graph consumer.
In the inductive setting, \tool is trained and tested on different graphs. The training procedure is similar to the transductive setting. For inference, the nodes $\Vc^I$ and their features $\feat^I$ from the new graph are provided to the graph provider. It is assumed that provider already has the new graph $\Gc^I:(\Vc^I, \Ec^I)$ which is usually the case with evolving graphs or when the provider has graphs of disjoint set of users from sometimes different countries~\cite{musae}. \ash{If the training and inference graphs are different (call it the {\em inductive-different} setting) then we use only the graph $\Gc^I$ to compute cluster degree vectors for \tool during inference time. In contrast, if the graph is evolving (call it the {\em inductive-evolving} setting) where the edges used for training are a subset of the edges used for inference then the entire graph $\Gc^I$ (that includes $\Gc^T$) is used to compute the cluster degree vectors. To summarize the inference procedure in the inductive setting, for each additional \mlp, \tool computes the cluster degree vectors on $\Gc^I$ and uses them along with the attributes of $\Vc^I$ to perform the forward pass on the \mlp. Finally, the last \mlp's logits are revealed to the graph consumer.} 

Algorithms~\ref{alg:lpgnet} and \ref{alg:lpgnet_inf} summarize the training and inference algorithms for \tool. During the training phase, the best intermediate MLPs are chosen based on the validation loss on the validation dataset whenever it is available. In the transductive setting, the nodes used for validation are from the same graph as the nodes used for training. Therefore, the cluster degree vectors are computed for all nodes once before training every additional \mlp in LPGNet and are reused for validation. In the inductive setting, the nodes used for validation may come from a different graph or from the unseen part of an evolving graph. So, we compute the cluster degree vectors again for validation (Lines $17$-$26$ in Algorithm~\ref{alg:lpgnet}).
\begin{algorithm}[t!]
\SetAlgoLined
\SetKwInOut{Input}{Input}
\SetKwInOut{Output}{Output}
\Input{Train graph $\Gc^{T}:(\Vc^{T},\Ec^{T})$, Train Features and labels: $(\feat^T, \set{Y}^T)$, Validation graph $\Gc^{V}:(\Vc^{V},\Ec^{V})$ (for inductive), Validation Features and labels: $(\feat^V, \set{Y}^V)$, Privacy budget: $\epsilon$, \tool size (\# of additional MLPs): $nl>=1$, Hyperparameters: learning rate ($lr$), dropout rate ($dr$), hidden layer size ($hid_s$), \# hidden layers ($hid_n$), \# of epochs ($e$)}
\Output{Trained \tool $\model$}

 $\model[0] = \cf{trainMLP}(\feat^T, \set{Y}^T, \feat^{V}, \set{Y}^V, hid_n, hid_s, lr, dr)$\;
 \tcp{select the best model across $e$ epochs using Adam optimizer and cross-entropy loss.}
 $i = 0$, $\feat_{0} = \feat^T, {\feat_{0}}^{V} = \feat^{V}$\;
 \While{$i < nl$}{
    $\emb_{i} = \model[i].\cf{forward}(\feat_i)$\;
    \tcp{get logits from one forward pass}
    $labels = \cf{softmax}(\emb_{i})$\;
    $\dv_i =$ $\cf{findDegreeVec}(\Gc^T,labels, \frac{\epsilon}{nl}$)\;
    \If{transductive setting}{
        Store to disk $\dv_i$\;
    }
    \If{i is 0}{
        $\feat_{i+1} = \emb_{i}\concat \dv_{i}$\; \tcp{$\concat$ denotes concatenation}
    }
    \Else{
        $\feat_{i+1} = \feat_{i}\concat(\emb_{i}\concat\dv_{i})$\;
    }
    \tcp{concatenate logits and degree vectors}
    ${\feat_{i+1}}^{V} = \feat_{i+1}$ \tcp{for transductive}
    \If{validation on different graph}{
        ${\emb_{i}}^{V} = \model[i].\cf{forward}(\feat_
        {i}^V)$\;
        $validation labels = \cf{softmax}({\emb_{i}}^{V})$\;
        ${\dv_{i}}^{V} =$ $\cf{findDegreeVec}(\Gc^V, validation labels, \frac{\epsilon}{nl}$)\;
        \If{i is 0}{
            ${\feat_{i+1}}^{V} = {{\emb_{i}}^{V}}\concat {\dv_{i}}^{V}$\;
        }
        \Else{
            ${\feat_{i+1}}^{V} = {{\feat_{i}}^{V}}\concat({{\emb_{i}}^{V}}\concat{\dv_{i}}^{V})$\;
        }
        \tcp{concatenate validation logits and degree vectors}
    }
    $\model[i+1] = \cf{trainMLP}(\feat_{i+1}, \set{Y}^T, {\feat_{i+1}}^{V}, \set{Y}^V, hid_n, hid_s, lr, dr)$\;
    $i= i+1$\;
 }
 \Return $\model = [\model[0],\ldots,\model[nl]]$
 \caption{The training algorithm for \tool. In the transductive setting, the validation graph and the training graph are the same. For inductive, they can be different hence we recompute the degree vectors on validation graph.}
 \label{alg:lpgnet}
\end{algorithm}

\begin{algorithm}[t]
\SetAlgoLined
\SetKwInOut{Input}{Input}
\SetKwInOut{Output}{Output}
\Input{Inference graph $\Gc^{I}:(\Vc^{I},\Ec^{I})$, Inference Features: $\feat^I$, Privacy budget: $\epsilon$, Trained model: $\model$.}
\Output{Node embeddings: $\emb$ for $\Vc^{I}$}
 $nl = $\# of additional MLPs in $\model$\;
 $i = 0, \feat_0 = \feat^I$\;
 \While{$i < nl$}{
    $\emb_{i} = \model[i].\cf{forward}(\feat_i)$\;
    \tcp{get logits from one forward pass}
    $labels = \cf{softmax}(\emb_{i})$\;
    \If{inductive setting and first time inference}{
        $\dv_i = \cf{findDegreeVec}(\Gc^I, labels, \frac{\epsilon}{nl})$\;
        Store to disk $\dv_i$\;\tcp{for future inference}
    }
    \Else{
        $\dv_i \xleftarrow[]{}$ Use Stored Degree Vectors\;
    }
    \If{i is 0}{
        $\feat_{i+1} = \emb_{i}\concat \dv_{i}$\;
    }
    \Else{
        $\feat_{i+1} = \feat_{i}\concat(\emb_{i}\concat\dv_{i})$\;
    }
    \tcp{concatenate logits and degreevectors}
    
    $i= i+1$\;
 }
 \Return $\emb = \model[nl].forward(\feat_{i})$
 \caption{The inference algorithm for \tool. In the transductive setting, the inference graph is same as the train graph $\Gc^T$ whereas in the inductive setting it is different. Therefore, in the inductive setting the degree vectors are computed on the new graph. Further, only the embeddings (logits) computed on the last MLP are released.}
 \label{alg:lpgnet_inf}
\end{algorithm}

\begin{algorithm}[h]
\SetAlgoLined
\SetKwInOut{Input}{Input}
\SetKwInOut{Output}{Output}
\Input{Graph $\Gc:(\Vc,\Ec)$, $labels$, Privacy budget: $\epsilon$}
\Output{Degree vectors: $\dv$}
\SetKwFunction{DV}{$\cf{findDegreeVec}$}
\SetKwProg{Fn}{Function}{:}{}

\Fn{\DV{$\Gc$, $labels$, $\epsilon$}}{
     $Cl = \cf{getClusters}(labels)$\;
     \tcp{nodes with same label is a cluster}
     \For{node $v$ in $\Gc$}{
        \For{cluster $c$ in $Cl$}{
            $n$ = \# of $\cf{neighbors}(v)$ in $c$\; 
            $\dv[v][c] = n + \cf{Lap}(0, \frac{2}{\epsilon})$\;
            \tcp{no need to noise for non-DP}
        }
     }
     \Return $\dv$
 }
 \caption{Compute degree vectors.}
 \label{alg:lpgnet_dv}
\end{algorithm}

\subsection{LPGNet's Privacy Details}
\label{sec:ppandimpl}

\paragraph{Differential privacy} 
To satisfy edge-DP, \tool adds noise sampled from $\cf{Lap}(0,\frac{2}{\epsilon})$ to all counts in the cluster degree vectors whenever they are computed from the graph (Lines $6$, $20$ in Algorithm~\ref{alg:lpgnet} and Line $7$ in Algorithm~\ref{alg:lpgnet_inf}). Without DP, \tool can train many additional MLPs, but with DP \tool can train only a few since the privacy budget adds up for every such \mlp. For instance, if $\epsilon=2$ for computing the cluster degree vectors once then after $5$ additional MLPs the resulting $\epsilon$ is $10$.

\ash{In our implementation, we fix a privacy budget that is consumed during the training, validation and inference phases. We split that budget between all possible degree vector queries across the three phases. Whenever possible, we save the privacy budget by reusing cluster degree vectors during validation and inference phases (Lines $8, 11$ in Algorithms~\ref{alg:lpgnet} and~\ref{alg:lpgnet_inf} respectively). \tool preserves $\epsilon$ edge-DP based on the post-processing property since the all computations build on the DP cluster degree vectors. Concretely, we prove the following theorems to show that \tool provides $\epsilon$ edge-DP in both the transductive and inductive settings.}

\begin{restatable}{theorem}{thma}
\label{thm:t1}
LPGNet satisfies $\epsilon$ edge-DP for the transductive setting.
\end{restatable}

\begin{restatable}{theorem}{thmb}
\label{thm:t2}
LPGNet satisfies $\epsilon$ edge-DP for the inductive setting.
\end{restatable}

We provide the details of how \tool uses its privacy budget and the proofs for the aforementioned theorems in Appendix~\ref{appdx:privacy_details}.


\section{Empirical Evaluation}
\label{sec:eval-setup}

Our primary goals are three-fold. First, we want to evaluate how well \tool will perform in the node-classification task as compared to \mlp and \gcn. Second, we want to measure how much utility is traded off when \tool is trained with a DP guarantee and check if \tool performs better than MLP and \baseline. Third, we want to compare the privacy-utility tradeoffs offered by \tool and \baseline when the privacy is measured by how well the $2$ state-of-the-art attacks--- \sotaAttack~\cite{wu2021linkteller} and \baselineAttack~\cite{he2021stealing}---work. 


\subsection{Methodology}

\paragraph{Datasets.}
\ash{We use $6$ standard datasets for the node classification task used in prior work~\cite{kipf2016semi,musae,wu2021linkteller}. In the transductive setting, we use \fbpage, \cora, \seer, and \pubmed datasets. \fbpage is a social network and the rest are citation networks. In the inductive setting, we use \twitch and \flickr datasets and both are social networks. \twitch is a collection of disjoint graphs, one from each country. Across all graphs the node features have the same dimension and the semantic meaning. The task is to classify whether a \twitch node (streamer) uses explicit language. We train GNNs on the graph corresponding to Spain and use other graphs for testing as done previously~\cite{wu2021linkteller}. \flickr dataset contains an evolving graph. We train GNNs on one part of the graph and test it on the evolved graph. We provide details of the datasets, their graph statistics and the test-train splits in Appendix~\ref{appdx:eval-setup}.}

 
\paragraph{Model architectures.}
We have $3$ model types for all our experiments: \gcn (or \baseline), \mlp, and \tool. For \gcn, we adopt the standard architecture as described in Section~\ref{sec:problem} along with ReLU activation and dropout for regularization in every hidden layer. For \mlp, we use a fully-connected neural network with ReLU activation and dropout in every hidden layer. In \tool, we use the same \mlp architecture for all stacked layers however with different input size for each layer. We use a softmax layer to compute the labels from the logits generated by every model. The number of hidden layers, dropout rate and size of the hidden layers are hyper-parameters which can be tuned. We find the right parameters by performing a grid search over a set of possible values in the non-DP setting which we provide in Appendix~\ref{appdx:eval-setup}. In the DP setting, we evaluate using the best hyper-parameter values from the non-DP setting. There may be better values for the DP setting but finding them would require using additional privacy budgets and we leave such tuning for future work. We also believe that our conclusions will only be stronger with better parameter tuning for DP models.

\paragraph{Training procedure and metrics.}
\ash{We train our models using cross-entropy loss, Adam's optimizer, and weight decay. We evaluate the node-classification performance using the micro-averaged F1 score. As proposed by the \sotaAttack paper, we use a modified F1 score for \md{Twitch} datasets which computes the F1 score for the rare class as there is a significant class imbalance in the datasets. We provide more details for our training procedure and metrics in Appendix~\ref{appdx:eval-setup}.}

\paragraph{Existing attacks.}
\sotaAttack is based on the observation that two neighboring nodes will influence each other's predictions more than two non-neighboring nodes. \ash{To compute the influence, \sotaAttack perturbs the features of a target node and observes the changes in the GNN outputs for every other node. For a $k$-layer \gcn, only the nodes within $k$ hop distance of the target node on the graph are influenced. For a $1$-layer \gcn the \sotaAttack reveals the exact edges of the graph used during the inference time. Therefore, \gcn and similar GNN architectures leak edges to \sotaAttack since the influence scores are propagated through edges of the graph in such architectures.} We also consider a more general attack, \baselineAttack, that is not tuned to any specific architecture given by He et al.~\cite{he2021stealing}. \baselineAttack has many attack scenarios based on the partial knowledge available with the attacker. Here, we choose the one that is most applicable in our setting, i.e., the attacker has access to only the inference node features $\feat^I$. Specifically, \baselineAttack uses the features and the obtained embeddings $\emb^I$ to predict the likelihood of having an edge. Nodes with more similar embeddings are ranked higher for the likelihood of having an edge. We reproduce both of these attacks successfully by following their protocols exactly.

\paragraph{Attack procedure and metrics.}
We follow the same attack procedures set by the previous works~\cite{he2021stealing,wu2021linkteller}. For the transductive setting, we sample a set of $500$ edges and $500$ non-edges to measure how well the attacks can separate the edges from non-edges. For the inductive setting, we sample a set of $500$ inference nodes $\Vc^I$ to create a subgraph that has not been trained on and then we evaluate the attacks on all the edges and non-edges in the subgraph. Further, \sotaAttack measures the performance on high and low-degree nodes separately to understand how their attack performs on nodes with different degrees. They find out that the attack performs better on high-degree nodes. We also evaluate the attack on these additional scenarios on the datasets they evaluate and report the values in the Appendix~\ref{appdx:ind-appx}. \ash{We use the Area Under Receiver Operating Curve (AUC) metric to measure the attack performance as done in the \baselineAttack and \sotaAttack works. This metric has also been extensively used for measuring inference attack performance on other kinds of private data in prior work~\cite{he2021stealing,wu2021linkteller,Pyrgelis2018KnockKW,choquette2021label}. For graph data, the current methodology to measure AUC asks how well an attack can identify a randomly sampled existing edge from a non-edge. Higher AUC is typically interpreted as the attack being more successful in identifying the edges used by the model, i.e., the model is less resilient to the attack. We report the average performance of each attack across $5$ seeds.}

\begin{table}[t]
\centering
\caption{Node classification results for the transductive setting using micro-average F1 scores when the models are not trained with DP. \tool significantly outperforms \mlp, in all datasets, which confirms the generality of  our key ideas.}

\begin{tabular}{c c c c c c}
\toprule
Dataset &  \mlp  & \toola & \toolb  & \gcn\\\hline

\fbpage & $0.74 \pm 0.01$ & $0.83 \pm 0.01$ & $0.87 \pm 0.01$  & $0.93 \pm 0.01$\\
\seer & $0.60 \pm 0.01$ & $0.65 \pm 0.01$ & $0.67 \pm 0.01$ & $0.72 \pm 0.0$\\
\cora & $0.60 \pm 0.01$ & $0.69 \pm 0.02$ & $0.73 \pm 0.01$ & $0.81 \pm 0.0$\\
\pubmed & $0.73 \pm 0.01$ & $0.75 \pm 0.0$ & $0.75 \pm 0.0$ & $0.79 \pm 0.0$\\
\hline
\end{tabular}
\label{tab:utlity-nondp-trans}
\end{table}

\begin{figure*}[ht!]
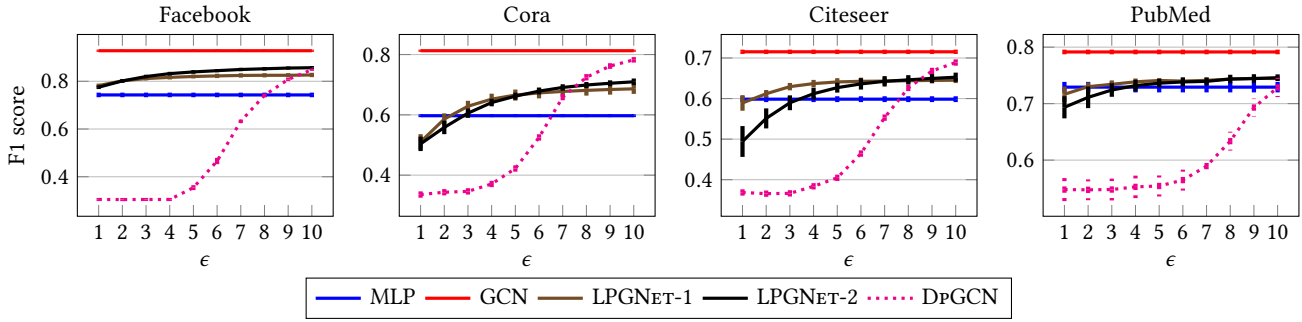

    \centering
    \begin{tikzpicture}
        \begin{groupplot}[group style = {group size = 4 by 1, horizontal sep = 25pt},
                width=0.28*\textwidth,
                height = 4.0cm]
            \input{figures/utility_dp_fb_trans}
            \input{figures/utility_dp_cora_trans}
            \input{figures/utility_dp_citeseer_trans}
            \input{figures/utility_dp_pubmed_trans}
        \end{groupplot}
        \node at ($(group c2r1) + (1.75cm,-2.3cm)$) {\ref{grouplegend2}};
    \end{tikzpicture}
    \caption{Utility of DP models in the transductive setting for various $\epsilon$.}
    \label{fig:utility-dp-trans}
\end{figure*}

\subsection{Results: Transductive Settings}
\label{sec:eval-perf}
\begin{figure*}[htbp]
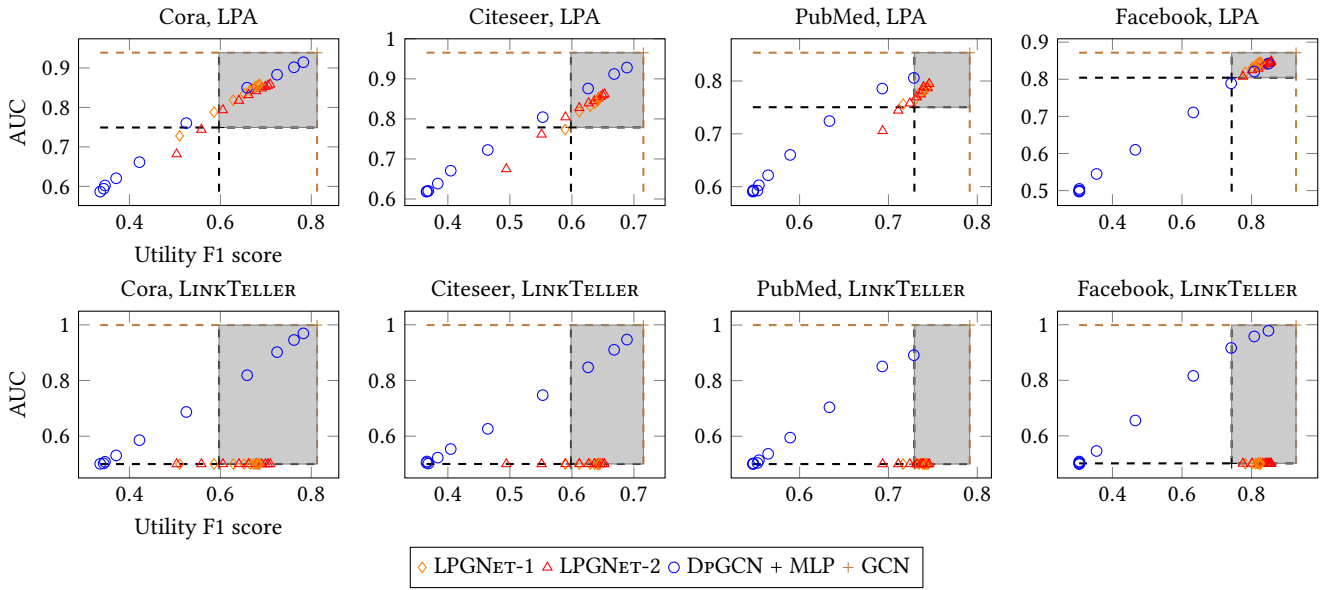

    \centering
    \begin{tikzpicture}
        \begin{groupplot}[group style = {group size = 4 by 2, horizontal sep = 25pt, vertical sep = 40pt},
                width=1/3.53*\textwidth,
                height = 3.8cm]
            \input{figures/utility_attack_baseline_cora}
            \input{figures/utility_attack_baseline_citeseer}
            \input{figures/utility_attack_baseline_pubmed}
            \input{figures/utility_attack_baseline_facebook_page}
            \input{figures/utility_attack_efficient_cora}
            \input{figures/utility_attack_efficient_citeseer}
            \input{figures/utility_attack_efficient_pubmed}
            \input{figures/utility_attack_efficient_facebook_page}
        \end{groupplot}
        \node at ($(group c2r2) + (1.75cm,-2.3cm)$) {\ref{grouplegend2}};
    \end{tikzpicture}
    \caption{Attack AUC vs Utility F1-score in the transductive setting for two state-of-the-art attacks. If we fix the utility level, \tool models always have better resilience to attacks than the corresponding \baseline model with that utility.}
    \label{fig:utility-vs-attack-trans}
\end{figure*}
%
We will denote \tool with $nl$ additional \mlp as \tool-$nl$. 

\paragraph{Utility in the non-DP setting. }
We compare the utility offered by \tool to that of \mlp and \gcn in the non-DP setup to gauge how well our key observations in Section~\ref{sec:approach} generalize to all datasets. In Table~\ref{tab:utlity-nondp-trans} we present the F1 scores of the best parameter-tuned models. First, observe that the \gcn outperforms the \mlp by $6$-$19\%$ across all datasets. Second, \toola outperforms the \mlp as well by $2$-$9\%$ with just one additional stack layer. With every additional stack layer, \tool's performance improves across all datasets with \toolc (not shown here) performing $4$-$15\%$ better than \mlp and only $4$-$6\%$ worse than the \gcn, respectively. This shows that our idea of iteratively improving the homophily signal in the clusters captured by \mlp by using \degreevecs works well on all datasets in the transductive setting. Note that \tool achieves this performance using the coarse-grained \degreevecs instead of using fine-grained edge information as in \gcn.

\paragraph{Utility in the DP setting.}
When we add noise, we observe that \tool's utility is not affected as much as that of \baseline models. Consequently, on all datasets in the transductive setting, \tool performs better than \mlp and the \baseline at a majority of privacy budgets $\epsilon \in [1,10]$ we evaluate. 
We compare \baseline with only \toola and \toolb as the privacy budgets increase with every additional stack layer used in \tool. Figure~\ref{fig:utility-dp-trans} shows the comparison between \tool, \mlp, and \baseline. \tool models perform better than the \mlp on all datasets for $\epsilon\geq2$. 
In contrast, \baseline performs up to $2.6\times$ worse than \tool at $\epsilon\leq2$. \tool provides better utility than \baseline for $2 < \epsilon < 7$ on all datasets as well. With $\epsilon\geq7$, \tool performs better than \baseline on $2$ out of the $4$ evaluated datasets. 

\custombox{1}{\tool outperforms \mlp for all $\epsilon\geq2$ and up to $2.6\times$ better than \baseline for relatively low  $\epsilon \in [1,7]$ on all datasets.}

\subsubsection*{Privacy resilience measured via attacks}
\label{sec:eval-tradeoffs}
The privacy budget $\epsilon$ is a theoretical upper bound on the amount of privacy leaked by a DP-trained model. Therefore, it may not be meaningful to say that two models offer same level of actual privacy if trained with same $\epsilon$. To estimate the true privacy resilience, we evaluate how well state-of-the-art link stealing attacks, \sotaAttack and \baselineAttack, perform on \tool and \baseline. Attack-based evaluations are common for other DP systems as well~\cite{jayaraman2019evaluating,nasr2021adversary}. 

Figure~\ref{fig:utility-vs-attack-trans} shows the tradeoffs with utility (classification accuracy) plotted on the x-axis and attack accuracy on y-axis. Each point on the plot is for a distinct choice of model and $\epsilon$. The performance of non-DP models, namely vanilla \mlp and \gcn, are baselines to compare DP solutions with and shown with dashed lines. We expect the attack resilience to be the best (lowest attack accuracy) for \mlp since it does not use edges at all. We also expect (vanilla) \gcn to offer the best utility since it uses the full raw edge structure with no noise. We color in grey the sweet spot region on the plot, where the utility is better than \mlp but worse than that of \gcn, and the attack resilience better than \gcn but worst than \mlp.

\ash{Figure~\ref{fig:utility-vs-attack-trans} shows that, compared to \baseline,  \tool's models are tightly concentrated in the sweet spot region and more towards the south-east side of that region. This implies that its utility is closer to \gcn and attack AUC is closer to \mlp.}

\ash{We zoom in to the only $10$ configurations where \baseline  has better utility than \mlp and compare its attack AUC with \tool. We provide the results in Appendix~\ref{appdx:eval-trans-addn}, Table $7$. \tool outperforms \baseline in both utility and attack resilience in \fbpage and \pubmed datasets. For \cora and \seer, the attack resilience of \tool is always better than \baseline. The best attack has an AUC of $0.85$-$0.98$ on \baseline which is about $10$-$22\%$ higher than the AUC on \mlp in absolute difference. In contrast, the best attack has an AUC of $0.79$-$0.86$ on \tool which is about $4$-$11\%$ higher than the AUC on \mlp. Thus, \tool offers better privacy-utility tradeoffs compared to \baseline\xspace{\em consistently} across all datasets.}

\ash{The AUCs already show that \tool leaks less edge information than \dpgcn. However, the AUC of $0.86$ for \tool is still high. So, does that mean \tool is leaking a lot of edge information? In short, the answer is {\em No} and we discuss this in Section~\ref{sec:discuss-eval}.}


\custombox{2}{\tool often hits the sweet spots and offers significantly better attack resilience than \baseline at similar utility.}

Due to space constraints, the full utility and attack resilience data for different $\epsilon$ values is left to the Appendix~\ref{appdx:eval-trans-addn}.

\subsection{Results: Inductive Setting}
\label{sec:eval-util-inductive}

\begin{figure}[t]
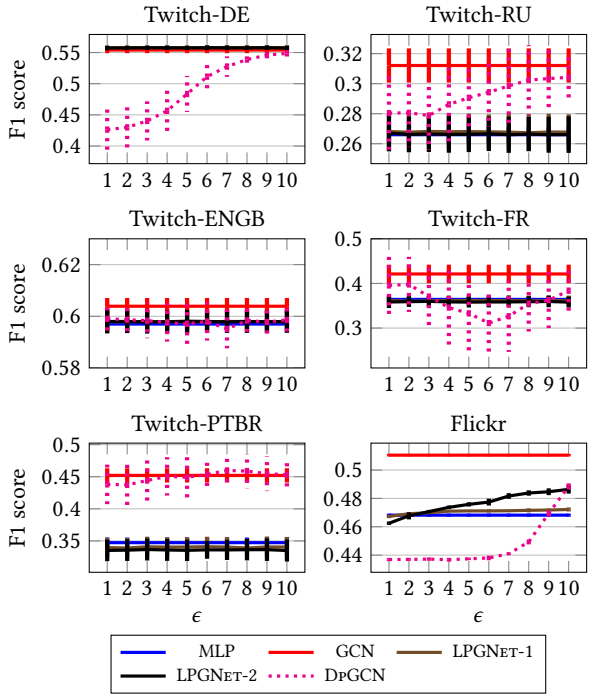

    \centering
    \begin{tikzpicture}
        \begin{groupplot}[group style = {group size = 2 by 3, horizontal sep = 25pt},
                width=1/4.0*\textwidth,
                height = 3.3cm]
            \input{figures/utility_dp_de_ind}
            \input{figures/utility_dp_ru_ind}
            \input{figures/utility_dp_engb_ind}
            \input{figures/utility_dp_fr_ind}
            \input{figures/utility_dp_ptbr_ind}
            \input{figures/utility_dp_flickr_ind}
        \end{groupplot}
        \node at ($(group c2r2) + (-2.0cm,-4.8cm)$) {\ref{grouplegend2}};
    \end{tikzpicture}
    \caption{Utility of DP models in the inductive settings.}
    \label{fig:utils-dp-ind}
\end{figure}

The inductive setting is challenging for many state-of-the-art GNN architectures as they have to transfer their knowledge to a different graph~\cite{kipf2016semi,hamilton2017inductive}. Here, we measure how well \tool generalizes to unseen graphs. Recall that we have a different evaluation procedure for both utility and attack performance in this setting (see Section~\ref{sec:eval-setup}).

\paragraph{Utility in the non-DP setting.}
We observe that all the models perform significantly worse than in the transductive setting and their performances are very close to each other as well. The vanilla \gcn performs only $4$-$11\%$ better than the \mlp. In the \md{Twitch} datasets, \tool is either on-par or improves by $1\%$ over \mlp  in all of its configurations ($5$ datasets $\times 2$ \tool architectures). For \flickr dataset, \tool improves over \mlp by $1$-$3\%$ when \gcn itself is $4\%$ better than \tool. Therefore, even in the inductive setting, \tool is a better choice than \mlp utility-wise as it is on-par or bridges the gap between \mlp and \gcn across all datasets.

\paragraph{Utility in the DP setting.}
For all $\epsilon \in [1,10]$, we observe that \toolb has better utility than \mlp in \flickr and is on-par with \mlp in \twitch. Figure~\ref{fig:utils-dp-ind} shows utility of \tool and \dpgcn models. \toola's performance changes at most by $1\%$ at all levels of $\epsilon$ and \toolb performs better than \toola by up to $2\%$ across all datasets. In $4$ out of $6$ datasets, \toolb performs on-par or better than \baseline at all $\epsilon$s, therefore, achieving better utility at the same theoretical privacy guarantee. \tool models have worse utility than \baseline models for \twitchptbr and \twitchru.

On \twitch datasets, \tool's lack of improvement over \mlp can be explained due to less useful information obtained from the cluster degree vectors. In \twitch, every node has similar number of neighbors from both clusters (on average) making it difficult to distinguish between the degree vectors of nodes from both clusters. For instance, nodes from cluster-$1$ in \twitchru have $75\%$ of their neighbors from cluster-$1$ and $25\%$ neighbors from cluster-$2$ as expected where as nodes from cluster-$2$ also have $70\%$  of their neighbors from cluster-$1$ and only $30\%$ from cluster-$2$. We observe a similar pattern for all \twitch datasets. This lack of homophily affects all architectures and \tool is the most affected since it depends on cluster degree vectors rather than the exact edge information. Nevertheless, even with DP noise, \tool reliably performs at least as good as \mlp unlike \dpgcn.


\custombox{3}{\toolb reliably performs on-par or better than \mlp for both \flickr and \twitch datasets.}

\paragraph{Privacy resilience measured via attacks.} 
We first observe that \tool has good resilience against both attacks, i.e., attack AUC is lesser than $0.65$ across all evaluated configurations.  \baseline, in contrast, has significantly worse attack resilience than \tool for \twitch and \flickr datasets with best attack AUC often reaching up to $0.98$ even at the same utility level as \tool. In $2$ out of $5$ \twitch datasets where \dpgcn has better utility than \tool the attack AUC is above $0.9$ for most $\epsilon$s whereas the attack AUC on \tool is almost $0.5$. For \flickr, we zoom in on cases where {\em both} \tool and \baseline have utility better than \mlp (see Table $12$ in Appendix~\ref{appdx:ind-appx}). \tool offers {\em significantly} better resilience against attacks than the corresponding \baseline with the same level of utility. The best attack on \tool has an AUC of $0.56$, at $\epsilon=10$, but $0.90$ on \baseline ($34\%$ worse than \tool). Therefore, our Result $2$ from Section~\ref{sec:eval-tradeoffs} is true for the inductive setting as well.

We provide the detailed results including utility and attack AUCs of all configurations for the inductive setting in the Appendix~\ref{appdx:ind-appx}.




\subsection{Discussion: Interpreting the Results}
\label{sec:discuss-eval}

\ash{We explain some implications of our findings which may not be immediately obvious from knowledge of prior attacks~\cite{he2021stealing,wu2021linkteller}.}

\subsubsection{\baselineAttack has relatively high AUC against \tool. So, is \tool leaking the edges against \baselineAttack?}
\ash{The answer depends on the nature of the attack and its evaluation strategy. In prior work, the attack AUC is evaluated on the ability to distinguish a randomly sampled edge from a non-edge. The AUC can be high even when no individual edges are used (memorized) by the trained model. For example, \baselineAttack's high AUC is an artifact of the distribution of the graphs that are being learned by the GNNs. To understand this, consider homophilous graphs which have $80\%$ of their edges connecting nodes from the same cluster and only $20\%$ of the time there is no edge connecting nodes within the same cluster. If the clusters learned by a GNN are revealed, then it is easy to guess which nodes are more likely to have an edge between them. A simple ``attack'' would predict that an edge exists between two nodes within the same cluster, otherwise not, and have high AUC. This case would arise even if an \mlp, that uses just node features, discovers these clusters. \tool (or any edge-DP mechanism) does not prevent such attacks that predict edges because they belong to a distribution that is being learned by the ML model. To confirm that \baselineAttack is our simple attack, we take the top-$|\Ec|$ edges as predicted by \baselineAttack for \gcn trained on the \cora dataset which has the aforementioned distribution of edges and non-edges. Almost $99\%$ of its edge predictions were intra-cluster. }
\begin{figure}[tbp]
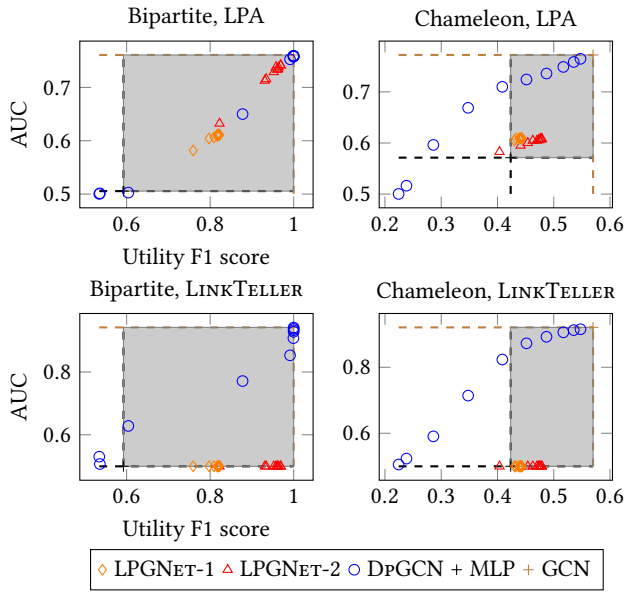

    \centering
    \begin{tikzpicture}
        \begin{groupplot}[group style = {group size = 2 by 2, horizontal sep = 25pt, vertical sep = 40pt},
                width=1/3.8*\textwidth,
                height = 3.8cm]
            \input{figures/utility_attack_baseline_bipartite}
            \input{figures/utility_attack_baseline_chameleon}
            \input{figures/utility_attack_efficient_bipartite}
            \input{figures/utility_attack_efficient_chameleon}
        \end{groupplot}
        \node at ($(group c1r2) + (2.0cm,-2.3cm)$) {\ref{grouplegend3}};
    \end{tikzpicture}
    \caption{Attack AUC vs Utility in the transductive setting for \bipartite and \chameleon datasets, with low homophily.}
    \label{fig:utility-vs-attack-trans-extra}
\end{figure}
\ash{Unlike \baselineAttack, \sotaAttack's high AUC is an artifact of its ability to tell whether a specific edge is in the training graph irrespective of the distribution of node features and edges. Therefore, a high AUC for \sotaAttack suggests that the model is responsible for leaking the edges used for training or inference. It is thus expected that \sotaAttack's AUC on \mlp should be $0.5$. \tool has the edge-DP guarantee, so it protects against such attacks when the model uses individual edges for training and inference.}

\ash{In summary, it is misleading to conclude that GNNs surely leak individual edges based on just the attack's AUC; instead, we must take into account the attack details (e.g., \baselineAttack vs. \sotaAttack).} 

\subsubsection{Why is \mlp a good baseline?}
\ash{
When we compute the attack AUC, as we explained, it is possible to get a high score whether or not the edges have been used by the GNN.~\footnote{The precision and recall will be similar to AUC for this evaluation methodology.} If we want to marginally compute the effect of using the edges in training then we would have to {\em causally} remove the edges from training and measure the attack performance. \mlp offers that baseline---it uses no edges. Therefore, the increase in attack AUC on a GNN vs \mlp indicates how much a GNN leaks about the edges which the attacker already does not know from its background knowledge about node features and the distribution of graphs to which the training graph belongs. For \tool, the increase in attack AUC over \mlp is smaller than in other architectures (see Figure~\ref{fig:utility-vs-attack-trans}).}

\subsubsection{Can \baselineAttack have low AUC on \mlp and \tool?}
\ash{Indeed, this is possible. Consider graphs where the node features do not correlate well with the clusters. In such graphs, the AUC of \baselineAttack on \mlp can be close to $0.5$ but the classification accuracy would be low as well. However, \tool can have a high classification accuracy while having a low \baselineAttack attack AUC. \tool, in this case, will almost entirely rely on the edge structure to improve the classification. We show this phenomena using two datasets. We first construct a synthetic bipartite graph, inspired by dating networks, so that the node features are not very correlated with the graph structure. We then take a real-world citation network (based on Wikipedia) that is popularly used as a challenge dataset for evaluating GNNs for its lack of homophily~\cite{pei2020geom,rozemberczki2021multi,zhu2020beyond}. \tool has low \baselineAttack AUC but higher classification accuracy than \mlp in these examples. We provide more details about the datasets in Appendix~\ref{sec:appdx-low-auc}.}

\ash{In Figure~\ref{fig:utility-vs-attack-trans-extra}, we provide the attack AUC vs utility plots for these two datasets. Observe that in both datasets, the AUC of best attack on \mlp is close to $0.5$. \tool has better utility than \mlp in almost all evaluated configurations. In the \bipartite graph, \toola itself is better than \mlp by $16-22\%$. In the \chameleon graph, \tool offers up to $6\%$ better utility than \mlp. \tool performs well on these datasets because two nodes within the same cluster are more likely to have similar distribution of labels in their neighborhood, hence similar cluster degree vectors. For instance, in the \bipartite graph, nodes of same cluster always have neighbors from the opposite cluster even though, by definition, it is not homophilous. \tool design can work for such non-homophilous graphs.}

\ash{The best attack on \toola has at most $0.61$ AUC on both datasets. The best attack on \toolb has an AUC of $0.74$ in the \bipartite graph and $0.61$ in the \chameleon graph. The attacks perform better on \dpgcn than \tool and the best attack has close to $0.9$ AUC on \dpgcn even at the same utility level as \tool.}

\subsubsection{How to prevent edge leakage from node attributes?}
\ash{It is natural to be concerned about how the node attributes themselves can leak edge information. However, the definitions of edge-DP or node-DP are not designed to address such privacy concerns. In our problem setup, the edge leakage from attributes corresponds to a different notion of privacy which asks for bounding the public information known to an attacker from knowing the attributes beforehand. One way to address this kind of leakage is to change the application setup itself. For instance, if an application allows users to store both of the node attributes and edges locally, then better privacy can be expected by using local DP for attributes and edges. Such setups are important future work and there is no existing work that designs DP algorithms for them, to the best of our knowledge.}

\section{Related Work}
\label{sec:related}

In this paper, we provide a differentially private learning technique for node classification on private graph-structured data.

\paragraph{Semi-supervised learning from graph data.}
Semi-supervised learning techniques on graphs combine the knowledge obtained from the graph structure and the node features for various applications such as node classification. Graph neural networks have become the de-facto standard to perform semi-supervised learning on graphs. Initial GNN architectures, inspired from recurrent neural networks~\cite{bengio1994learning}, apply the same parameterized function over node features recursively. The parameterized function performs a weighted average over the features of a node's neighbors to update its features~\cite{gori2005new,scarselli2008graph}. Current GNN architectures are inspired from convolutional neural networks~\cite{lecun1998gradient} where the input features are passed through {\em different} parameterized functions (convolutions) to compute the outputs. They mainly fall into two categories called spatial and spectral, depending on the nature of parameterized function used. The function could perform a weighted average over the features of neighboring nodes for each node (spatial)~\cite{gao2018large,xu2018powerful,velivckovic2017graph,hamilton2017inductive,chen2018fastgcn} or select the important features using the spectral decomposition of the graph Laplacian (spectral)~\cite{henaff2015deep,defferrard2016convolutional,kipf2016semi}. Refer to this extensive survey for existing GNN architectures~\cite{wu2020comprehensive}. In all these architectures, the adjacency matrix of the graph, i.e., the knowledge of exact neighbors is essential. Therefore, \sotaAttack is expected to perform well on them as demonstrated on two architectures, GCNs and graph attention networks~\cite{wu2021linkteller}. Further, in Sections~\ref{sec:problem} and \ref{sec:eval-perf} we show that noising the adjacency matrix using \baseline will adversely affect their utility due to severely perturbed graph structure.

Our idea is to  move away from the traditional GNN architectures and not use the adjacency matrix for propagation in intermediate layers. Rather, we propose to use only MLPs for node classification and use the graph structure in the form of \degreevecs as their input features to iteratively improve their classification performance. Our idea is in the same spirit of recent works that design purely MLP-based architectures to achieve comparable performance to state-of-the-art nets in vision~\cite{tolstikhin2021mlp,melas2021you}.

\paragraph{Attacks on graph neural networks.}
Attacks on graph neural networks to steal (infer) the edges is a recent phenomena. Duddu et al.~\cite{duddu2020quantifying} showed several attacks to predict the membership of both nodes and links with blackbox and whitebox~\footnote{Only the node embeddings at intermediate layers are visible} access to train the GCNs. The attack for identifying links uses another graph from the same distribution and node embeddings obtained from the victim GCN. He et al.~\cite{he2021stealing} showed link stealing attacks with blackbox access to a trained GCN (\baselineAttack) and more recently, Wu et al.~\cite{wu2021linkteller} showed, \sotaAttack, a stronger attack with blackbox access to the learned GCN and without needing access to features of nodes used in training. We have discussed these attacks in Section~\ref{sec:eval-setup}. Other attacks on GNNs have focused on poisoning the training dataset to affect the final classification results~\cite{dai2018adversarial,bojchevski2019adversarial}.   

\paragraph{Existing defenses against the attacks.}
\baseline is the only defense which provides an edge-DP guarantee against the aforementioned attacks for any graph. Zhou et al.~\cite{zhou2020privacy} proposes training a GCN in the federated setup but only adds noise after aggregating the features from the neighbors which is not sufficient for differential privacy. Wu et al.~\cite{wu2021fedgnn} proposes using private training of GNN for recommendations, but they model only bipartite graphs of user-item edges and their techniques do not extend to general graphs. A concurrent work proposes to train a GNN with node-DP guarantee which may require high $\epsilon$s for our setup as shown by the high privacy budgets used in that work ($\epsilon > 10$) to get a utility better than an MLP~\cite{daigavane2021node}. Further, they provide their privacy analysis only for single layer GCNs and do not evaluate the state-of-the-art attacks. Therefore, it is not clear if their models provide any privacy resilience against the attacks.  

\paragraph{Differentially private queries on graphs.}
For the edge-DP setup, many works design differentially private algorithms for estimating degree distributions~\cite{karwa2012differentially,blocki2013differentially,nissim2007smooth}, subgraph counts~\cite{nissim2007smooth}, synthetic graphs~\cite{lu2014exponential,xiao2014differentially,sala2011sharing,wang2013differential,wang2013preserving,mir2012differentially} and communities~\cite{mulle2015privacy,nguyen2016detecting,ji2019differentially}. Similarly, for the node-DP setup, there are algorithms for computing degree distributions~\cite{day2016publishing,raskhodnikova2015efficient} and subgraph counts~\cite{chen2013recursive,ding2018privacy}. All of the aforementioned algorithms are in the centralized DP setup where a central server is trusted. In the LDP setup where there is no trusted centralized server, there are edge-LDP algorithms for estimating aggregate statistics on graphs~\cite{wei2020asgldp}, community detection~\cite{qin2017generating}, and hierarchical clustering~\cite{kolluri2021private}. Our proposed technique can be implemented in both the centralized and local setups since computing \degreevecs is an operation that requires only the knowledge of one's neighbors and their cluster labels.


\section{Conclusion}
\label{sec:conclusion}

We have presented \tool, a new stacked neural network architecture for learning graphs with privacy-sensitive edges. \tool is carefully designed to strengthen the property of homophily, when present in the graph. \tool provides differential privacy for edges. It exhibits meaningful utility-privacy tradeoffs compared to other existing architectures that either use edge-DP or are trivially edge-private in most evaluated datasets and configurations.

\begin{acks}
We thank Aneet Kumar Dutta and Bo Wang for helping us with parsing the results and generating tables for an earlier version of this paper. We are grateful for the constructive feedback from the anonymous reviewers and for conducting the revision procedure. We thank Kunwar Preet Singh for his help while revising the paper. We thank Crystal Center and its sponsors, Singapore Ministry of Education (MoE), National University of Singapore (NUS), and National Research Foundation Singapore (NRF) for their generous grants. Aashish Kolluri is supported by~\grantsponsor{CC}{Crystal Centre}{} under the Grant No.:~\grantnum{CC}{E-251-00-0105-01} and~\grantsponsor{moe}{MoE}{} under the Grant No.:~\grantnum{moe}{A-0008530-00-00}.
Teodora Baluta is supported by~\grantsponsor{nrf}{NRF}{} under its NRF Fellowship Programme ~\grantnum{nrf}{NRF-MRFFAI1-2019-0004} and \grantsponsor{moe}{MoE}{} under the Grant No.:~\grantnum{moe}{T2EP20121-0011}. Teodora Baluta is also supported by the Google PhD Fellowship.
Bryan Hooi is supported in part by~\grantsponsor{nus}{NUS}{} ODPRT grant under the Grant No.:~\grantnum{nus}{R252-000-A81-133}.
\end{acks}

\bibliographystyle{ACM-Reference-Format}
\bibliography{paper}

\appendix
\section{\tool's Privacy Details}
\label{appdx:privacy_details}

\begin{table}[h]
    \centering
    \caption{The budget split across different phases in $\tool$.}
    \begin{tabular}{c|c|c|c}
         Setting &  Training & Validation & Inference\\\hline
         Transductive & $\frac{\epsilon}{nl}$ & NA & NA \\
         Inductive-different & $\frac{\epsilon}{nl}$ & NA & $\frac{\epsilon}{nl}$ \\
         Inductive-evolving & $\frac{\epsilon}{3\cdot nl}$ & $\frac{\epsilon}{3\cdot nl}$ & $\frac{\epsilon}{3\cdot nl}$ \\\hline
    \end{tabular}
    \label{tab:budgetsplit}
\end{table}

In this section we explain \tool's privacy budget split in detail and prove that it satisfies $\epsilon$ edge-DP.

\paragraph{Privacy budget split} \ash{In the transductive setting, we reuse the degree vectors computed during the training time for inference since they are computed on the same graph (Lines $8, 11$ in Algorithms~\ref{alg:lpgnet} and~\ref{alg:lpgnet_inf} respectively). In the inductive setting, we only care about privacy of edges from the graph used during the inference time in our threat model. Therefore, for the inductive-different setting, we use the full privacy budget for computing the cluster degree vectors on the graph used for inference. Nevertheless, for completeness we ensure edge-DP while learning on the training graph as well with full privacy budget. For the inductive-evolving setting, we have to recompute the cluster degree vectors during training, validation and inference phases for every additional \mlp used in \tool. Therefore, we split the privacy budget across the training, validation and inference phases equally and further split the budget for every time we compute the cluster degree vectors. An example of the privacy budget split for a total budget $\epsilon$ is given in the Table~\ref{tab:budgetsplit}}.

We use this privacy budget split to show that \tool satisfies $\epsilon$ edge-DP, i.e., to prove the two theorems in Section~\ref{sec:ppandimpl}.

\thma*
\begin{proof}
\ash{The cluster degree vector computation satisfies $\epsilon$ edge-DP (see Algorithm~\ref{alg:lpgnet_dv}) as it uses the Laplace mechanism with corresponding sensitivity for edge-DP. \tool only accesses the private graph to compute cluster degree vectors during the training phase (Line $6$ in the Algorithm~\ref{alg:lpgnet}). It computes these vectors for all nodes before training each additional MLP with a privacy budget $\frac{\epsilon}{nl}$. Therefore, the total privacy budget for training \tool with $nl$ additional MLPs will be $\epsilon$  as given by the sequential composition of DP~\cite{privacybook}. For validation and inference the same degree vectors that are computed during training are used. Therefore, by post-processing property of DP, \tool satisfies $\epsilon$ edge-DP over the course of its training, validation and inference.}
\end{proof}

\thmb*
\begin{proof}
\ash{In the inductive-different setting, \tool computes cluster degree vectors on the inference graph just once for each additional \mlp with a budget $\frac{\epsilon}{nl}$ (Line $7$, Algorithm~\ref{alg:lpgnet_inf}). Therefore, for $nl$ additional MLPs the total privacy budget will be $\epsilon$ as given by the sequential composition. Since, \tool never uses the inference graph again, by post-processing property, it satisfies $\epsilon$ edge-DP.}

\ash{In the inductive-evolving setting, \tool computes degree vectors during all three phases because it is an evolving graph (Lines $6,20$ in Algorithm~\ref{alg:lpgnet}, and Line $7$ in Algorithm~\ref{alg:lpgnet_inf}). Therefore, according to our privacy budget split (in  Table~\ref{tab:budgetsplit}), the total budget used during training and validation is $\frac{2\cdot\epsilon}{3\cdot nl}\cdot nl$. The total budget used during inference is $\frac{\epsilon}{3\cdot nl}\cdot nl$. Therefore, by sequential composition, across all three phases the total privacy budget is $\epsilon$. Further, by post-processing property \tool satisfies $\epsilon$ edge-DP.}
\end{proof}
\section{Evaluation Setup}
\label{appdx:eval-setup}

Here we provide the additional details for our evaluation setup.

\begin{table}[ht]
    \centering
    \caption{Transductive setting}
    \resizebox{0.9\columnwidth}{!}{%
    \begin{tabularx}{1.1\columnwidth}{c c c c c c}
        \toprule
        Dataset & Nodes & Edges &  Density & Features & Classes\\ \hline
        \cora   & 2,708 & 5,429 & 0.0015 & 1,433 & 7 \\
        \seer   & 3,327 & 4,732 & 0.0008 & 3,703 & 6 \\
        \pubmed & 19,717 & 44,338 & 0.0002 & 500 & 3 \\
        \fbpage & 22,470 & 	171,002 & 0.0007 & 128 & 4 \\ \hline
    \end{tabularx}
    }
    \bigskip
    \caption{Inductive setting}
    \resizebox{0.9\columnwidth}{!}{%
    \begin{tabularx}{1.18\columnwidth}{c c c c c c}
        \toprule
         Dataset & Nodes & Edges &  Density & Features & Classes\\ \hline
         \md{Twitch-ES} & 4,648 & 59,382 & 0.0055 & 3,170 & 2 \\
         \md{Twitch-RU} & 4,385 & 37,304 & 0.0033 & 3,170 & 2 \\
         \md{Twitch-DE} & 9,498 & 153,138 & 0.0034 & 3,170 & 2 \\
         \md{Twitch-FR} & 6,549 & 112,666 & 0.0053 & 3,170 & 2 \\
         \md{Twitch-ENGB} & 7,126 & 35,324 & 0.0014 & 3,170 & 2 \\
         \md{Twitch-PTBR} & 1,912 & 31,299 & 0.0171 & 3,170 & 2 \\
         \flickr &  89,250 & 899,756 & 0.0002 & 500 & 7 \\ \hline
    \end{tabularx}
    }
    \label{tab:datastats}
\end{table}
\paragraph{Datasets}
\fbpage is a social network where nodes represent Facebook pages (sites) and the links represent mutual likes between the sites. Node features are extracted from the page descriptions provided by the site-owners. The latter three are citation networks where the nodes are documents and the edges are citations. The features represent existence/non-existence of certain key words. In the inductive setting, we choose two social networks \twitch and \flickr. \twitch provides a collection of disjoint graphs where nodes represent \twitch streamers from different countries and edges represent their mutual friendships. The features for each streamer are based on the games played and streaming habits. Across all graphs the features have the same meaning which makes transfer learning possible from one graph to another. The task is to classify whether a streamer uses explicit language. We use the graph corresponding to Spain to train the models and the graphs for $5$ other countries for testing as done previously. Hence, there are $6$ datasets in inductive setup: 5 for \twitch and 1 for \flickr.  Finally, the \flickr dataset is taken from the social network where users showcase their images. The nodes in this dataset represent the images and links represent similar image metadata such as images uploaded by , user's friends, same location, same group, and so on. The network is evolving and the task is to classify unseen images by transferring the knowledge from the seen ones.

The dataset statistics are provided in the Table~\ref{tab:datastats}. We follow the same data splits for training for \fbpage, \cora, \seer and \pubmed as considered by the previous works~\cite{kipf2016semi,yang2016revisiting}. The train set consists of randomly chosen $20$ labels per class , the  test set consists of $1000$ nodes and the validation set consists of $500$ labeled examples. The entire graph structure is used for training. For \twitch we do not use any validation set, instead, we use the models with the least training loss. For \flickr, we use a $50$-$25$-$25$ split to create the train, test and validation sets respectively. We do not use the validation or test labels to train the models and use only the graph formed by the nodes in the training and validation set. While evaluating the models we use the whole graph of Flickr and evaluate it on the labels of test nodes.

\paragraph{Training procedure and metrics}
In all datasets, except \twitch, we train the models, including each stacked layer in \tool, for $500$ epochs and checkpoint the models after every epoch to find the best performing model on the validation set during training. This helps us to measure the privacy leakage for the model with best utility. For \twitch we do not have a validation set, so instead we train for $200$ epochs in order to reproduce the findings of the \sotaAttack paper~\cite{wu2021linkteller}. We observe that all models converge within that time as reported there. We evaluate the node-classification performance using the micro-averaged F1 score. Since \md{Twitch} presents a binary classification problem, to compute the rare F1 score we identify the minor class and consider them positive samples. We run the training for $30$ seeds for all models and report the standard deviation. 
The micro averaged F1 score is defined as follows.

\begin{definition}[Micro averaged F1 score]
If there are $C$ classes and $TP_i$, $FP_i$, $FN_i$ represent the number of true positives, false positives and false negatives for a class $i$ the the micro-averaged precision, recall and F1 scores are defined as follows:
\begin{align*}
    Precision_{micro} &= \frac{\sum_{i=1}^{C} TP_i}{\sum_{i=1}^{C} TP_i + \sum_{i=1}^{C} FP_i}\\
    Recall_{micro} &= \frac{\sum_{i=1}^{C} TP_i}{\sum_{i=1}^{C} TP_i + \sum_{i=1}^{C} FN_i}\\
    F1_{micro} &= \cf{HarmonicMean(Precision_{micro}, Recall_{micro})} 
\end{align*}
\end{definition}

\paragraph{Attack Metrics}
For evaluating the attacks we use the Area Under Curve (AUC) metric. The AUC measures the area under the receiver operating characteristic (ROC) curve which plots the true positive rate against the false positive rate of a binary classifier. A high AUC implies that the classifier is able to tell apart a randomly chosen positive sample from a randomly chosen negative sample with high probability. A classifier which predicts labels randomly would have an AUC 0.5 which implies the classifier cannot tell apart the positive samples from the negative. The best AUC is 1 which implies the model can achieve a true positive rate of 1 even at the lowest false positive rate 0 indicating that the model perfectly classifies the samples. In our context, an attack achieves high AUC when it can distinguish randomly sampled edges from the non-edges most of the time.

\paragraph{Hyperparameters}
We choose the following values for hyperparameter tuning, using grid search, for all models. 
\begin{itemize}
    \item learning rate: $[0.005, 0.001, 0.01, 0.05]$
    \item Hidden layer size: $[16, 64, 256]$
    \item Hidden layers: $[2,3]$ for GCN and all MLPs.
    \item dropout: $[0.1, 0.3, 0.5]$
\end{itemize}
Further, we use two kinds of normalization of adjacency matrices for GCN (see the architecture in Section~\ref{sec:problem}). Apart from \twitch, for all other datasets, we use the Augmented Normalized Adjacency technique and for \twitch we use the First Order GCN normalization technique similar to previous works. Specifically, they are defined as follows:
\begin{align*}
    \cf{FirstOrderGCN} &: \arr{I} + \arr{D^{-0.5}}\cdot\arr{A}\cdot\arr{D^{-0.5}}\\
    \cf{AugNormAdj} &: (\arr{D}+\arr{I})^{-0.5}\cdot\arr{A}\cdot(\arr{D}+\arr{I})^{-0.5}\\
    \text{where }\arr{D} &= \cf{diag}(d_0, d_1, \cdots, d_{|\set{V}|-1})\text{, }d_i=\cf{degree}(v_i)
\end{align*}
For \flickr we choose the hyperparameters outside of our grid search which are known to achieve good performance for GCN and MLP i.e., $2$ hidden layers with size $256$ each, learning rate $0.0005$ and dropout rate $0.2$. We use these hyperparameters for \tool as well as we observe it to be better than the models we trained. 

\section{Additional Evaluation}
\label{sec:appdx-addn-eval}
\subsection{Transductive Setting}
\label{appdx:eval-trans-addn}
\begin{table}[htbp]
\caption{Utility and best attack AUC scores for configurations where \dpgcn and \tool are better than the private \mlp in the transductive setting. \toola has even better attack resilience than \toolb (not shown).}
\label{tab:final-relevant-attack-utility}
\resizebox{\columnwidth}{!}{%
\begin{tabular}{cccccc}
Dataset & $\epsilon$ & \begin{tabular}[c]{@{}c@{}}F1-score\\  (\dpgcn)\end{tabular} & \begin{tabular}[c]{@{}c@{}}Best Attack\\  (\dpgcn)\end{tabular} & \begin{tabular}[c]{@{}c@{}}F1-score\\ (\toolb)\end{tabular} & \begin{tabular}[c]{@{}c@{}}Best Attack\\ (\toolb)\end{tabular} \\ \hline
\multirow{4}{*}{\seer}
 & 8 & 0.63 & 0.88 & 0.65 & 0.86 \\
 & 9 & 0.67 & 0.92 & 0.65 & 0.86 \\
 & 10 & 0.69 & 0.95 & 0.65 & 0.86 \\ \hline
\multirow{5}{*}{\cora}
 & 7 & 0.66 & 0.85 & 0.69 & 0.85 \\
 & 8 & 0.72 & 0.90 & 0.70 & 0.85 \\
 & 9 & 0.76 & 0.95 & 0.70 & 0.86 \\
 & 10 & 0.78 & 0.97 & 0.71 & 0.86 \\ \hline
\multirow{3}{*}{\fbpage}
 & 9 & 0.81 & 0.96 & 0.86 & 0.85 \\
 & 10 & 0.85 & 0.98 & 0.86 & 0.85 \\ \hline
\pubmed
 & 10 & 0.73 & 0.89 & 0.75 & 0.79\\\hline
\end{tabular}%
}
\end{table}

Tables~\ref{tab:cora-dp-ind},~\ref{tab:citeseer-dp-ind},~\ref{tab:pubmed-dp-ind} and~\ref{tab:facebook-dp-ind} detail the utility results for the transductive setting. We report the attack performance in the Table~\ref{tab:trans_attack}.

\ash{Further, we zoom in to the $10$ configurations where \baseline performs better than \mlp in Table~\ref{tab:final-relevant-attack-utility}.}


\begin{table}[h!]
\centering
\caption{Node classification utility on Cora.}
\resizebox{\columnwidth}{!}{
\begin{tabular}{c c c c c c}
\toprule
Epsilon &  \mlp  & \toola & \toolb  & \gcn  &  \dpgcn \\\hline

1.0 & $0.6 \pm 0.0$ & $0.51 \pm 0.03$ & $0.5 \pm 0.02$ & $0.81 \pm 0.0$ & $0.34 \pm 0.02$\\
2.0 & $0.6 \pm 0.0$ & $0.59 \pm 0.02$ & $0.56 \pm 0.02$ & $0.81 \pm 0.0$ & $0.34 \pm 0.02$\\
3.0 & $0.6 \pm 0.0$ & $0.63 \pm 0.02$ & $0.61 \pm 0.02$ & $0.81 \pm 0.0$ & $0.35 \pm 0.03$\\
4.0 & $0.6 \pm 0.0$ & $0.65 \pm 0.02$ & $0.64 \pm 0.01$ & $0.81 \pm 0.0$ & $0.37 \pm 0.02$\\
5.0 & $0.6 \pm 0.0$ & $0.67 \pm 0.02$ & $0.66 \pm 0.01$ & $0.81 \pm 0.0$ & $0.42 \pm 0.01$\\
6.0 & $0.6 \pm 0.0$ & $0.67 \pm 0.02$ & $0.68 \pm 0.01$ & $0.81 \pm 0.0$ & $0.53 \pm 0.02$\\
7.0 & $0.6 \pm 0.0$ & $0.68 \pm 0.02$ & $0.69 \pm 0.01$ & $0.81 \pm 0.0$ & $0.66 \pm 0.01$\\
8.0 & $0.6 \pm 0.0$ & $0.68 \pm 0.02$ & $0.7 \pm 0.01$ & $0.81 \pm 0.0$ & $0.72 \pm 0.01$\\
9.0 & $0.6 \pm 0.0$ & $0.68 \pm 0.02$ & $0.7 \pm 0.01$ & $0.81 \pm 0.0$ & $0.76 \pm 0.01$\\
10.0 & $0.6 \pm 0.0$ & $0.69 \pm 0.02$ & $0.71 \pm 0.01$ & $0.81 \pm 0.0$ & $0.78 \pm 0.01$\\

\hline
\end{tabular}
}
\label{tab:cora-dp-ind}
\end{table}

\begin{table}[h!]
\centering
\caption{Node classification utility on Citeseer.}
\resizebox{\columnwidth}{!}{
\begin{tabular}{c c c c c c}
\toprule
Epsilon &  \mlp  & \toola & \toolb  & \gcn  &  \dpgcn \\\hline

1.0 & $0.6 \pm 0.01$ & $0.59 \pm 0.02$ & $0.49 \pm 0.04$ & $0.72 \pm 0.01$ & $0.37 \pm 0.02$\\
2.0 & $0.6 \pm 0.01$ & $0.61 \pm 0.01$ & $0.55 \pm 0.02$ & $0.72 \pm 0.01$ & $0.37 \pm 0.02$\\
3.0 & $0.6 \pm 0.01$ & $0.63 \pm 0.01$ & $0.59 \pm 0.02$ & $0.72 \pm 0.01$ & $0.37 \pm 0.02$\\
4.0 & $0.6 \pm 0.01$ & $0.64 \pm 0.01$ & $0.61 \pm 0.01$ & $0.72 \pm 0.01$ & $0.38 \pm 0.02$\\
5.0 & $0.6 \pm 0.01$ & $0.64 \pm 0.01$ & $0.63 \pm 0.01$ & $0.72 \pm 0.01$ & $0.4 \pm 0.02$\\
6.0 & $0.6 \pm 0.01$ & $0.64 \pm 0.01$ & $0.64 \pm 0.01$ & $0.72 \pm 0.01$ & $0.46 \pm 0.02$\\
7.0 & $0.6 \pm 0.01$ & $0.64 \pm 0.01$ & $0.64 \pm 0.01$ & $0.72 \pm 0.01$ & $0.55 \pm 0.02$\\
8.0 & $0.6 \pm 0.01$ & $0.64 \pm 0.01$ & $0.65 \pm 0.01$ & $0.72 \pm 0.01$ & $0.63 \pm 0.01$\\
9.0 & $0.6 \pm 0.01$ & $0.64 \pm 0.01$ & $0.65 \pm 0.01$ & $0.72 \pm 0.01$ & $0.67 \pm 0.01$\\
10.0 & $0.6 \pm 0.01$ & $0.65 \pm 0.01$ & $0.65 \pm 0.01$ & $0.72 \pm 0.01$ & $0.69 \pm 0.01$\\

\hline
\end{tabular}
}
\label{tab:citeseer-dp-ind}
\end{table}

\begin{table}[h!]
\centering
\caption{Node classification utility on Pubmed.}
\resizebox{\columnwidth}{!}{
\begin{tabular}{c c c c c c}
\toprule
Epsilon &  \mlp  & \toola & \toolb  & \gcn  &  \dpgcn \\\hline

1.0 & $0.73 \pm 0.01$ & $0.72 \pm 0.01$ & $0.69 \pm 0.02$ & $0.79 \pm 0.0$ & $0.55 \pm 0.02$\\
2.0 & $0.73 \pm 0.01$ & $0.73 \pm 0.01$ & $0.71 \pm 0.02$ & $0.79 \pm 0.0$ & $0.55 \pm 0.02$\\
3.0 & $0.73 \pm 0.01$ & $0.73 \pm 0.01$ & $0.72 \pm 0.01$ & $0.79 \pm 0.0$ & $0.55 \pm 0.02$\\
4.0 & $0.73 \pm 0.01$ & $0.74 \pm 0.01$ & $0.73 \pm 0.01$ & $0.79 \pm 0.0$ & $0.55 \pm 0.02$\\
5.0 & $0.73 \pm 0.01$ & $0.74 \pm 0.0$ & $0.74 \pm 0.01$ & $0.79 \pm 0.0$ & $0.55 \pm 0.02$\\
6.0 & $0.73 \pm 0.01$ & $0.74 \pm 0.01$ & $0.74 \pm 0.01$ & $0.79 \pm 0.0$ & $0.56 \pm 0.02$\\
7.0 & $0.73 \pm 0.01$ & $0.74 \pm 0.0$ & $0.74 \pm 0.01$ & $0.79 \pm 0.0$ & $0.59 \pm 0.01$\\
8.0 & $0.73 \pm 0.01$ & $0.74 \pm 0.0$ & $0.74 \pm 0.01$ & $0.79 \pm 0.0$ & $0.63 \pm 0.02$\\
9.0 & $0.73 \pm 0.01$ & $0.74 \pm 0.0$ & $0.74 \pm 0.01$ & $0.79 \pm 0.0$ & $0.69 \pm 0.02$\\
10.0 & $0.73 \pm 0.01$ & $0.74 \pm 0.0$ & $0.75 \pm 0.01$ & $0.79 \pm 0.0$ & $0.73 \pm 0.02$\\

\hline
\end{tabular}
}
\label{tab:pubmed-dp-ind}
\end{table}

\begin{table}[h!]
\centering
\caption{Node classification utility on Facebook.}
\resizebox{\columnwidth}{!}{
\begin{tabular}{c c c c c c}
\toprule
Epsilon &  \mlp  & \toola & \toolb  & \gcn  &  \dpgcn \\\hline

1.0 & $0.74 \pm 0.01$ & $0.78 \pm 0.01$ & $0.78 \pm 0.01$ & $0.93 \pm 0.0$ & $0.3 \pm 0.01$\\
2.0 & $0.74 \pm 0.01$ & $0.8 \pm 0.01$ & $0.8 \pm 0.01$ & $0.93 \pm 0.0$ & $0.3 \pm 0.01$\\
3.0 & $0.74 \pm 0.01$ & $0.81 \pm 0.01$ & $0.82 \pm 0.01$ & $0.93 \pm 0.0$ & $0.3 \pm 0.01$\\
4.0 & $0.74 \pm 0.01$ & $0.82 \pm 0.01$ & $0.83 \pm 0.01$ & $0.93 \pm 0.0$ & $0.3 \pm 0.01$\\
5.0 & $0.74 \pm 0.01$ & $0.82 \pm 0.01$ & $0.84 \pm 0.01$ & $0.93 \pm 0.0$ & $0.35 \pm 0.01$\\
6.0 & $0.74 \pm 0.01$ & $0.82 \pm 0.01$ & $0.84 \pm 0.01$ & $0.93 \pm 0.0$ & $0.47 \pm 0.03$\\
7.0 & $0.74 \pm 0.01$ & $0.82 \pm 0.01$ & $0.85 \pm 0.01$ & $0.93 \pm 0.0$ & $0.63 \pm 0.01$\\
8.0 & $0.74 \pm 0.01$ & $0.83 \pm 0.01$ & $0.85 \pm 0.01$ & $0.93 \pm 0.0$ & $0.74 \pm 0.01$\\
9.0 & $0.74 \pm 0.01$ & $0.83 \pm 0.01$ & $0.86 \pm 0.01$ & $0.93 \pm 0.0$ & $0.81 \pm 0.01$\\
10.0 & $0.74 \pm 0.01$ & $0.83 \pm 0.01$ & $0.86 \pm 0.01$ & $0.93 \pm 0.0$ & $0.85 \pm 0.01$\\

\hline
\end{tabular}
}
\label{tab:facebook-dp-ind}
\end{table}

\subsection{Inductive Setting}
\label{appdx:ind-appx}

\begin{table}[h!]
    \centering
    \caption{The performance of best attack against \tool and \baseline when both of them perform better than \mlp in the \flickr dataset. \tool is significantly more attack resilient at higher utility. In \twitch, we observe similar patterns.}
    \resizebox{\columnwidth}{!}{
    \begin{tabular}{c c c c c c}
        \toprule
        Nodes &  $\epsilon$ & \begin{tabular}[c]{@{}c@{}}F1-score\\  (\dpgcn)\end{tabular} & \begin{tabular}[c]{@{}c@{}}Best Attack\\  (\dpgcn)\end{tabular} & \begin{tabular}[c]{@{}c@{}}F1-score\\ (\toolb)\end{tabular} & \begin{tabular}[c]{@{}c@{}}Best Attack\\ (\toolb)\end{tabular}\\\hline
        \multirow{3}{*}{Low Degree} 
         & 9 & 0.47 & 0.74 & 0.49 & 0.45\\
         & 10 & 0.49 & 0.87 & 0.49 & 0.58\\\hline
        \multirow{3}{*}{All Degree} 
         & 9 & 0.47 & 0.81 & 0.49 & 0.55\\
         & 10 & 0.49 & 0.90 & 0.49 & 0.56\\\hline
         \multirow{3}{*}{High Degree} 
         & 9 & 0.47 & 0.82 & 0.49 & 0.64\\
         & 10 & 0.49 & 0.90 & 0.49 & 0.65\\\hline
    \end{tabular}
    }
    \label{tab:flickr}
\end{table}

First, we show the attack performance in configurations where both \tool and \baseline perform better than \mlp in Table~\ref{tab:flickr}.  

Finally, we report the full utility and attack results for the inductive setting. The Tables~\ref{tab:de-dp-ind},~\ref{tab:ptbr-dp-ind},~\ref{tab:engb-dp-ind},~\ref{tab:ru-dp-ind},~\ref{tab:fr-dp-ind} and~\ref{tab:flickr-dp-ind} report utility values. The attack performance of existing attacks is given in the Tables~\ref{tab:ind_1_attack_unbalanced} and~\ref{tab:ind_2_attack_unbalanced}. We also provide attack performances on nodes of high and low degrees to reconcile with the evaluation settings of the LinkTeller paper in Tables~\ref{tab:ind_1_attack_unbalanced_hi},~\ref{tab:ind_2_attack_unbalanced_hi},~\ref{tab:ind_1_attack_unbalanced_lo} and~\ref{tab:ind_2_attack_unbalanced_lo}. 

\begin{table}[h!]
\centering
\caption{Node classification utility on TwitchDE.}
\resizebox{\columnwidth}{!}{
\begin{tabular}{c c c c c c}
\toprule
Epsilon &  \mlp  & \toola & \toolb  & \gcn  &  \dpgcn \\\hline

1.0 & $0.56 \pm 0.0$ & $0.56 \pm 0.0$ & $0.56 \pm 0.0$ & $0.55 \pm 0.01$ & $0.43 \pm 0.04$\\
2.0 & $0.56 \pm 0.0$ & $0.56 \pm 0.0$ & $0.56 \pm 0.0$ & $0.55 \pm 0.01$ & $0.43 \pm 0.03$\\
3.0 & $0.56 \pm 0.0$ & $0.56 \pm 0.0$ & $0.56 \pm 0.0$ & $0.55 \pm 0.01$ & $0.44 \pm 0.03$\\
4.0 & $0.56 \pm 0.0$ & $0.56 \pm 0.0$ & $0.56 \pm 0.0$ & $0.55 \pm 0.01$ & $0.46 \pm 0.03$\\
5.0 & $0.56 \pm 0.0$ & $0.56 \pm 0.0$ & $0.56 \pm 0.0$ & $0.55 \pm 0.01$ & $0.48 \pm 0.03$\\
6.0 & $0.56 \pm 0.0$ & $0.56 \pm 0.0$ & $0.56 \pm 0.0$ & $0.55 \pm 0.01$ & $0.51 \pm 0.02$\\
7.0 & $0.56 \pm 0.0$ & $0.56 \pm 0.0$ & $0.56 \pm 0.0$ & $0.55 \pm 0.01$ & $0.53 \pm 0.02$\\
8.0 & $0.56 \pm 0.0$ & $0.56 \pm 0.0$ & $0.56 \pm 0.0$ & $0.55 \pm 0.01$ & $0.54 \pm 0.01$\\
9.0 & $0.56 \pm 0.0$ & $0.56 \pm 0.0$ & $0.56 \pm 0.0$ & $0.55 \pm 0.01$ & $0.55 \pm 0.01$\\
10.0 & $0.56 \pm 0.0$ & $0.56 \pm 0.0$ & $0.56 \pm 0.0$ & $0.55 \pm 0.01$ & $0.55 \pm 0.01$\\

\hline
\end{tabular}
}
\label{tab:de-dp-ind}
\end{table}

\begin{table}[h!]
\centering
\caption{Node classification utility on TwitchPTBR.}
\resizebox{\columnwidth}{!}{
\begin{tabular}{c c c c c c}
\toprule
Epsilon &  \mlp  & \toola & \toolb  & \gcn  &  \dpgcn \\\hline

1.0 & $0.35 \pm 0.01$ & $0.34 \pm 0.02$ & $0.34 \pm 0.02$ & $0.45 \pm 0.01$ & $0.44 \pm 0.03$\\
2.0 & $0.35 \pm 0.01$ & $0.34 \pm 0.02$ & $0.34 \pm 0.02$ & $0.45 \pm 0.01$ & $0.44 \pm 0.03$\\
3.0 & $0.35 \pm 0.01$ & $0.34 \pm 0.02$ & $0.34 \pm 0.02$ & $0.45 \pm 0.01$ & $0.44 \pm 0.03$\\
4.0 & $0.35 \pm 0.01$ & $0.34 \pm 0.02$ & $0.34 \pm 0.02$ & $0.45 \pm 0.01$ & $0.45 \pm 0.03$\\
5.0 & $0.35 \pm 0.01$ & $0.34 \pm 0.02$ & $0.34 \pm 0.02$ & $0.45 \pm 0.01$ & $0.45 \pm 0.03$\\
6.0 & $0.35 \pm 0.01$ & $0.34 \pm 0.02$ & $0.34 \pm 0.01$ & $0.45 \pm 0.01$ & $0.45 \pm 0.03$\\
7.0 & $0.35 \pm 0.01$ & $0.34 \pm 0.02$ & $0.34 \pm 0.02$ & $0.45 \pm 0.01$ & $0.46 \pm 0.02$\\
8.0 & $0.35 \pm 0.01$ & $0.34 \pm 0.01$ & $0.34 \pm 0.02$ & $0.45 \pm 0.01$ & $0.46 \pm 0.03$\\
9.0 & $0.35 \pm 0.01$ & $0.34 \pm 0.02$ & $0.34 \pm 0.02$ & $0.45 \pm 0.01$ & $0.45 \pm 0.03$\\
10.0 & $0.35 \pm 0.01$ & $0.34 \pm 0.02$ & $0.34 \pm 0.02$ & $0.45 \pm 0.01$ & $0.45 \pm 0.02$\\

\hline
\end{tabular}
}
\label{tab:ptbr-dp-ind}
\end{table}

\begin{table}[h!]
\centering
\caption{Node classification utility on TwitchENGB.}
\resizebox{\columnwidth}{!}{
\begin{tabular}{c c c c c c}
\toprule
Epsilon &  \mlp  & \toola & \toolb  & \gcn  &  \dpgcn \\\hline

1.0 & $0.6 \pm 0.0$ & $0.6 \pm 0.0$ & $0.6 \pm 0.0$ & $0.6 \pm 0.0$ & $0.6 \pm 0.01$\\
2.0 & $0.6 \pm 0.0$ & $0.6 \pm 0.0$ & $0.6 \pm 0.0$ & $0.6 \pm 0.0$ & $0.6 \pm 0.01$\\
3.0 & $0.6 \pm 0.0$ & $0.6 \pm 0.0$ & $0.6 \pm 0.0$ & $0.6 \pm 0.0$ & $0.6 \pm 0.01$\\
4.0 & $0.6 \pm 0.0$ & $0.6 \pm 0.0$ & $0.6 \pm 0.0$ & $0.6 \pm 0.0$ & $0.6 \pm 0.01$\\
5.0 & $0.6 \pm 0.0$ & $0.6 \pm 0.0$ & $0.6 \pm 0.0$ & $0.6 \pm 0.0$ & $0.6 \pm 0.01$\\
6.0 & $0.6 \pm 0.0$ & $0.6 \pm 0.0$ & $0.6 \pm 0.0$ & $0.6 \pm 0.0$ & $0.6 \pm 0.01$\\
7.0 & $0.6 \pm 0.0$ & $0.6 \pm 0.0$ & $0.6 \pm 0.0$ & $0.6 \pm 0.0$ & $0.6 \pm 0.01$\\
8.0 & $0.6 \pm 0.0$ & $0.6 \pm 0.0$ & $0.6 \pm 0.0$ & $0.6 \pm 0.0$ & $0.6 \pm 0.01$\\
9.0 & $0.6 \pm 0.0$ & $0.6 \pm 0.0$ & $0.6 \pm 0.0$ & $0.6 \pm 0.0$ & $0.6 \pm 0.01$\\
10.0 & $0.6 \pm 0.0$ & $0.6 \pm 0.0$ & $0.6 \pm 0.0$ & $0.6 \pm 0.0$ & $0.6 \pm 0.01$\\

\hline
\end{tabular}
}
\label{tab:engb-dp-ind}
\end{table}

\begin{table}[h!]
\centering
\caption{Node classification utility on TwitchRU.}
\resizebox{\columnwidth}{!}{
\begin{tabular}{c c c c c c}
\toprule
Epsilon &  \mlp  & \toola & \toolb  & \gcn  &  \dpgcn \\\hline

1.0 & $0.27 \pm 0.01$ & $0.27 \pm 0.01$ & $0.27 \pm 0.01$ & $0.31 \pm 0.01$ & $0.28 \pm 0.02$\\
2.0 & $0.27 \pm 0.01$ & $0.27 \pm 0.01$ & $0.27 \pm 0.01$ & $0.31 \pm 0.01$ & $0.28 \pm 0.02$\\
3.0 & $0.27 \pm 0.01$ & $0.27 \pm 0.01$ & $0.27 \pm 0.01$ & $0.31 \pm 0.01$ & $0.28 \pm 0.02$\\
4.0 & $0.27 \pm 0.01$ & $0.27 \pm 0.01$ & $0.27 \pm 0.01$ & $0.31 \pm 0.01$ & $0.29 \pm 0.02$\\
5.0 & $0.27 \pm 0.01$ & $0.27 \pm 0.01$ & $0.27 \pm 0.01$ & $0.31 \pm 0.01$ & $0.29 \pm 0.02$\\
6.0 & $0.27 \pm 0.01$ & $0.27 \pm 0.01$ & $0.27 \pm 0.01$ & $0.31 \pm 0.01$ & $0.29 \pm 0.02$\\
7.0 & $0.27 \pm 0.01$ & $0.27 \pm 0.01$ & $0.27 \pm 0.01$ & $0.31 \pm 0.01$ & $0.3 \pm 0.03$\\
8.0 & $0.27 \pm 0.01$ & $0.27 \pm 0.01$ & $0.27 \pm 0.01$ & $0.31 \pm 0.01$ & $0.3 \pm 0.02$\\
9.0 & $0.27 \pm 0.01$ & $0.27 \pm 0.01$ & $0.27 \pm 0.01$ & $0.31 \pm 0.01$ & $0.3 \pm 0.02$\\
10.0 & $0.27 \pm 0.01$ & $0.27 \pm 0.01$ & $0.27 \pm 0.01$ & $0.31 \pm 0.01$ & $0.3 \pm 0.02$\\

\hline
\end{tabular}
}
\label{tab:ru-dp-ind}
\end{table}

\begin{table}[h!]
\centering
\caption{Node classification utility on TwitchFR.}
\resizebox{\columnwidth}{!}{
\begin{tabular}{c c c c c c}
\toprule
Epsilon &  \mlp  & \toola & \toolb  & \gcn  &  \dpgcn \\\hline

1.0 & $0.36 \pm 0.01$ & $0.36 \pm 0.01$ & $0.36 \pm 0.01$ & $0.42 \pm 0.02$ & $0.4 \pm 0.07$\\
2.0 & $0.36 \pm 0.01$ & $0.36 \pm 0.01$ & $0.36 \pm 0.01$ & $0.42 \pm 0.02$ & $0.4 \pm 0.06$\\
3.0 & $0.36 \pm 0.01$ & $0.36 \pm 0.01$ & $0.36 \pm 0.01$ & $0.42 \pm 0.02$ & $0.37 \pm 0.07$\\
4.0 & $0.36 \pm 0.01$ & $0.36 \pm 0.01$ & $0.36 \pm 0.01$ & $0.42 \pm 0.02$ & $0.35 \pm 0.09$\\
5.0 & $0.36 \pm 0.01$ & $0.36 \pm 0.01$ & $0.36 \pm 0.01$ & $0.42 \pm 0.02$ & $0.33 \pm 0.09$\\
6.0 & $0.36 \pm 0.01$ & $0.36 \pm 0.01$ & $0.36 \pm 0.01$ & $0.42 \pm 0.02$ & $0.31 \pm 0.07$\\
7.0 & $0.36 \pm 0.01$ & $0.36 \pm 0.01$ & $0.36 \pm 0.01$ & $0.42 \pm 0.02$ & $0.33 \pm 0.08$\\
8.0 & $0.36 \pm 0.01$ & $0.36 \pm 0.01$ & $0.36 \pm 0.01$ & $0.42 \pm 0.02$ & $0.35 \pm 0.06$\\
9.0 & $0.36 \pm 0.01$ & $0.36 \pm 0.01$ & $0.36 \pm 0.01$ & $0.42 \pm 0.02$ & $0.36 \pm 0.06$\\
10.0 & $0.36 \pm 0.01$ & $0.36 \pm 0.01$ & $0.36 \pm 0.01$ & $0.42 \pm 0.02$ & $0.38 \pm 0.05$\\

\hline
\end{tabular}
}
\label{tab:fr-dp-ind}
\end{table}

\begin{table}[h!]
\centering
\caption{Node classification utility on Flickr.}
\resizebox{\columnwidth}{!}{
\begin{tabular}{c c c c c c}
\toprule
Epsilon &  \mlp  & \toola & \toolb  & \gcn  &  \dpgcn \\\hline

1.0 & $0.47 \pm 0.0$ & $0.47 \pm 0.0$ & $0.46 \pm 0.0$ & $0.51 \pm 0.0$ & $0.44 \pm 0.0$\\
2.0 & $0.47 \pm 0.0$ & $0.47 \pm 0.0$ & $0.47 \pm 0.0$ & $0.51 \pm 0.0$ & $0.44 \pm 0.0$\\
3.0 & $0.47 \pm 0.0$ & $0.47 \pm 0.0$ & $0.47 \pm 0.0$ & $0.51 \pm 0.0$ & $0.44 \pm 0.0$\\
4.0 & $0.47 \pm 0.0$ & $0.47 \pm 0.0$ & $0.47 \pm 0.0$ & $0.51 \pm 0.0$ & $0.44 \pm 0.0$\\
5.0 & $0.47 \pm 0.0$ & $0.47 \pm 0.0$ & $0.48 \pm 0.0$ & $0.51 \pm 0.0$ & $0.44 \pm 0.0$\\
6.0 & $0.47 \pm 0.0$ & $0.47 \pm 0.0$ & $0.48 \pm 0.0$ & $0.51 \pm 0.0$ & $0.44 \pm 0.0$\\
7.0 & $0.47 \pm 0.0$ & $0.47 \pm 0.0$ & $0.48 \pm 0.0$ & $0.51 \pm 0.0$ & $0.44 \pm 0.0$\\
8.0 & $0.47 \pm 0.0$ & $0.47 \pm 0.0$ & $0.48 \pm 0.0$ & $0.51 \pm 0.0$ & $0.45 \pm 0.0$\\
9.0 & $0.47 \pm 0.0$ & $0.47 \pm 0.0$ & $0.48 \pm 0.0$ & $0.51 \pm 0.0$ & $0.47 \pm 0.0$\\
10.0 & $0.47 \pm 0.0$ & $0.47 \pm 0.0$ & $0.49 \pm 0.0$ & $0.51 \pm 0.0$ & $0.49 \pm 0.0$\\

\hline
\end{tabular}
}
\label{tab:flickr-dp-ind}
\end{table}

\begin{table*}[th!]
\centering
\caption{Attack in transductive setting}
\resizebox{\textwidth}{!}{
\begin{tabular}{c c c c c c c c c c }
\toprule
Model & Epsilon & \multicolumn{2}{c}{Cora} & \multicolumn{2}{c}{Facebook} & \multicolumn{2}{c}{Citeseer} & \multicolumn{2}{c}{Pubmed}\\\hline
 & & \baselineAttack & \sotaAttack & \baselineAttack & \sotaAttack & \baselineAttack & \sotaAttack & \baselineAttack & \sotaAttack\\\hline

\mlp
& $\infty$ & $0.75 \pm 0.01$ & $0.5 \pm 0.0$ & $0.8 \pm 0.02$ & $0.5 \pm 0.0$ & $0.78 \pm 0.01$ & $0.5 \pm 0.0$ & $0.75 \pm 0.02$ & $0.5 \pm 0.0$\\\hline
\gcn
& $\infty$ & $0.94 \pm 0.0$ & $1.0 \pm 0.0$ & $0.87 \pm 0.01$ & $1.0 \pm 0.0$ & $0.97 \pm 0.0$ & $1.0 \pm 0.0$ & $0.85 \pm 0.0$ & $1.0 \pm 0.0$\\\hline
\toola
& $\infty$ & $0.88 \pm 0.01$ & $0.5 \pm 0.0$ & $0.84 \pm 0.01$ & $0.5 \pm 0.0$ & $0.86 \pm 0.01$ & $0.5 \pm 0.0$ & $0.8 \pm 0.01$ & $0.5 \pm 0.0$\\\hline
\toolb
& $\infty$ & $0.89 \pm 0.01$ & $0.5 \pm 0.0$ & $0.86 \pm 0.02$ & $0.5 \pm 0.0$ & $0.89 \pm 0.01$ & $0.5 \pm 0.0$ & $0.81 \pm 0.0$ & $0.5 \pm 0.0$\\\hline

\multirow{10}{*}{\dpgcn}
& $1.0$ & $0.59 \pm 0.01$ & $0.5 \pm 0.0$ & $0.5 \pm 0.01$ & $0.5 \pm 0.0$ & $0.62 \pm 0.02$ & $0.5 \pm 0.0$ & $0.59 \pm 0.01$ & $0.5 \pm 0.0$\\

& $2.0$ & $0.59 \pm 0.02$ & $0.5 \pm 0.0$ & $0.5 \pm 0.01$ & $0.5 \pm 0.0$ & $0.62 \pm 0.02$ & $0.5 \pm 0.0$ & $0.59 \pm 0.01$ & $0.5 \pm 0.0$\\

& $3.0$ & $0.6 \pm 0.02$ & $0.51 \pm 0.0$ & $0.5 \pm 0.02$ & $0.5 \pm 0.0$ & $0.62 \pm 0.02$ & $0.51 \pm 0.0$ & $0.59 \pm 0.01$ & $0.5 \pm 0.0$\\

& $4.0$ & $0.62 \pm 0.01$ & $0.53 \pm 0.0$ & $0.5 \pm 0.01$ & $0.51 \pm 0.0$ & $0.64 \pm 0.02$ & $0.52 \pm 0.0$ & $0.59 \pm 0.01$ & $0.5 \pm 0.0$\\

& $5.0$ & $0.66 \pm 0.02$ & $0.59 \pm 0.0$ & $0.54 \pm 0.01$ & $0.55 \pm 0.0$ & $0.67 \pm 0.02$ & $0.55 \pm 0.01$ & $0.6 \pm 0.02$ & $0.51 \pm 0.0$\\

& $6.0$ & $0.76 \pm 0.01$ & $0.69 \pm 0.01$ & $0.61 \pm 0.03$ & $0.66 \pm 0.01$ & $0.72 \pm 0.01$ & $0.63 \pm 0.01$ & $0.62 \pm 0.02$ & $0.54 \pm 0.0$\\

& $7.0$ & $0.85 \pm 0.01$ & $0.82 \pm 0.01$ & $0.71 \pm 0.02$ & $0.82 \pm 0.01$ & $0.8 \pm 0.01$ & $0.75 \pm 0.01$ & $0.66 \pm 0.01$ & $0.6 \pm 0.01$\\

& $8.0$ & $0.88 \pm 0.01$ & $0.9 \pm 0.0$ & $0.79 \pm 0.02$ & $0.92 \pm 0.01$ & $0.88 \pm 0.01$ & $0.85 \pm 0.01$ & $0.72 \pm 0.01$ & $0.7 \pm 0.01$\\

& $9.0$ & $0.9 \pm 0.01$ & $0.95 \pm 0.0$ & $0.82 \pm 0.01$ & $0.96 \pm 0.0$ & $0.91 \pm 0.01$ & $0.91 \pm 0.01$ & $0.79 \pm 0.0$ & $0.85 \pm 0.01$\\

& $10.0$ & $0.91 \pm 0.0$ & $0.97 \pm 0.0$ & $0.84 \pm 0.01$ & $0.98 \pm 0.0$ & $0.93 \pm 0.01$ & $0.95 \pm 0.01$ & $0.81 \pm 0.01$ & $0.89 \pm 0.01$\\
\hline

\multirow{10}{*}{\toola}
& $1.0$ & $0.73 \pm 0.02$ & $0.5 \pm 0.0$ & $0.82 \pm 0.01$ & $0.5 \pm 0.0$ & $0.77 \pm 0.02$ & $0.5 \pm 0.0$ & $0.76 \pm 0.02$ & $0.5 \pm 0.0$\\

& $2.0$ & $0.79 \pm 0.02$ & $0.5 \pm 0.0$ & $0.83 \pm 0.01$ & $0.5 \pm 0.0$ & $0.82 \pm 0.01$ & $0.5 \pm 0.0$ & $0.76 \pm 0.01$ & $0.5 \pm 0.0$\\

& $3.0$ & $0.82 \pm 0.02$ & $0.5 \pm 0.0$ & $0.84 \pm 0.01$ & $0.5 \pm 0.0$ & $0.83 \pm 0.01$ & $0.5 \pm 0.0$ & $0.78 \pm 0.01$ & $0.5 \pm 0.0$\\

& $4.0$ & $0.83 \pm 0.02$ & $0.5 \pm 0.0$ & $0.84 \pm 0.01$ & $0.5 \pm 0.0$ & $0.84 \pm 0.01$ & $0.5 \pm 0.0$ & $0.78 \pm 0.01$ & $0.5 \pm 0.0$\\

& $5.0$ & $0.84 \pm 0.01$ & $0.5 \pm 0.0$ & $0.84 \pm 0.01$ & $0.5 \pm 0.0$ & $0.84 \pm 0.01$ & $0.5 \pm 0.0$ & $0.78 \pm 0.01$ & $0.5 \pm 0.0$\\

& $6.0$ & $0.85 \pm 0.01$ & $0.5 \pm 0.0$ & $0.84 \pm 0.01$ & $0.5 \pm 0.0$ & $0.85 \pm 0.01$ & $0.5 \pm 0.0$ & $0.78 \pm 0.01$ & $0.5 \pm 0.0$\\

& $7.0$ & $0.85 \pm 0.01$ & $0.5 \pm 0.0$ & $0.84 \pm 0.01$ & $0.5 \pm 0.0$ & $0.85 \pm 0.01$ & $0.5 \pm 0.0$ & $0.78 \pm 0.01$ & $0.5 \pm 0.0$\\

& $8.0$ & $0.85 \pm 0.01$ & $0.5 \pm 0.0$ & $0.84 \pm 0.01$ & $0.5 \pm 0.0$ & $0.85 \pm 0.01$ & $0.5 \pm 0.0$ & $0.79 \pm 0.01$ & $0.5 \pm 0.0$\\

& $9.0$ & $0.86 \pm 0.01$ & $0.5 \pm 0.0$ & $0.84 \pm 0.01$ & $0.5 \pm 0.0$ & $0.85 \pm 0.01$ & $0.5 \pm 0.0$ & $0.79 \pm 0.01$ & $0.5 \pm 0.0$\\

& $10.0$ & $0.86 \pm 0.01$ & $0.5 \pm 0.0$ & $0.85 \pm 0.01$ & $0.5 \pm 0.0$ & $0.85 \pm 0.01$ & $0.5 \pm 0.0$ & $0.79 \pm 0.01$ & $0.5 \pm 0.0$\\
\hline

\multirow{10}{*}{\toolb}
& $1.0$ & $0.68 \pm 0.02$ & $0.5 \pm 0.0$ & $0.81 \pm 0.01$ & $0.5 \pm 0.0$ & $0.67 \pm 0.02$ & $0.5 \pm 0.0$ & $0.71 \pm 0.02$ & $0.5 \pm 0.0$\\

& $2.0$ & $0.74 \pm 0.01$ & $0.5 \pm 0.0$ & $0.82 \pm 0.01$ & $0.5 \pm 0.0$ & $0.76 \pm 0.02$ & $0.5 \pm 0.0$ & $0.74 \pm 0.02$ & $0.5 \pm 0.0$\\

& $3.0$ & $0.79 \pm 0.01$ & $0.5 \pm 0.0$ & $0.83 \pm 0.01$ & $0.5 \pm 0.0$ & $0.8 \pm 0.01$ & $0.5 \pm 0.0$ & $0.76 \pm 0.01$ & $0.5 \pm 0.0$\\

& $4.0$ & $0.82 \pm 0.01$ & $0.5 \pm 0.0$ & $0.84 \pm 0.01$ & $0.5 \pm 0.0$ & $0.83 \pm 0.01$ & $0.5 \pm 0.0$ & $0.77 \pm 0.01$ & $0.5 \pm 0.0$\\

& $5.0$ & $0.83 \pm 0.01$ & $0.5 \pm 0.0$ & $0.84 \pm 0.01$ & $0.5 \pm 0.0$ & $0.84 \pm 0.01$ & $0.5 \pm 0.0$ & $0.77 \pm 0.02$ & $0.5 \pm 0.0$\\

& $6.0$ & $0.84 \pm 0.02$ & $0.5 \pm 0.0$ & $0.84 \pm 0.01$ & $0.5 \pm 0.0$ & $0.84 \pm 0.01$ & $0.5 \pm 0.0$ & $0.78 \pm 0.01$ & $0.5 \pm 0.0$\\

& $7.0$ & $0.85 \pm 0.02$ & $0.5 \pm 0.0$ & $0.84 \pm 0.01$ & $0.5 \pm 0.0$ & $0.85 \pm 0.01$ & $0.5 \pm 0.0$ & $0.79 \pm 0.01$ & $0.5 \pm 0.0$\\

& $8.0$ & $0.85 \pm 0.01$ & $0.5 \pm 0.0$ & $0.84 \pm 0.01$ & $0.5 \pm 0.0$ & $0.86 \pm 0.01$ & $0.5 \pm 0.0$ & $0.79 \pm 0.01$ & $0.5 \pm 0.0$\\

& $9.0$ & $0.86 \pm 0.01$ & $0.5 \pm 0.0$ & $0.85 \pm 0.02$ & $0.5 \pm 0.0$ & $0.86 \pm 0.01$ & $0.5 \pm 0.0$ & $0.79 \pm 0.01$ & $0.5 \pm 0.0$\\

& $10.0$ & $0.86 \pm 0.01$ & $0.5 \pm 0.0$ & $0.85 \pm 0.01$ & $0.5 \pm 0.0$ & $0.86 \pm 0.01$ & $0.5 \pm 0.0$ & $0.79 \pm 0.01$ & $0.5 \pm 0.0$\\
\hline

\end{tabular}
}
\label{tab:trans_attack}
\end{table*}

\begin{table*}[th!]
\centering
\caption{Attack in inductive setting for nodes of all degrees}
\resizebox{\textwidth}{!}{
\begin{tabular}{c c c c c c c c }
\toprule
Model & Epsilon & \multicolumn{2}{c}{TwitchDE} & \multicolumn{2}{c}{TwitchPTBR} & \multicolumn{2}{c}{TwitchRU}\\\hline
 & & \baselineAttack & \sotaAttack & \baselineAttack & \sotaAttack & \baselineAttack & \sotaAttack\\\hline

\mlp
& $\infty$ & $0.5 \pm 0.01$ & $0.5 \pm 0.0$ & $0.45 \pm 0.01$ & $0.5 \pm 0.0$ & $0.49 \pm 0.03$ & $0.5 \pm 0.0$\\\hline
\gcn
& $\infty$ & $0.54 \pm 0.02$ & $1.0 \pm 0.0$ & $0.57 \pm 0.05$ & $0.99 \pm 0.0$ & $0.52 \pm 0.01$ & $0.99 \pm 0.0$\\\hline
\toola
& $\infty$ & $0.5 \pm 0.02$ & $0.5 \pm 0.0$ & $0.46 \pm 0.02$ & $0.5 \pm 0.0$ & $0.48 \pm 0.01$ & $0.5 \pm 0.0$\\\hline
\toolb
& $\infty$ & $0.5 \pm 0.01$ & $0.5 \pm 0.0$ & $0.45 \pm 0.03$ & $0.5 \pm 0.0$ & $0.48 \pm 0.01$ & $0.5 \pm 0.0$\\\hline

\multirow{10}{*}{\dpgcn}
& $1.0$ & $0.48 \pm 0.02$ & $0.54 \pm 0.02$ & $0.52 \pm 0.02$ & $0.56 \pm 0.01$ & $0.49 \pm 0.02$ & $0.51 \pm 0.03$\\

& $2.0$ & $0.5 \pm 0.03$ & $0.57 \pm 0.01$ & $0.52 \pm 0.02$ & $0.62 \pm 0.0$ & $0.49 \pm 0.02$ & $0.53 \pm 0.04$\\

& $3.0$ & $0.5 \pm 0.02$ & $0.6 \pm 0.03$ & $0.53 \pm 0.02$ & $0.74 \pm 0.02$ & $0.5 \pm 0.02$ & $0.6 \pm 0.04$\\

& $4.0$ & $0.52 \pm 0.01$ & $0.71 \pm 0.01$ & $0.54 \pm 0.03$ & $0.85 \pm 0.01$ & $0.52 \pm 0.02$ & $0.7 \pm 0.03$\\

& $5.0$ & $0.51 \pm 0.02$ & $0.8 \pm 0.01$ & $0.55 \pm 0.04$ & $0.93 \pm 0.01$ & $0.49 \pm 0.02$ & $0.82 \pm 0.02$\\

& $6.0$ & $0.52 \pm 0.02$ & $0.91 \pm 0.01$ & $0.54 \pm 0.03$ & $0.96 \pm 0.01$ & $0.52 \pm 0.02$ & $0.91 \pm 0.01$\\

& $7.0$ & $0.53 \pm 0.02$ & $0.96 \pm 0.01$ & $0.55 \pm 0.01$ & $0.95 \pm 0.0$ & $0.53 \pm 0.02$ & $0.95 \pm 0.01$\\

& $8.0$ & $0.52 \pm 0.02$ & $0.97 \pm 0.0$ & $0.57 \pm 0.02$ & $0.98 \pm 0.0$ & $0.54 \pm 0.01$ & $0.97 \pm 0.01$\\

& $9.0$ & $0.55 \pm 0.02$ & $0.98 \pm 0.0$ & $0.55 \pm 0.03$ & $0.98 \pm 0.0$ & $0.52 \pm 0.02$ & $0.98 \pm 0.01$\\

& $10.0$ & $0.54 \pm 0.01$ & $0.99 \pm 0.0$ & $0.57 \pm 0.03$ & $0.98 \pm 0.0$ & $0.52 \pm 0.01$ & $0.99 \pm 0.0$\\
\hline

\multirow{10}{*}{\toola}
& $1.0$ & $0.49 \pm 0.01$ & $0.5 \pm 0.0$ & $0.47 \pm 0.04$ & $0.5 \pm 0.0$ & $0.51 \pm 0.02$ & $0.5 \pm 0.0$\\

& $2.0$ & $0.5 \pm 0.01$ & $0.5 \pm 0.0$ & $0.47 \pm 0.04$ & $0.5 \pm 0.0$ & $0.51 \pm 0.01$ & $0.5 \pm 0.0$\\

& $3.0$ & $0.48 \pm 0.02$ & $0.5 \pm 0.0$ & $0.46 \pm 0.02$ & $0.5 \pm 0.0$ & $0.51 \pm 0.01$ & $0.5 \pm 0.0$\\

& $4.0$ & $0.5 \pm 0.01$ & $0.5 \pm 0.0$ & $0.46 \pm 0.02$ & $0.5 \pm 0.0$ & $0.51 \pm 0.01$ & $0.5 \pm 0.0$\\

& $5.0$ & $0.49 \pm 0.01$ & $0.5 \pm 0.0$ & $0.47 \pm 0.01$ & $0.5 \pm 0.0$ & $0.51 \pm 0.01$ & $0.5 \pm 0.0$\\

& $6.0$ & $0.5 \pm 0.01$ & $0.5 \pm 0.0$ & $0.47 \pm 0.03$ & $0.5 \pm 0.0$ & $0.51 \pm 0.02$ & $0.5 \pm 0.0$\\

& $7.0$ & $0.49 \pm 0.02$ & $0.5 \pm 0.0$ & $0.48 \pm 0.02$ & $0.5 \pm 0.0$ & $0.51 \pm 0.01$ & $0.5 \pm 0.0$\\

& $8.0$ & $0.49 \pm 0.03$ & $0.5 \pm 0.0$ & $0.46 \pm 0.03$ & $0.5 \pm 0.0$ & $0.5 \pm 0.02$ & $0.5 \pm 0.0$\\

& $9.0$ & $0.49 \pm 0.01$ & $0.5 \pm 0.0$ & $0.47 \pm 0.04$ & $0.5 \pm 0.0$ & $0.51 \pm 0.01$ & $0.5 \pm 0.0$\\

& $10.0$ & $0.48 \pm 0.02$ & $0.5 \pm 0.0$ & $0.45 \pm 0.02$ & $0.5 \pm 0.0$ & $0.51 \pm 0.01$ & $0.5 \pm 0.0$\\
\hline

\multirow{10}{*}{\toolb}
& $1.0$ & $0.49 \pm 0.02$ & $0.5 \pm 0.0$ & $0.46 \pm 0.02$ & $0.5 \pm 0.0$ & $0.5 \pm 0.01$ & $0.5 \pm 0.0$\\

& $2.0$ & $0.5 \pm 0.01$ & $0.5 \pm 0.0$ & $0.46 \pm 0.03$ & $0.5 \pm 0.0$ & $0.51 \pm 0.01$ & $0.5 \pm 0.0$\\

& $3.0$ & $0.5 \pm 0.02$ & $0.5 \pm 0.0$ & $0.46 \pm 0.03$ & $0.5 \pm 0.0$ & $0.51 \pm 0.02$ & $0.5 \pm 0.0$\\

& $4.0$ & $0.5 \pm 0.01$ & $0.5 \pm 0.0$ & $0.47 \pm 0.02$ & $0.5 \pm 0.0$ & $0.51 \pm 0.01$ & $0.5 \pm 0.0$\\

& $5.0$ & $0.5 \pm 0.01$ & $0.5 \pm 0.0$ & $0.46 \pm 0.02$ & $0.5 \pm 0.0$ & $0.51 \pm 0.01$ & $0.5 \pm 0.0$\\

& $6.0$ & $0.5 \pm 0.02$ & $0.5 \pm 0.0$ & $0.45 \pm 0.03$ & $0.5 \pm 0.0$ & $0.51 \pm 0.01$ & $0.5 \pm 0.0$\\

& $7.0$ & $0.49 \pm 0.02$ & $0.5 \pm 0.0$ & $0.45 \pm 0.03$ & $0.5 \pm 0.0$ & $0.51 \pm 0.01$ & $0.5 \pm 0.0$\\

& $8.0$ & $0.5 \pm 0.02$ & $0.5 \pm 0.0$ & $0.46 \pm 0.03$ & $0.5 \pm 0.0$ & $0.51 \pm 0.02$ & $0.5 \pm 0.0$\\

& $9.0$ & $0.5 \pm 0.02$ & $0.5 \pm 0.0$ & $0.45 \pm 0.04$ & $0.5 \pm 0.0$ & $0.51 \pm 0.02$ & $0.5 \pm 0.0$\\

& $10.0$ & $0.5 \pm 0.02$ & $0.5 \pm 0.0$ & $0.47 \pm 0.02$ & $0.5 \pm 0.0$ & $0.51 \pm 0.02$ & $0.5 \pm 0.0$\\
\hline

\end{tabular}
}
\label{tab:ind_1_attack_unbalanced}
\end{table*}

\begin{table*}[th!]
\centering
\caption{Attack in inductive setting for nodes of all degrees}
\resizebox{\textwidth}{!}{
\begin{tabular}{c c c c c c c c }
\toprule
Model & Epsilon & \multicolumn{2}{c}{TwitchENGB} & \multicolumn{2}{c}{TwitchFR} & \multicolumn{2}{c}{Flickr}\\\hline
 & & \baselineAttack & \sotaAttack & \baselineAttack & \sotaAttack & \baselineAttack & \sotaAttack\\\hline

\mlp
& $\infty$ & $0.47 \pm 0.02$ & $0.5 \pm 0.0$ & $0.48 \pm 0.02$ & $0.5 \pm 0.0$ & $0.5 \pm 0.05$ & $0.5 \pm 0.0$\\\hline
\gcn
& $\infty$ & $0.54 \pm 0.02$ & $1.0 \pm 0.0$ & $0.52 \pm 0.02$ & $0.99 \pm 0.0$ & $0.61 \pm 0.07$ & $1.0 \pm 0.0$\\\hline
\toola
& $\infty$ & $0.47 \pm 0.03$ & $0.5 \pm 0.0$ & $0.49 \pm 0.04$ & $0.5 \pm 0.0$ & $0.54 \pm 0.08$ & $0.5 \pm 0.0$\\\hline
\toolb
& $\infty$ & $0.49 \pm 0.03$ & $0.5 \pm 0.0$ & $0.49 \pm 0.02$ & $0.5 \pm 0.0$ & $0.59 \pm 0.09$ & $0.5 \pm 0.0$\\\hline

\multirow{10}{*}{\dpgcn}
& $1.0$ & $0.5 \pm 0.02$ & $0.5 \pm 0.01$ & $0.51 \pm 0.01$ & $0.53 \pm 0.03$ & $0.49 \pm 0.06$ & $0.5 \pm 0.0$\\

& $2.0$ & $0.48 \pm 0.01$ & $0.5 \pm 0.01$ & $0.53 \pm 0.02$ & $0.57 \pm 0.03$ & $0.48 \pm 0.04$ & $0.5 \pm 0.0$\\

& $3.0$ & $0.47 \pm 0.02$ & $0.51 \pm 0.02$ & $0.51 \pm 0.03$ & $0.64 \pm 0.03$ & $0.48 \pm 0.04$ & $0.5 \pm 0.0$\\

& $4.0$ & $0.48 \pm 0.05$ & $0.55 \pm 0.03$ & $0.51 \pm 0.01$ & $0.74 \pm 0.03$ & $0.48 \pm 0.04$ & $0.5 \pm 0.0$\\

& $5.0$ & $0.48 \pm 0.01$ & $0.66 \pm 0.02$ & $0.52 \pm 0.02$ & $0.86 \pm 0.02$ & $0.47 \pm 0.07$ & $0.5 \pm 0.0$\\

& $6.0$ & $0.5 \pm 0.03$ & $0.79 \pm 0.02$ & $0.52 \pm 0.04$ & $0.92 \pm 0.01$ & $0.50 \pm 0.05$ & $0.51 \pm 0.01$\\

& $7.0$ & $0.5 \pm 0.04$ & $0.91 \pm 0.01$ & $0.52 \pm 0.02$ & $0.96 \pm 0.01$ & $0.52 \pm 0.01$ & $0.57 \pm 0.03$\\

& $8.0$ & $0.49 \pm 0.03$ & $0.95 \pm 0.0$ & $0.51 \pm 0.02$ & $0.97 \pm 0.0$ & $0.50 \pm 0.08$ & $0.62 \pm 0.04$\\

& $9.0$ & $0.52 \pm 0.02$ & $0.97 \pm 0.01$ & $0.52 \pm 0.02$ & $0.98 \pm 0.0$ & $0.52 \pm 0.07$ & $0.81 \pm 0.04$\\

& $10.0$ & $0.53 \pm 0.04$ & $0.98 \pm 0.01$ & $0.53 \pm 0.01$ & $0.99 \pm 0.0$ & $0.54 \pm 0.02$ & $0.90 \pm 0.02$\\
\hline

\multirow{10}{*}{\toola}
& $1.0$ & $0.47 \pm 0.02$ & $0.5 \pm 0.0$ & $0.48 \pm 0.02$ & $0.5 \pm 0.0$ & $0.52 \pm 0.03$ & $0.5 \pm 0.0$\\

& $2.0$ & $0.47 \pm 0.02$ & $0.5 \pm 0.0$ & $0.48 \pm 0.01$ & $0.5 \pm 0.0$ & $0.53 \pm 0.08$ & $0.5 \pm 0.0$\\

& $3.0$ & $0.49 \pm 0.02$ & $0.5 \pm 0.0$ & $0.46 \pm 0.03$ & $0.5 \pm 0.0$ & $0.55 \pm 0.06$ & $0.5 \pm 0.0$\\

& $4.0$ & $0.48 \pm 0.02$ & $0.5 \pm 0.0$ & $0.47 \pm 0.03$ & $0.5 \pm 0.0$ & $0.54 \pm 0.08$ & $0.5 \pm 0.0$\\

& $5.0$ & $0.48 \pm 0.02$ & $0.5 \pm 0.0$ & $0.47 \pm 0.03$ & $0.5 \pm 0.0$ & $0.56 \pm 0.07$ & $0.5 \pm 0.0$\\

& $6.0$ & $0.48 \pm 0.02$ & $0.5 \pm 0.0$ & $0.48 \pm 0.02$ & $0.5 \pm 0.0$ & $0.54 \pm 0.07$ & $0.5 \pm 0.0$\\

& $7.0$ & $0.46 \pm 0.03$ & $0.5 \pm 0.0$ & $0.47 \pm 0.03$ & $0.5 \pm 0.0$ & $0.54 \pm 0.07$ & $0.5 \pm 0.0$\\

& $8.0$ & $0.48 \pm 0.03$ & $0.5 \pm 0.0$ & $0.46 \pm 0.03$ & $0.5 \pm 0.0$ & $0.55 \pm 0.06$ & $0.5 \pm 0.0$\\

& $9.0$ & $0.48 \pm 0.03$ & $0.5 \pm 0.0$ & $0.47 \pm 0.01$ & $0.5 \pm 0.0$ & $0.54 \pm 0.07$ & $0.5 \pm 0.0$\\

& $10.0$ & $0.48 \pm 0.03$ & $0.5 \pm 0.0$ & $0.46 \pm 0.03$ & $0.5 \pm 0.0$ & $0.54 \pm 0.07$ & $0.5 \pm 0.0$\\
\hline

\multirow{10}{*}{\toolb}
& $1.0$ & $0.47 \pm 0.03$ & $0.5 \pm 0.0$ & $0.48 \pm 0.02$ & $0.5 \pm 0.0$ & $0.48 \pm 0.03$ & $0.5 \pm 0.0$\\

& $2.0$ & $0.47 \pm 0.02$ & $0.5 \pm 0.0$ & $0.48 \pm 0.02$ & $0.5 \pm 0.0$ & $0.52 \pm 0.04$ & $0.5 \pm 0.0$\\

& $3.0$ & $0.49 \pm 0.04$ & $0.5 \pm 0.0$ & $0.47 \pm 0.02$ & $0.5 \pm 0.0$ & $0.52 \pm 0.08$ & $0.5 \pm 0.0$\\

& $4.0$ & $0.48 \pm 0.02$ & $0.5 \pm 0.0$ & $0.49 \pm 0.02$ & $0.5 \pm 0.0$ & $0.53 \pm 0.09$ & $0.5 \pm 0.0$\\

& $5.0$ & $0.48 \pm 0.03$ & $0.5 \pm 0.0$ & $0.49 \pm 0.0$ & $0.5 \pm 0.0$ & $0.53 \pm 0.08$ & $0.5 \pm 0.0$\\

& $6.0$ & $0.47 \pm 0.02$ & $0.5 \pm 0.0$ & $0.48 \pm 0.02$ & $0.5 \pm 0.0$ & $0.54 \pm 0.05$ & $0.5 \pm 0.0$\\

& $7.0$ & $0.48 \pm 0.02$ & $0.5 \pm 0.0$ & $0.49 \pm 0.01$ & $0.5 \pm 0.0$ & $0.58 \pm 0.05$ & $0.5 \pm 0.0$\\

& $8.0$ & $0.48 \pm 0.02$ & $0.5 \pm 0.0$ & $0.45 \pm 0.04$ & $0.5 \pm 0.0$ & $0.55 \pm 0.07$ & $0.5 \pm 0.0$\\

& $9.0$ & $0.48 \pm 0.02$ & $0.5 \pm 0.0$ & $0.48 \pm 0.02$ & $0.5 \pm 0.0$ & $0.55 \pm 0.09$ & $0.5 \pm 0.0$\\

& $10.0$ & $0.47 \pm 0.02$ & $0.5 \pm 0.0$ & $0.48 \pm 0.02$ & $0.5 \pm 0.0$ & $0.56 \pm 0.07$ & $0.5 \pm 0.0$\\
\hline

\end{tabular}
}
\label{tab:ind_2_attack_unbalanced}
\end{table*}

\begin{table*}[th!]
\centering
\caption{Attack in inductive setting for high degree nodes}
\resizebox{\textwidth}{!}{
\begin{tabular}{c c c c c c c c }
\toprule
Model & Epsilon & \multicolumn{2}{c}{TwitchDE} & \multicolumn{2}{c}{TwitchPTBR} & \multicolumn{2}{c}{TwitchRU}\\\hline
 & & \baselineAttack & \sotaAttack & \baselineAttack & \sotaAttack & \baselineAttack & \sotaAttack\\\hline

\mlp
& $\infty$ & $0.51 \pm 0.02$ & $0.5 \pm 0.0$ & $0.48 \pm 0.01$ & $0.5 \pm 0.0$ & $0.52 \pm 0.01$ & $0.5 \pm 0.0$\\\hline
\gcn
& $\infty$ & $0.53 \pm 0.01$ & $0.99 \pm 0.0$ & $0.54 \pm 0.01$ & $0.98 \pm 0.0$ & $0.53 \pm 0.01$ & $0.99 \pm 0.0$\\\hline
\toola
& $\infty$ & $0.52 \pm 0.02$ & $0.5 \pm 0.0$ & $0.49 \pm 0.01$ & $0.5 \pm 0.0$ & $0.52 \pm 0.02$ & $0.5 \pm 0.0$\\\hline
\toolb
& $\infty$ & $0.52 \pm 0.02$ & $0.5 \pm 0.0$ & $0.48 \pm 0.03$ & $0.5 \pm 0.0$ & $0.52 \pm 0.02$ & $0.5 \pm 0.0$\\\hline

\multirow{10}{*}{\dpgcn}
& $1.0$ & $0.47 \pm 0.02$ & $0.52 \pm 0.02$ & $0.51 \pm 0.01$ & $0.56 \pm 0.01$ & $0.5 \pm 0.01$ & $0.53 \pm 0.01$\\

& $2.0$ & $0.49 \pm 0.02$ & $0.55 \pm 0.01$ & $0.52 \pm 0.01$ & $0.61 \pm 0.01$ & $0.52 \pm 0.01$ & $0.55 \pm 0.01$\\

& $3.0$ & $0.49 \pm 0.01$ & $0.59 \pm 0.01$ & $0.52 \pm 0.02$ & $0.73 \pm 0.0$ & $0.52 \pm 0.01$ & $0.61 \pm 0.02$\\

& $4.0$ & $0.52 \pm 0.02$ & $0.69 \pm 0.01$ & $0.53 \pm 0.02$ & $0.84 \pm 0.0$ & $0.51 \pm 0.01$ & $0.69 \pm 0.02$\\

& $5.0$ & $0.53 \pm 0.02$ & $0.8 \pm 0.01$ & $0.54 \pm 0.03$ & $0.91 \pm 0.0$ & $0.53 \pm 0.01$ & $0.81 \pm 0.01$\\

& $6.0$ & $0.54 \pm 0.02$ & $0.9 \pm 0.0$ & $0.54 \pm 0.02$ & $0.94 \pm 0.0$ & $0.54 \pm 0.01$ & $0.9 \pm 0.01$\\

& $7.0$ & $0.54 \pm 0.02$ & $0.94 \pm 0.0$ & $0.55 \pm 0.02$ & $0.93 \pm 0.01$ & $0.54 \pm 0.02$ & $0.93 \pm 0.01$\\

& $8.0$ & $0.53 \pm 0.01$ & $0.97 \pm 0.0$ & $0.55 \pm 0.02$ & $0.96 \pm 0.0$ & $0.53 \pm 0.02$ & $0.96 \pm 0.0$\\

& $9.0$ & $0.53 \pm 0.0$ & $0.98 \pm 0.0$ & $0.55 \pm 0.01$ & $0.97 \pm 0.0$ & $0.52 \pm 0.01$ & $0.97 \pm 0.0$\\

& $10.0$ & $0.54 \pm 0.02$ & $0.98 \pm 0.0$ & $0.55 \pm 0.01$ & $0.97 \pm 0.0$ & $0.53 \pm 0.01$ & $0.98 \pm 0.0$\\
\hline

\multirow{10}{*}{\toola}
& $1.0$ & $0.52 \pm 0.02$ & $0.5 \pm 0.0$ & $0.47 \pm 0.01$ & $0.5 \pm 0.0$ & $0.51 \pm 0.01$ & $0.5 \pm 0.0$\\

& $2.0$ & $0.52 \pm 0.02$ & $0.5 \pm 0.0$ & $0.49 \pm 0.01$ & $0.5 \pm 0.0$ & $0.51 \pm 0.01$ & $0.5 \pm 0.0$\\

& $3.0$ & $0.52 \pm 0.02$ & $0.5 \pm 0.0$ & $0.48 \pm 0.01$ & $0.5 \pm 0.0$ & $0.51 \pm 0.01$ & $0.5 \pm 0.0$\\

& $4.0$ & $0.52 \pm 0.01$ & $0.5 \pm 0.0$ & $0.47 \pm 0.01$ & $0.5 \pm 0.0$ & $0.51 \pm 0.0$ & $0.5 \pm 0.0$\\

& $5.0$ & $0.51 \pm 0.01$ & $0.5 \pm 0.0$ & $0.48 \pm 0.01$ & $0.5 \pm 0.0$ & $0.51 \pm 0.01$ & $0.5 \pm 0.0$\\

& $6.0$ & $0.51 \pm 0.02$ & $0.5 \pm 0.0$ & $0.48 \pm 0.01$ & $0.5 \pm 0.0$ & $0.51 \pm 0.0$ & $0.5 \pm 0.0$\\

& $7.0$ & $0.52 \pm 0.02$ & $0.5 \pm 0.0$ & $0.48 \pm 0.01$ & $0.5 \pm 0.0$ & $0.51 \pm 0.01$ & $0.5 \pm 0.0$\\

& $8.0$ & $0.51 \pm 0.02$ & $0.5 \pm 0.0$ & $0.48 \pm 0.01$ & $0.5 \pm 0.0$ & $0.51 \pm 0.01$ & $0.5 \pm 0.0$\\

& $9.0$ & $0.52 \pm 0.02$ & $0.5 \pm 0.0$ & $0.48 \pm 0.0$ & $0.5 \pm 0.0$ & $0.51 \pm 0.01$ & $0.5 \pm 0.0$\\

& $10.0$ & $0.52 \pm 0.01$ & $0.5 \pm 0.0$ & $0.48 \pm 0.01$ & $0.5 \pm 0.0$ & $0.51 \pm 0.01$ & $0.5 \pm 0.0$\\
\hline

\multirow{10}{*}{\toolb}
& $1.0$ & $0.52 \pm 0.02$ & $0.5 \pm 0.0$ & $0.49 \pm 0.01$ & $0.5 \pm 0.0$ & $0.5 \pm 0.01$ & $0.5 \pm 0.0$\\

& $2.0$ & $0.5 \pm 0.5$ & $0.5 \pm 0.0$ & $0.49 \pm 0.01$ & $0.5 \pm 0.0$ & $0.52 \pm 0.01$ & $0.5 \pm 0.0$\\

& $3.0$ & $0.52 \pm 0.02$ & $0.5 \pm 0.0$ & $0.48 \pm 0.02$ & $0.5 \pm 0.0$ & $0.5 \pm 0.01$ & $0.5 \pm 0.0$\\

& $4.0$ & $0.53 \pm 0.01$ & $0.5 \pm 0.0$ & $0.49 \pm 0.01$ & $0.5 \pm 0.0$ & $0.51 \pm 0.01$ & $0.5 \pm 0.0$\\

& $5.0$ & $0.52 \pm 0.02$ & $0.5 \pm 0.0$ & $0.49 \pm 0.01$ & $0.5 \pm 0.0$ & $0.51 \pm 0.01$ & $0.5 \pm 0.0$\\

& $6.0$ & $0.53 \pm 0.03$ & $0.5 \pm 0.0$ & $0.49 \pm 0.01$ & $0.5 \pm 0.0$ & $0.51 \pm 0.01$ & $0.5 \pm 0.0$\\

& $7.0$ & $0.53 \pm 0.02$ & $0.5 \pm 0.0$ & $0.49 \pm 0.02$ & $0.5 \pm 0.0$ & $0.51 \pm 0.01$ & $0.5 \pm 0.0$\\

& $8.0$ & $0.52 \pm 0.02$ & $0.5 \pm 0.0$ & $0.47 \pm 0.01$ & $0.5 \pm 0.0$ & $0.51 \pm 0.01$ & $0.5 \pm 0.0$\\

& $9.0$ & $0.52 \pm 0.02$ & $0.5 \pm 0.0$ & $0.48 \pm 0.01$ & $0.5 \pm 0.0$ & $0.52 \pm 0.01$ & $0.5 \pm 0.0$\\

& $10.0$ & $0.52 \pm 0.02$ & $0.5 \pm 0.0$ & $0.49 \pm 0.01$ & $0.5 \pm 0.0$ & $0.51 \pm 0.01$ & $0.5 \pm 0.0$\\
\hline

\end{tabular}
}
\label{tab:ind_1_attack_unbalanced_hi}
\end{table*}

\begin{table*}[th!]
\centering
\caption{Attack in inductive setting for high degree nodes}
\resizebox{\textwidth}{!}{
\begin{tabular}{c c c c c c c c }
\toprule
Model & Epsilon & \multicolumn{2}{c}{TwitchENGB} & \multicolumn{2}{c}{TwitchFR} & \multicolumn{2}{c}{Flickr}\\\hline
 & & \baselineAttack & \sotaAttack & \baselineAttack & \sotaAttack & \baselineAttack & \sotaAttack\\\hline

\mlp
& $\infty$ & $0.51 \pm 0.01$ & $0.5 \pm 0.0$ & $0.49 \pm 0.01$ & $0.5 \pm 0.0$ & $0.53 \pm 0.01$ & $0.5 \pm 0.0$\\\hline
\gcn
& $\infty$ & $0.54 \pm 0.02$ & $0.99 \pm 0.0$ & $0.52 \pm 0.01$ & $0.99 \pm 0.0$ & $0.66 \pm 0.02$ & $0.98 \pm 0.0$\\\hline
\toola
& $\infty$ & $0.52 \pm 0.01$ & $0.5 \pm 0.0$ & $0.52 \pm 0.03$ & $0.5 \pm 0.0$ & $0.63 \pm 0.02$ & $0.5 \pm 0.0$\\\hline
\toolb
& $\infty$ & $0.52 \pm 0.01$ & $0.5 \pm 0.0$ & $0.51 \pm 0.02$ & $0.5 \pm 0.0$ & $0.65 \pm 0.03$ & $0.5 \pm 0.0$\\\hline

\multirow{10}{*}{\dpgcn}
& $1.0$ & $0.5 \pm 0.01$ & $0.5 \pm 0.0$ & $0.5 \pm 0.01$ & $0.54 \pm 0.01$ & $0.51 \pm 0.02$ & $0.5 \pm 0.0$\\

& $2.0$ & $0.5 \pm 0.01$ & $0.5 \pm 0.0$ & $0.51 \pm 0.01$ & $0.58 \pm 0.02$ & $0.51 \pm 0.02$ & $0.5 \pm 0.0$\\

& $3.0$ & $0.5 \pm 0.01$ & $0.52 \pm 0.0$ & $0.52 \pm 0.02$ & $0.65 \pm 0.03$ & $0.51 \pm 0.02$ & $0.5 \pm 0.0$\\

& $4.0$ & $0.51 \pm 0.01$ & $0.56 \pm 0.01$ & $0.51 \pm 0.01$ & $0.74 \pm 0.03$ & $0.51 \pm 0.02$ & $0.5 \pm 0.0$\\

& $5.0$ & $0.51 \pm 0.02$ & $0.66 \pm 0.01$ & $0.52 \pm 0.01$ & $0.85 \pm 0.02$ & $0.51 \pm 0.01$ & $0.51 \pm 0.0$\\

& $6.0$ & $0.51 \pm 0.02$ & $0.81 \pm 0.01$ & $0.52 \pm 0.01$ & $0.91 \pm 0.01$ & $0.52 \pm 0.02$ & $0.53 \pm 0.00$\\

& $7.0$ & $0.53 \pm 0.01$ & $0.91 \pm 0.01$ & $0.53 \pm 0.02$ & $0.95 \pm 0.0$ & $0.54 \pm 0.01$ & $0.58 \pm 0.01$\\

& $8.0$ & $0.54 \pm 0.01$ & $0.94 \pm 0.0$ & $0.53 \pm 0.01$ & $0.97 \pm 0.0$ & $0.56 \pm 0.02$ & $0.68 \pm 0.02$\\

& $9.0$ & $0.53 \pm 0.01$ & $0.97 \pm 0.0$ & $0.53 \pm 0.02$ & $0.98 \pm 0.0$ & $0.60 \pm 0.01$ & $0.82 \pm 0.02$\\

& $10.0$ & $0.53 \pm 0.01$ & $0.98 \pm 0.0$ & $0.53 \pm 0.01$ & $0.98 \pm 0.0$ & $0.64 \pm 0.02$ & $0.90 \pm 0.02$\\
\hline

\multirow{10}{*}{\toola}
& $1.0$ & $0.52 \pm 0.0$ & $0.5 \pm 0.0$ & $0.5 \pm 0.01$ & $0.5 \pm 0.0$ & $0.59 \pm 0.01$ & $0.5 \pm 0.0$\\

& $2.0$ & $0.52 \pm 0.01$ & $0.5 \pm 0.0$ & $0.5 \pm 0.01$ & $0.5 \pm 0.0$ & $0.61 \pm 0.02$ & $0.5 \pm 0.0$\\

& $3.0$ & $0.52 \pm 0.01$ & $0.5 \pm 0.0$ & $0.5 \pm 0.02$ & $0.5 \pm 0.0$ & $0.62 \pm 0.02$ & $0.5 \pm 0.0$\\

& $4.0$ & $0.52 \pm 0.0$ & $0.5 \pm 0.0$ & $0.5 \pm 0.01$ & $0.5 \pm 0.0$ & $0.62 \pm 0.02$ & $0.5 \pm 0.0$\\

& $5.0$ & $0.52 \pm 0.01$ & $0.5 \pm 0.0$ & $0.5 \pm 0.0$ & $0.5 \pm 0.0$ & $0.62 \pm 0.01$ & $0.5 \pm 0.0$\\

& $6.0$ & $0.52 \pm 0.01$ & $0.5 \pm 0.0$ & $0.49 \pm 0.02$ & $0.5 \pm 0.0$ & $0.62 \pm 0.01$ & $0.5 \pm 0.0$\\

& $7.0$ & $0.52 \pm 0.01$ & $0.5 \pm 0.0$ & $0.5 \pm 0.01$ & $0.5 \pm 0.0$ & $0.63 \pm 0.01$ & $0.5 \pm 0.0$\\

& $8.0$ & $0.52 \pm 0.01$ & $0.5 \pm 0.0$ & $0.48 \pm 0.02$ & $0.5 \pm 0.0$ & $0.62 \pm 0.02$ & $0.5 \pm 0.0$\\

& $9.0$ & $0.51 \pm 0.01$ & $0.5 \pm 0.0$ & $0.5 \pm 0.01$ & $0.5 \pm 0.0$ & $0.63 \pm 0.02$ & $0.5 \pm 0.0$\\

& $10.0$ & $0.52 \pm 0.01$ & $0.5 \pm 0.0$ & $0.49 \pm 0.01$ & $0.5 \pm 0.0$ & $0.63 \pm 0.02$ & $0.5 \pm 0.0$\\
\hline

\multirow{10}{*}{\toolb}
& $1.0$ & $0.52 \pm 0.01$ & $0.5 \pm 0.0$ & $0.49 \pm 0.02$ & $0.5 \pm 0.0$ & $0.55 \pm 0.02$ & $0.5 \pm 0.0$\\

& $2.0$ & $0.52 \pm 0.0$ & $0.5 \pm 0.0$ & $0.49 \pm 0.01$ & $0.5 \pm 0.0$ & $0.57 \pm 0.01$ & $0.5 \pm 0.0$\\

& $3.0$ & $0.52 \pm 0.01$ & $0.5 \pm 0.0$ & $0.49 \pm 0.01$ & $0.5 \pm 0.0$ & $0.6 \pm 0.01$ & $0.5 \pm 0.0$\\

& $4.0$ & $0.52 \pm 0.01$ & $0.5 \pm 0.0$ & $0.5 \pm 0.02$ & $0.5 \pm 0.0$ & $0.62 \pm 0.02$ & $0.5 \pm 0.0$\\

& $5.0$ & $0.52 \pm 0.01$ & $0.5 \pm 0.0$ & $0.5 \pm 0.01$ & $0.5 \pm 0.0$ & $0.63 \pm 0.02$ & $0.5 \pm 0.0$\\

& $6.0$ & $0.52 \pm 0.0$ & $0.5 \pm 0.0$ & $0.49 \pm 0.01$ & $0.5 \pm 0.0$ & $0.62 \pm 0.01$ & $0.5 \pm 0.0$\\

& $7.0$ & $0.52 \pm 0.01$ & $0.5 \pm 0.0$ & $0.49 \pm 0.01$ & $0.5 \pm 0.0$ & $0.64 \pm 0.01$ & $0.5 \pm 0.0$\\

& $8.0$ & $0.51 \pm 0.0$ & $0.5 \pm 0.0$ & $0.5 \pm 0.0$ & $0.5 \pm 0.0$ & $0.64 \pm 0.01$ & $0.5 \pm 0.0$\\

& $9.0$ & $0.52 \pm 0.01$ & $0.5 \pm 0.0$ & $0.5 \pm 0.01$ & $0.5 \pm 0.0$ & $0.64 \pm 0.02$ & $0.5 \pm 0.0$\\

& $10.0$ & $0.52 \pm 0.0$ & $0.5 \pm 0.0$ & $0.49 \pm 0.02$ & $0.5 \pm 0.0$ & $0.65 \pm 0.02$ & $0.5 \pm 0.0$\\
\hline

\end{tabular}
}
\label{tab:ind_2_attack_unbalanced_hi}
\end{table*}

\begin{table*}[th!]
\centering
\caption{Attack in inductive setting for low degree nodes}
\resizebox{\textwidth}{!}{
\begin{tabular}{c c c c c c c c }
\toprule
Model & Epsilon & \multicolumn{2}{c}{TwitchDE} & \multicolumn{2}{c}{TwitchPTBR} & \multicolumn{2}{c}{TwitchRU}\\\hline
 & & \baselineAttack & \sotaAttack & \baselineAttack & \sotaAttack & \baselineAttack & \sotaAttack\\\hline

\mlp
& $\infty$ & $0.46 \pm 0.03$ & $0.5 \pm 0.0$ & $0.48 \pm 0.01$ & $0.5 \pm 0.0$ & $0.43 \pm 0.06$ & $0.5 \pm 0.0$\\\hline
\gcn
& $\infty$ & $0.6 \pm 0.1$ & $1.0 \pm 0.0$ & $0.64 \pm 0.04$ & $1.0 \pm 0.0$ & $0.6 \pm 0.07$ & $1.0 \pm 0.0$\\\hline
\toola
& $\infty$ & $0.48 \pm 0.08$ & $0.5 \pm 0.0$ & $0.47 \pm 0.02$ & $0.5 \pm 0.0$ & $0.47 \pm 0.05$ & $0.5 \pm 0.0$\\\hline
\toolb
& $\infty$ & $0.47 \pm 0.08$ & $0.5 \pm 0.0$ & $0.48 \pm 0.02$ & $0.5 \pm 0.0$ & $0.46 \pm 0.05$ & $0.5 \pm 0.0$\\\hline

\multirow{10}{*}{\dpgcn}
& $1.0$ & $0.51 \pm 0.06$ & $0.47 \pm 0.03$ & $0.48 \pm 0.02$ & $0.53 \pm 0.01$ & $0.53 \pm 0.11$ & $0.53 \pm 0.03$\\

& $2.0$ & $0.51 \pm 0.07$ & $0.5 \pm 0.07$ & $0.48 \pm 0.04$ & $0.54 \pm 0.03$ & $0.52 \pm 0.07$ & $0.54 \pm 0.05$\\

& $3.0$ & $0.5 \pm 0.12$ & $0.52 \pm 0.05$ & $0.46 \pm 0.02$ & $0.63 \pm 0.02$ & $0.5 \pm 0.07$ & $0.58 \pm 0.05$\\

& $4.0$ & $0.54 \pm 0.09$ & $0.55 \pm 0.07$ & $0.5 \pm 0.04$ & $0.8 \pm 0.04$ & $0.53 \pm 0.03$ & $0.62 \pm 0.05$\\

& $5.0$ & $0.53 \pm 0.12$ & $0.66 \pm 0.04$ & $0.5 \pm 0.05$ & $0.89 \pm 0.05$ & $0.56 \pm 0.06$ & $0.68 \pm 0.09$\\

& $6.0$ & $0.5 \pm 0.05$ & $0.78 \pm 0.1$ & $0.54 \pm 0.03$ & $0.94 \pm 0.03$ & $0.58 \pm 0.11$ & $0.8 \pm 0.04$\\

& $7.0$ & $0.46 \pm 0.1$ & $0.83 \pm 0.1$ & $0.57 \pm 0.03$ & $0.97 \pm 0.01$ & $0.58 \pm 0.07$ & $0.87 \pm 0.03$\\

& $8.0$ & $0.52 \pm 0.08$ & $0.88 \pm 0.1$ & $0.61 \pm 0.04$ & $0.98 \pm 0.01$ & $0.6 \pm 0.08$ & $0.95 \pm 0.03$\\

& $9.0$ & $0.57 \pm 0.07$ & $0.95 \pm 0.04$ & $0.58 \pm 0.02$ & $0.99 \pm 0.01$ & $0.64 \pm 0.04$ & $0.98 \pm 0.02$\\

& $10.0$ & $0.6 \pm 0.11$ & $0.93 \pm 0.04$ & $0.6 \pm 0.03$ & $0.99 \pm 0.01$ & $0.6 \pm 0.06$ & $0.98 \pm 0.02$\\
\hline

\multirow{10}{*}{\toola}
& $1.0$ & $0.51 \pm 0.06$ & $0.5 \pm 0.0$ & $0.46 \pm 0.02$ & $0.5 \pm 0.0$ & $0.5 \pm 0.06$ & $0.5 \pm 0.0$\\

& $2.0$ & $0.51 \pm 0.03$ & $0.5 \pm 0.0$ & $0.45 \pm 0.02$ & $0.5 \pm 0.0$ & $0.49 \pm 0.07$ & $0.5 \pm 0.0$\\

& $3.0$ & $0.51 \pm 0.05$ & $0.5 \pm 0.0$ & $0.46 \pm 0.01$ & $0.5 \pm 0.0$ & $0.5 \pm 0.08$ & $0.5 \pm 0.0$\\

& $4.0$ & $0.49 \pm 0.05$ & $0.5 \pm 0.0$ & $0.45 \pm 0.02$ & $0.5 \pm 0.0$ & $0.49 \pm 0.06$ & $0.5 \pm 0.0$\\

& $5.0$ & $0.5 \pm 0.05$ & $0.5 \pm 0.0$ & $0.45 \pm 0.02$ & $0.5 \pm 0.0$ & $0.49 \pm 0.06$ & $0.5 \pm 0.0$\\

& $6.0$ & $0.48 \pm 0.06$ & $0.5 \pm 0.0$ & $0.46 \pm 0.03$ & $0.5 \pm 0.0$ & $0.5 \pm 0.07$ & $0.5 \pm 0.0$\\

& $7.0$ & $0.51 \pm 0.05$ & $0.5 \pm 0.0$ & $0.46 \pm 0.02$ & $0.5 \pm 0.0$ & $0.49 \pm 0.06$ & $0.5 \pm 0.0$\\

& $8.0$ & $0.5 \pm 0.05$ & $0.5 \pm 0.0$ & $0.45 \pm 0.02$ & $0.5 \pm 0.0$ & $0.49 \pm 0.06$ & $0.5 \pm 0.0$\\

& $9.0$ & $0.51 \pm 0.03$ & $0.5 \pm 0.0$ & $0.45 \pm 0.03$ & $0.5 \pm 0.0$ & $0.5 \pm 0.07$ & $0.5 \pm 0.0$\\

& $10.0$ & $0.5 \pm 0.07$ & $0.5 \pm 0.0$ & $0.45 \pm 0.01$ & $0.5 \pm 0.0$ & $0.5 \pm 0.08$ & $0.5 \pm 0.0$\\
\hline

\multirow{10}{*}{\toolb}
& $1.0$ & $0.48 \pm 0.11$ & $0.5 \pm 0.0$ & $0.47 \pm 0.02$ & $0.5 \pm 0.0$ & $0.49 \pm 0.07$ & $0.5 \pm 0.0$\\

& $2.0$ & $0.5 \pm 0.0$ & $0.5 \pm 0.0$ & $0.45 \pm 0.02$ & $0.5 \pm 0.0$ & $0.5 \pm 0.08$ & $0.5 \pm 0.0$\\

& $3.0$ & $0.48 \pm 0.1$ & $0.5 \pm 0.0$ & $0.46 \pm 0.02$ & $0.5 \pm 0.0$ & $0.49 \pm 0.06$ & $0.5 \pm 0.0$\\

& $4.0$ & $0.49 \pm 0.05$ & $0.5 \pm 0.0$ & $0.45 \pm 0.02$ & $0.5 \pm 0.0$ & $0.49 \pm 0.07$ & $0.5 \pm 0.0$\\

& $5.0$ & $0.49 \pm 0.11$ & $0.5 \pm 0.0$ & $0.46 \pm 0.02$ & $0.5 \pm 0.0$ & $0.49 \pm 0.07$ & $0.5 \pm 0.0$\\

& $6.0$ & $0.5 \pm 0.05$ & $0.5 \pm 0.0$ & $0.46 \pm 0.02$ & $0.5 \pm 0.0$ & $0.49 \pm 0.07$ & $0.5 \pm 0.0$\\

& $7.0$ & $0.49 \pm 0.08$ & $0.5 \pm 0.0$ & $0.47 \pm 0.02$ & $0.5 \pm 0.0$ & $0.49 \pm 0.06$ & $0.5 \pm 0.0$\\

& $8.0$ & $0.5 \pm 0.07$ & $0.5 \pm 0.0$ & $0.45 \pm 0.03$ & $0.5 \pm 0.0$ & $0.49 \pm 0.07$ & $0.5 \pm 0.0$\\

& $9.0$ & $0.5 \pm 0.07$ & $0.5 \pm 0.0$ & $0.46 \pm 0.02$ & $0.5 \pm 0.0$ & $0.49 \pm 0.07$ & $0.5 \pm 0.0$\\

& $10.0$ & $0.47 \pm 0.06$ & $0.5 \pm 0.0$ & $0.47 \pm 0.02$ & $0.5 \pm 0.0$ & $0.49 \pm 0.07$ & $0.5 \pm 0.0$\\
\hline

\end{tabular}
}
\label{tab:ind_1_attack_unbalanced_lo}
\end{table*}

\begin{table*}[th!]
\centering
\caption{Attack in inductive setting for low degree nodes}
\resizebox{\textwidth}{!}{
\begin{tabular}{c c c c c c c c }
\toprule
Model & Epsilon & \multicolumn{2}{c}{TwitchENGB} & \multicolumn{2}{c}{TwitchFR} & \multicolumn{2}{c}{Flickr}\\\hline
 & & \baselineAttack & \sotaAttack & \baselineAttack & \sotaAttack & \baselineAttack & \sotaAttack\\\hline

\mlp
& $\infty$ & $0.44 \pm 0.06$ & $0.5 \pm 0.0$ & $0.45 \pm 0.06$ & $0.5 \pm 0.0$ & $0.55 \pm 0.15$ & $0.5 \pm 0.0$\\\hline
\gcn
& $\infty$ & $0.64 \pm 0.02$ & $1.0 \pm 0.0$ & $0.6 \pm 0.06$ & $1.0 \pm 0.0$ & $0.72 \pm 0.14$ & $1.0 \pm 0.0$\\\hline
\toola
& $\infty$ & $0.5 \pm 0.05$ & $0.5 \pm 0.0$ & $0.48 \pm 0.05$ & $0.5 \pm 0.0$ & $0.59 \pm 0.12$ & $0.5 \pm 0.0$\\\hline
\toolb
& $\infty$ & $0.49 \pm 0.07$ & $0.5 \pm 0.0$ & $0.46 \pm 0.07$ & $0.5 \pm 0.0$ & $0.49 \pm 0.09$ & $0.5 \pm 0.0$\\\hline

\multirow{10}{*}{\dpgcn}
& $1.0$ & $0.54 \pm 0.03$ & $0.5 \pm 0.03$ & $0.46 \pm 0.1$ & $0.51 \pm 0.06$ & $0.51 \pm 0.04$ & $0.5 \pm 0.0$\\

& $2.0$ & $0.51 \pm 0.03$ & $0.5 \pm 0.04$ & $0.49 \pm 0.12$ & $0.49 \pm 0.1$ & $0.51 \pm 0.04$ & $0.5 \pm 0.0$\\

& $3.0$ & $0.52 \pm 0.11$ & $0.5 \pm 0.04$ & $0.44 \pm 0.11$ & $0.52 \pm 0.09$ & $0.52 \pm 0.04$ & $0.5 \pm 0.0$\\

& $4.0$ & $0.52 \pm 0.08$ & $0.52 \pm 0.05$ & $0.52 \pm 0.08$ & $0.59 \pm 0.09$ & $0.52 \pm 0.04$ & $0.5 \pm 0.0$\\

& $5.0$ & $0.47 \pm 0.04$ & $0.54 \pm 0.05$ & $0.47 \pm 0.06$ & $0.71 \pm 0.1$ & $0.52 \pm 0.04$ & $0.50 \pm 0.00$\\

& $6.0$ & $0.47 \pm 0.02$ & $0.63 \pm 0.1$ & $0.52 \pm 0.06$ & $0.86 \pm 0.05$ & $0.61 \pm 0.08$ & $0.56 \pm 0.08$\\

& $7.0$ & $0.55 \pm 0.12$ & $0.8 \pm 0.07$ & $0.48 \pm 0.04$ & $0.92 \pm 0.07$ & $0.56 \pm 0.06$ & $0.58 \pm 0.07$\\

& $8.0$ & $0.51 \pm 0.09$ & $0.87 \pm 0.07$ & $0.55 \pm 0.06$ & $0.95 \pm 0.07$ & $0.60 \pm 0.17$ & $0.71 \pm 0.08$\\

& $9.0$ & $0.53 \pm 0.06$ & $0.91 \pm 0.05$ & $0.65 \pm 0.1$ & $0.97 \pm 0.05$ & $0.62 \pm 0.15$ & $0.74 \pm 0.07$\\

& $10.0$ & $0.55 \pm 0.03$ & $0.94 \pm 0.06$ & $0.56 \pm 0.09$ & $0.98 \pm 0.03$ & $0.62 \pm 0.08$ & $0.87 \pm 0.09$\\
\hline

\multirow{10}{*}{\toola}
& $1.0$ & $0.5 \pm 0.07$ & $0.5 \pm 0.0$ & $0.5 \pm 0.06$ & $0.5 \pm 0.0$ & $0.54 \pm 0.16$ & $0.5 \pm 0.0$\\

& $2.0$ & $0.5 \pm 0.08$ & $0.5 \pm 0.0$ & $0.52 \pm 0.07$ & $0.5 \pm 0.0$ & $0.56 \pm 0.12$ & $0.5 \pm 0.0$\\

& $3.0$ & $0.51 \pm 0.06$ & $0.5 \pm 0.0$ & $0.52 \pm 0.07$ & $0.5 \pm 0.0$ & $0.6 \pm 0.16$ & $0.5 \pm 0.0$\\

& $4.0$ & $0.51 \pm 0.06$ & $0.5 \pm 0.0$ & $0.5 \pm 0.07$ & $0.5 \pm 0.0$ & $0.58 \pm 0.15$ & $0.5 \pm 0.0$\\

& $5.0$ & $0.49 \pm 0.07$ & $0.5 \pm 0.0$ & $0.51 \pm 0.03$ & $0.5 \pm 0.0$ & $0.52 \pm 0.22$ & $0.5 \pm 0.0$\\

& $6.0$ & $0.52 \pm 0.08$ & $0.5 \pm 0.0$ & $0.5 \pm 0.05$ & $0.5 \pm 0.0$ & $0.64 \pm 0.16$ & $0.5 \pm 0.0$\\

& $7.0$ & $0.49 \pm 0.07$ & $0.5 \pm 0.0$ & $0.51 \pm 0.04$ & $0.5 \pm 0.0$ & $0.64 \pm 0.14$ & $0.5 \pm 0.0$\\

& $8.0$ & $0.49 \pm 0.07$ & $0.5 \pm 0.0$ & $0.52 \pm 0.06$ & $0.5 \pm 0.0$ & $0.64 \pm 0.14$ & $0.5 \pm 0.0$\\

& $9.0$ & $0.51 \pm 0.07$ & $0.5 \pm 0.0$ & $0.49 \pm 0.05$ & $0.5 \pm 0.0$ & $0.64 \pm 0.14$ & $0.5 \pm 0.0$\\

& $10.0$ & $0.49 \pm 0.06$ & $0.5 \pm 0.0$ & $0.51 \pm 0.04$ & $0.5 \pm 0.0$ & $0.64 \pm 0.13$ & $0.5 \pm 0.0$\\
\hline

\multirow{10}{*}{\toolb}
& $1.0$ & $0.51 \pm 0.09$ & $0.5 \pm 0.0$ & $0.52 \pm 0.05$ & $0.5 \pm 0.0$ & $0.53 \pm 0.1$ & $0.5 \pm 0.0$\\

& $2.0$ & $0.49 \pm 0.06$ & $0.5 \pm 0.0$ & $0.51 \pm 0.04$ & $0.5 \pm 0.0$ & $0.57 \pm 0.15$ & $0.5 \pm 0.0$\\

& $3.0$ & $0.5 \pm 0.07$ & $0.5 \pm 0.0$ & $0.5 \pm 0.06$ & $0.5 \pm 0.0$ & $0.54 \pm 0.14$ & $0.5 \pm 0.0$\\

& $4.0$ & $0.51 \pm 0.08$ & $0.5 \pm 0.0$ & $0.51 \pm 0.06$ & $0.5 \pm 0.0$ & $0.59 \pm 0.2$ & $0.5 \pm 0.0$\\

& $5.0$ & $0.5 \pm 0.06$ & $0.5 \pm 0.0$ & $0.51 \pm 0.06$ & $0.5 \pm 0.0$ & $0.44 \pm 0.25$ & $0.5 \pm 0.0$\\

& $6.0$ & $0.51 \pm 0.07$ & $0.5 \pm 0.0$ & $0.49 \pm 0.09$ & $0.5 \pm 0.0$ & $0.53 \pm 0.08$ & $0.5 \pm 0.0$\\

& $7.0$ & $0.5 \pm 0.07$ & $0.5 \pm 0.0$ & $0.5 \pm 0.08$ & $0.5 \pm 0.0$ & $0.55 \pm 0.11$ & $0.5 \pm 0.0$\\

& $8.0$ & $0.49 \pm 0.05$ & $0.5 \pm 0.0$ & $0.5 \pm 0.07$ & $0.5 \pm 0.0$ & $0.56 \pm 0.17$ & $0.5 \pm 0.0$\\

& $9.0$ & $0.51 \pm 0.07$ & $0.5 \pm 0.0$ & $0.5 \pm 0.09$ & $0.5 \pm 0.0$ & $0.45 \pm 0.26$ & $0.5 \pm 0.0$\\

& $10.0$ & $0.51 \pm 0.08$ & $0.5 \pm 0.0$ & $0.5 \pm 0.07$ & $0.5 \pm 0.0$ & $0.58 \pm 0.17$ & $0.5 \pm 0.0$\\
\hline

\end{tabular}
}
\label{tab:ind_2_attack_unbalanced_lo}
\end{table*}

\subsection{LPA with low AUC on \mlp \& \tool}
\label{sec:appdx-low-auc}

\paragraph{Datasets} \ash{First, we construct a bipartite graph, as they appear in online dating applications, with two clusters. The first cluster has $500$ nodes and the second one has $400$ nodes. All the edges in this graph connect one node from the first cluster to another from the second cluster. For each such node pair we sample an edge with a probability $0.05$. The assigned cluster represents the label of each node. We design node features such that they do not correlate with the graph structure to minimize the leakage of edge information from the node features. Also, in such setting, \tool will have to depend on just the edge information to achieve better performance than \mlp. For that we first assign the features $[1, -1]$ for all nodes in the first cluster and $[-1, 1]$ for the ones in the second cluster. We then flip the features of $25\%$ of the nodes in the first cluster to $[-1, 1]$ and for about $62.5\%$ of the nodes in the second cluster to $[1, -1]$. We split the nodes in the dataset equally for training and testing. Out of the training nodes, we use $30\%$ for validation.}

\ash{Second, we consider a commonly used real-world citation dataset called Chameleon. It is used as a challenge dataset for GNNs as the features and edges do not correlate well with the ground-truth labels (clusters). It is typically considered as a heterophilous dataset. The dataset consists of $2,277$ nodes and $31,421$ edges. It is used in the transductive setting and already comes with $10$ random splits of the nodes with $48\%$ of them used for training, $32\%$ for validation, and $20\%$ for testing.}

\paragraph{Training procedure \& Metrics} \ash{We use the same architectures for \gcn, \mlp, and \tool as we did for other datasets to learn the embeddings. We first find the best model parameters and training hyperparameters for training  using grid search as described in Section~\ref{sec:eval-setup}. We then use the best performing models to determine the attack performance following the same procedures described in Section~\ref{sec:eval-setup}. We run both the training and attacks for $30$ seeds.}

\end{document}